\documentclass[letterpaper]{article} 
\usepackage[draft]{aaai2026} 
\usepackage[T1]{fontenc}
\usepackage{times}  
\usepackage{helvet}  
\usepackage{courier}  
\usepackage[hyphens]{url}  
\usepackage{graphicx} 
\usepackage{natbib}  
\usepackage{caption} 
\usepackage{newfloat}
\usepackage{listings}
\DeclareCaptionStyle{ruled}{labelfont=normalfont,labelsep=colon,strut=off} 
\DeclareFloatingEnvironment[fileext=lst,placement=tb,name=Listing]{listing}
\lstdefinestyle{prompt}{%
	basicstyle={\scriptsize\ttfamily},%
	columns=fullflexible,keepspaces=true,%
	breaklines=true,breakautoindent=false,breakindent=0pt,%
	showstringspaces=false,tabsize=2,upquote=true,xleftmargin=0pt,%
	escapeinside={(*}{*)},%
	literate={"}{{\textquotedbl}}1}

\usepackage{upquote} 
\usepackage{amsmath,amssymb}
\usepackage{booktabs}
\usepackage{multirow}
\usepackage{colortbl} 
\definecolor{cellbest}{RGB}{163,213,175}
\definecolor{cellnear}{RGB}{208,235,214}
\definecolor{cellclose}{RGB}{236,246,238}
\definecolor{cellmid}{RGB}{252,244,212}
\definecolor{cellfar}{RGB}{250,232,230}
\definecolor{cellworst}{RGB}{244,211,204}
\usepackage{tikz}
\usetikzlibrary{arrows.meta,positioning,fit,backgrounds,calc}
\usepackage{xspace}

\newcommand{\method}{TabDAG\xspace}

\newcommand{\rawarm}{Raw\xspace}
\newcommand{\openfe}{OpenFE\xspace}
\newcommand{\llmfe}{LLM-FE\xspace}
\newcommand{\llmfeof}{\llmfe$\cup$OF\xspace}
\newcommand{\methodof}{\method$\cup$OF\xspace}
\newcommand{\pythonarm}{Python\xspace}
\newcommand{\jsonarm}{JSON\xspace}

\newcommand{\catboost}{CatBoost\xspace}
\newcommand{\tabm}{TabM\xspace}
\newcommand{\tabpfn}{TabPFN\xspace}

\newcommand{\dfit}{\texttt{df\_fit}}
\newcommand{\yfit}{\texttt{y\_fit}}
\newcommand{\df}{\texttt{df}}

\definecolor{dagraw}{RGB}{236,236,236}
\definecolor{dagrawborder}{RGB}{115,115,115}
\definecolor{dagrow}{RGB}{219,234,254}
\definecolor{dagrowborder}{RGB}{54,104,168}
\definecolor{dagagg}{RGB}{253,228,207}
\definecolor{daggborder}{RGB}{197,108,40}
\definecolor{dagnew}{RGB}{213,240,219}
\definecolor{dagnewborder}{RGB}{47,132,72}
\definecolor{dagdel}{RGB}{249,218,214}
\definecolor{dagdelborder}{RGB}{182,71,60}
\definecolor{dagedit}{RGB}{253,243,205}
\definecolor{dageditborder}{RGB}{171,124,20}
\definecolor{dagest}{RGB}{245,245,245}
\tikzset{
  dagbase/.style={draw, rounded corners=1.5pt, align=left, inner sep=2.6pt,
                  font=\scriptsize, line width=0.7pt},
  rawstrip/.style={dagbase, fill=dagraw, draw=dagrawborder},
  rownode/.style={dagbase, fill=dagrow, draw=dagrowborder},
  aggnode/.style={dagbase, fill=dagagg, draw=daggborder},
  interm/.style={densely dashed},
  addednode/.style={fill=dagnew, draw=dagnewborder},
  deletednode/.style={fill=dagdel, draw=dagdelborder},
  editednode/.style={fill=dagedit, draw=dageditborder},
  keptnode/.style={fill=dagrow!45!white, draw=dagrowborder!45!white, text=black!55},
  estnode/.style={dagbase, fill=dagest, draw=black!70},
  dagedge/.style={-{Stealth[length=1.6mm]}, semithick, draw=black!55},
  dagnote/.style={font=\tiny\itshape, text=black!60, align=left},
}
\newcommand{\annot}[1]{{\color{black!55}\itshape #1}}

\title{TabDAG: Structured Evolutionary Search for Tabular Feature Engineering}
\author{
    Anastasia Volkova\textsuperscript{1}, Andrey Kuznetsov\textsuperscript{1},\\
    Ivan Oseledets\textsuperscript{2}, Valentin Khrulkov\textsuperscript{1,2}
}
\affiliations{
    \textsuperscript{1}FusionBrain Lab \quad \textsuperscript{2}AXXX\\
    \texttt{a.volkova@fusionbrainlab.com}, \texttt{kuznetsoff.andrey@gmail.com}\\
    \texttt{Ivan.oseledets@gmail.com}, \texttt{khrulkov.v@gmail.com}
}

\begin{document}

\maketitle
\pagestyle{plain}

\begin{abstract}
Feature engineering can improve tabular prediction, but useful transformations are compositional, data-dependent, and subject to leakage constraints. We present TabDAG, a framework that combines evolutionary search with a graph representation of feature-engineering programs. Nodes implement row-wise or data-fitted transformations; edges encode dependencies among raw variables, intermediate values, and final features. The representation supports interactions, statistical encodings, clustering- and neighborhood-based features, and multi-stage compositions in a single executable program. Mutations are structured edits constrained by a parent-specific schema and checked before evaluation. In our experiments, no structurally invalid program reached evaluation, although proposals could be rejected and valid graphs could still fail at runtime. Across ten regression and classification datasets and three tabular model families, evolved features improve predictive performance over raw inputs. TabDAG and its combination with OpenFE achieve the two best mean ranks among the compared methods; pairwise rank differences between feature-engineering methods do not survive multiple-testing correction. The LLM-based searches use the same backbone and candidate-evaluation budget. A controlled ablation compares free-form Python, whole-graph JSON, and constrained graph edits with the search engine, budget, and backbone held fixed. Constrained edits yield fewer failed evaluations than either rewrite baseline. The resulting graphs expose the transformations and dependencies selected by search, enabling feature-level analysis of the observed gains.
\end{abstract}

\section{Introduction}
\label{sec:intro}

On tabular data, inputs can matter as much as estimators: a ratio between raw
columns, or a per-group target statistic, may move a gradient-boosted tree
further than a change of model. Automating feature engineering must handle
three things at once: useful transformations are compositional (the
informative quantity is often a function of quantities built first),
data-dependent (target encodings and learned bins carry state fitted only on
the training split), and constrained (a feature that consults
its own row's target scores well under cross-validation and fails in use).

Two lines of work automate parts of this. Expansion-and-selection systems
apply a fixed operator vocabulary and prune the result statistically
\citep{zhang2023openfe,kanter2015dfs}, but what a column means
never enters the search. Language models supply that missing information:
prompted with column names and a task description, they propose transformations
reading as domain knowledge rather than arithmetic over anonymous variables
\citep{hollmann2023caafe,han2024featllm,nam2024octree,abhyankar2025llmfe}.
But the candidate is free-form code: nothing guarantees a valid column
reference, train-only fitting, or preservation of what worked in the parent,
and each failure surfaces only when the program runs.

We make these constraints explicit in the representation. A candidate is a typed feature
directed acyclic graph (DAG) whose nodes declare their input columns, typed
output columns and the code
computing them, under two interfaces separating row-wise computation from
computation fitting state on a frozen training frame. A mutation is a diff
against one parent, decoded under a schema regenerated for that parent, so a
kept-node reference to a node the parent lacks cannot be produced. The
decoded edit is then checked before evaluation; runtime and leakage probes
remain necessary for the generated code. Fig.~\ref{fig:anatomy} shows both objects as real
artifacts from our runs.

Our contributions are: (i)~a graph-native encoding carrying composition, output
types and the train/query boundary in the representation rather than by
convention; (ii)~a schema-constrained per-parent diff as the mutation operator,
with an accounting separating proposals rejected before evaluation from
programs failing at runtime; (iii)~an experiment placing this encoding against a
search-based and an LLM-based baseline under one protocol, with an ablation
isolating the encoding; and (iv)~an account of where the gain comes from: a
search-free control, a feature-level comparison of what each arm builds, and a
per-estimator reading of when engineered features pay.

\section{Related Work}
\label{sec:related}

\paragraph{Automated feature engineering.}
OpenFE enumerates binary and group operators over the raw frame and prunes them
by successive halving \citep{zhang2023openfe}; Deep Feature Synthesis composes
aggregation and transformation primitives across related tables
\citep{kanter2015dfs}; learned or evolutionary policies replace enumeration
over the same vocabulary \citep{katz2016explorekit,chen2019nfs,zhu2022difer},
some transferring a fitted policy to unseen datasets
\citep{nargesian2017lfe,khurana2018transgraph}. Composition to depth is
reachable; what a fixed vocabulary cannot supply is an operator
specific to the dataset --- and since tabular estimators are already
strong on raw inputs \citep{grinsztajn2022why}, those may be the columns worth
adding.

\paragraph{Language models as feature proposers.}
CAAFE keeps a code edit when a validation score improves
\citep{hollmann2023caafe}; FeatLLM translates textual rules into binary columns
\citep{han2024featllm}; OCTree refines a rule under feedback summarized from a
decision tree \citep{nam2024octree}; LLM-FE evolves complete Python feature
functions under an island model \citep{abhyankar2025llmfe}. In all four the
candidate is free-form code, so validity is established by execution. Such
proposals are dominated by simple binary arithmetic
\citep{kueken2024operatorbias}, and popular tabular benchmarks are partly
memorized by the models proposing the features \citep{bordt2024memorization}
--- two reasons to report what an arm builds, and how much gain precedes
search, rather than a final score alone.

\paragraph{Program search with language models.}
ELM pairs a diff-emitting code model with MAP-Elites
\citep{lehman2022elm,mouret2015mapelites}; FunSearch
\citep{romeraparedes2024funsearch}, AlphaEvolve \citep{novikov2025alphaevolve}
and EoH \citep{liu2024eoh} evolve programs under, respectively, an island
model, exact-match replacement blocks, and idea--implementation co-evolution.
In each the edit is text parsed after it is generated; we keep the loop and
change the object it moves.

\paragraph{Constrained decoding and reuse of experience.}
Grammar- and automaton-constrained decoding restricts what a language model
can emit, so everything it produces already respects a formal syntax
\citep{geng2023grammar,willard2023efficient}. That constraint is normally
written once for the task. Ours is rebuilt for every mutation from the
parent being edited: the only node ids the schema admits are the parent's
own, so an edit naming a node that does not exist cannot even be decoded.
In the free-form arms this reference error is the largest failure mode,
caught only after the candidate is generated. Reusing experience across an
agent's steps is likewise established, including for tabular feature
generation \citep{dong2026malmas}; the supplementary material describes
what our runs carry. We are not aware of a prior study that holds engine,
budget, backbone model and prompt information fixed and varies only how
the candidate program is written.

\section{Method}
\label{sec:method}

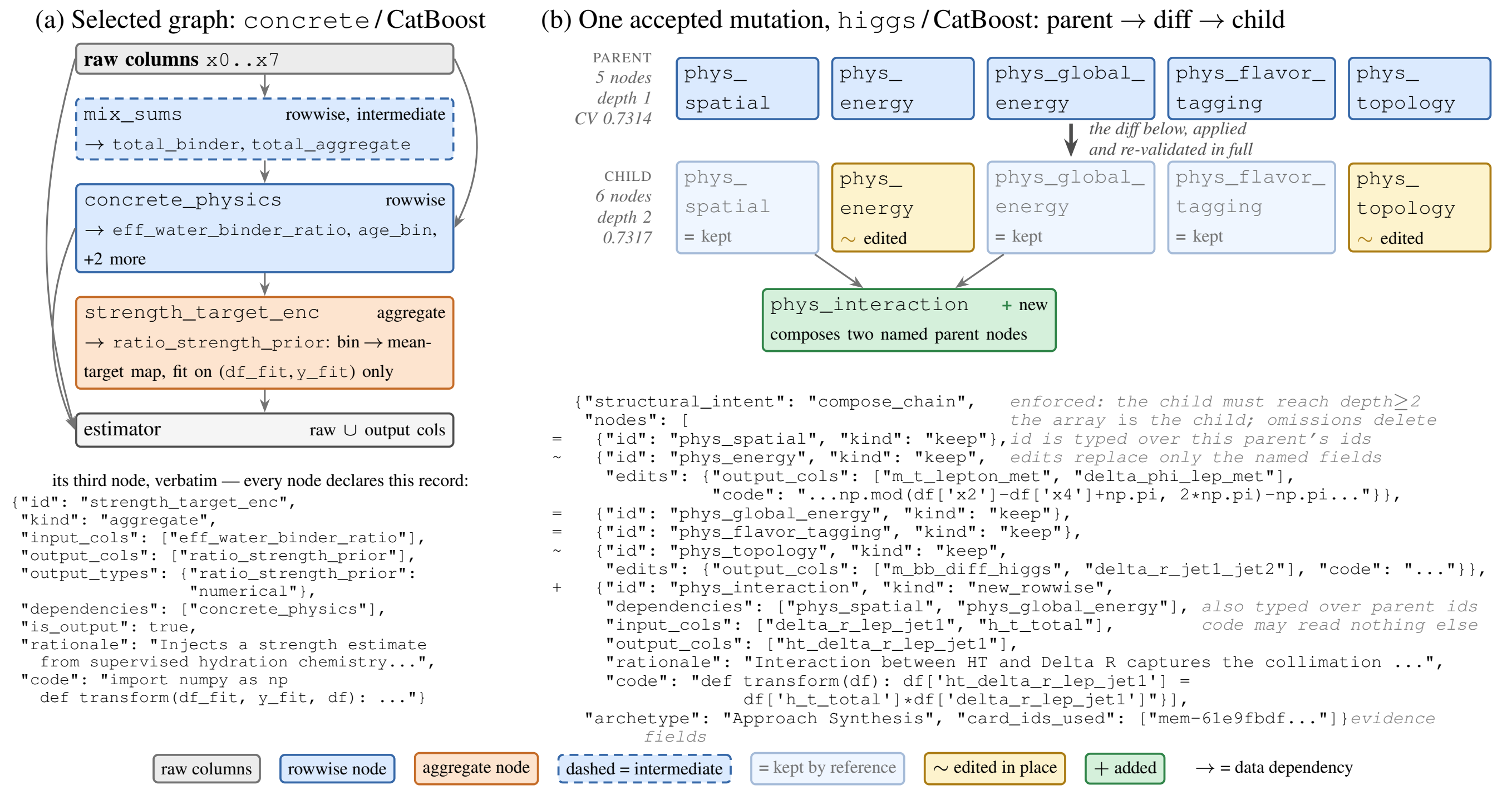
\begin{figure*}[t]
\centering
\begin{minipage}[t]{0.335\textwidth}
\centering
{\footnotesize (a) Selected graph: \texttt{concrete}\,/\,\catboost}\\[2pt]
\begin{tikzpicture}[node distance=2.7mm]
  \node[rawstrip, text width=43mm] (raw)
    {\textbf{raw columns} \texttt{x0..x7}};
  \node[rownode, interm, below=of raw, text width=43mm] (mix)
    {\texttt{mix\_sums}\hfill{\tiny rowwise, intermediate}\\
     {\tiny$\to$ \texttt{total\_binder}, \texttt{total\_aggregate}}};
  \node[rownode, below=of mix, text width=43mm] (phys)
    {\texttt{concrete\_physics}\hfill{\tiny rowwise}\\
     {\tiny$\to$ \texttt{eff\_water\_binder\_ratio}, \texttt{age\_bin},
      +2 more}};
  \node[aggnode, below=of phys, text width=43mm] (enc)
    {\texttt{strength\_target\_enc}\hfill{\tiny aggregate}\\
     {\tiny$\to$ \texttt{ratio\_strength\_prior}: bin\,$\to$\,mean-target map,
      fit on (\dfit,\,\yfit) only}};
  \node[estnode, below=of enc, text width=43mm] (est)
    {estimator \hfill {\tiny raw $\cup$ output cols}};
  \draw[dagedge] (raw) -- (mix);
  \draw[dagedge] (mix) -- (phys);
  \draw[dagedge] (phys) -- (enc);
  \draw[dagedge] (enc) -- (est);
  \draw[dagedge] (raw.east) .. controls +(3.5mm,-7mm) and +(3.5mm,7mm) .. (phys.east);
  \draw[dagedge] (raw.west) .. controls +(-4.5mm,-16mm) and +(-4.5mm,16mm) .. (est.west);
  \draw[dagedge] (phys.west) .. controls +(-3.5mm,-8mm) and +(-3.5mm,8mm) .. (est.west);
\end{tikzpicture}\\[2pt]
{\tiny its third node, verbatim --- every node declares this record:}\\[1pt]
\begin{lstlisting}[basicstyle={\fontsize{5.6}{6.0}\selectfont\ttfamily},columns=fixed]
{"id": "strength_target_enc",
 "kind": "aggregate",
 "input_cols": ["eff_water_binder_ratio"],
 "output_cols": ["ratio_strength_prior"],
 "output_types": {"ratio_strength_prior":
                  "numerical"},
 "dependencies": ["concrete_physics"],
 "is_output": true,
 "rationale": "Injects a strength estimate
   from supervised hydration chemistry...",
 "code": "import numpy as np
   def transform(df_fit, y_fit, df): ..."}
\end{lstlisting}
\end{minipage}\hfill
\begin{minipage}[t]{0.645\textwidth}
{\footnotesize (b) One accepted mutation, \texttt{higgs}\,/\,\catboost:
parent $\to$ diff $\to$ child}\\[2pt]
\begin{tikzpicture}[node distance=4.8mm and 1.4mm,
                    every node/.append style={font=\tiny}]
  \node[rownode, text width=15mm] (p1)
    {\texttt{phys\_}\\\texttt{spatial}};
  \node[rownode, text width=15mm, right=of p1] (p2)
    {\texttt{phys\_}\\\texttt{energy}};
  \node[rownode, text width=18mm, right=of p2] (p3)
    {\texttt{phys\_global\_}\\\texttt{energy}};
  \node[rownode, text width=18mm, right=of p3] (p4)
    {\texttt{phys\_flavor\_}\\\texttt{tagging}};
  \node[rownode, text width=15mm, right=of p4] (p5)
    {\texttt{phys\_}\\\texttt{topology}};
  \node[dagnote, left=1.6mm of p1, text width=12mm, align=right]
    {\textsc{parent}\\5 nodes\\depth 1\\CV 0.7314};
  \node[rownode, keptnode, text width=15mm, below=of p1] (c1)
    {\texttt{phys\_}\\\texttt{spatial}\\{\tiny\textbf{=} kept}};
  \node[rownode, editednode, text width=15mm, right=of c1] (c2)
    {\texttt{phys\_}\\\texttt{energy}\\{\tiny\textcolor{dageditborder}{\textbf{$\sim$}} edited}};
  \node[rownode, keptnode, text width=18mm, right=of c2] (c3)
    {\texttt{phys\_global\_}\\\texttt{energy}\\{\tiny\textbf{=} kept}};
  \node[rownode, keptnode, text width=18mm, right=of c3] (c4)
    {\texttt{phys\_flavor\_}\\\texttt{tagging}\\{\tiny\textbf{=} kept}};
  \node[rownode, editednode, text width=15mm, right=of c4] (c5)
    {\texttt{phys\_}\\\texttt{topology}\\{\tiny\textcolor{dageditborder}{\textbf{$\sim$}} edited}};
  \node[dagnote, left=1.6mm of c1, text width=12mm, align=right]
    {\textsc{child}\\6 nodes\\depth 2\\0.7317};
  \node[rownode, addednode, text width=33mm,
        below=4mm of $(c1.south)!0.5!(c3.south)$] (cnew)
    {\texttt{phys\_interaction}\hfill{\tiny\textcolor{dagnewborder}{\textbf{+}} new}\\
     {\tiny composes two named parent nodes}};
  \draw[dagedge] (c1) -- (cnew);
  \draw[dagedge] (c3) -- (cnew);
  \draw[-{Stealth[length=2.2mm]}, very thick, draw=black!70]
    ($(p3.south)+(0,-0.4mm)$) -- ($(c3.north)+(0,0.4mm)$)
    node[midway, right=0.8mm, font=\tiny\itshape, text=black!70, align=left]
    {the diff below, applied\\and re-validated in full};
\end{tikzpicture}\\[1pt]
\begin{lstlisting}[basicstyle={\fontsize{6}{6.3}\selectfont\ttfamily},columns=fixed,escapeinside={(*@}{@*)}]
   {"structural_intent": "compose_chain",   (*@\annot{enforced: the child must reach depth$\geq$2}@*)
    "nodes": [                              (*@\annot{the array \emph{is} the child; omissions delete}@*)
 =   {"id": "phys_spatial", "kind": "keep"},(*@\annot{\texttt{id} is typed over this parent's ids}@*)
 ~   {"id": "phys_energy", "kind": "keep",  (*@\annot{edits replace only the named fields}@*)
      "edits": {"output_cols": ["m_t_lepton_met", "delta_phi_lep_met"],
                "code": "...np.mod(df['x2']-df['x4']+np.pi, 2*np.pi)-np.pi..."}},
 =   {"id": "phys_global_energy", "kind": "keep"},
 =   {"id": "phys_flavor_tagging", "kind": "keep"},
 ~   {"id": "phys_topology", "kind": "keep",
      "edits": {"output_cols": ["m_bb_diff_higgs", "delta_r_jet1_jet2"], "code": "..."}},
 +   {"id": "phys_interaction", "kind": "new_rowwise",
      "dependencies": ["phys_spatial", "phys_global_energy"], (*@\annot{also typed over parent ids}@*)
      "input_cols": ["delta_r_lep_jet1", "h_t_total"],        (*@\annot{code may read nothing else}@*)
      "output_cols": ["ht_delta_r_lep_jet1"],
      "rationale": "Interaction between HT and Delta R captures the collimation ...",
      "code": "def transform(df): df['ht_delta_r_lep_jet1'] =
                   df['h_t_total']*df['delta_r_lep_jet1']"}],
    "archetype": "Approach Synthesis", "card_ids_used": ["mem-61e9fbdf..."]}(*@\annot{evidence fields}@*)
\end{lstlisting}
\end{minipage}\\[2pt]
\begin{tikzpicture}[node distance=1.2mm and 2.2mm]
  \node[rawstrip, font=\tiny] (l1) {raw columns};
  \node[rownode, font=\tiny, right=of l1] (l2) {rowwise node};
  \node[aggnode, font=\tiny, right=of l2] (l3) {aggregate node};
  \node[rownode, interm, font=\tiny, right=of l3] (l4) {dashed = intermediate};
  \node[rownode, keptnode, font=\tiny, right=of l4] (l5) {\textbf{=} kept by reference};
  \node[rownode, editednode, font=\tiny, right=of l5] (l6) {$\sim$ edited in place};
  \node[rownode, addednode, font=\tiny, right=of l6] (l7) {$+$ added};
  \node[font=\tiny, right=of l7] (l8) {$\to$ = data dependency};
\end{tikzpicture}
\caption{The representation, illustrated with two evaluated programs.
\textbf{(a)}~A genome is a typed DAG of feature nodes; the listing is its
third node, the record every node declares.
\emph{Rowwise} nodes map \df{} row-by-row and never see the target;
\emph{aggregate} nodes implement \texttt{transform(\dfit,\,\yfit,\,\df)} and may
fit statistics only on the frozen training frame --- here a supervised
bin-encoder computing its prior five-fold out-of-fold. \textbf{(b)}~A mutation
is a schema-constrained \emph{diff} against one parent, verbatim: it keeps three
nodes by reference, edits two, and adds one composing two named parents, raising
the graph from depth 1 to 2 as its \texttt{compose\_chain} intent requires; an
omitted parent node would be deleted. Markers and gray italics are ours.}
\label{fig:anatomy}
\end{figure*}

\method{} is evolutionary search over typed feature DAGs: an LLM proposes edits
to a graph-structured program, and an engine archives the evaluated graphs by
behavior.

\subsection{The Genome: a Typed Feature DAG}
\label{sec:genome}

A genome is a JSON document listing raw columns, at most 10 feature nodes,
optional hidden raw columns, and an optional invertible target transform
(regression only). Fig.~\ref{fig:anatomy}a shows the selected graph for
\texttt{concrete}\,/\,\catboost. Nodes declare inputs, outputs, dependencies,
output types (\texttt{numerical}, \texttt{categorical}, \texttt{binary},
\texttt{ordinal}), and pandas code \citep{mckinney2010pandas}. The estimator
receives visible raw columns and output-node columns; other nodes provide
intermediate values. Raw columns have positional names \texttt{x0..xN};
their semantics are supplied in the task description, except for the
anonymized datasets \texttt{house} and \texttt{otto}. Every run starts from
the empty graph, which reproduces the raw baseline.

The two node interfaces separate row-wise computation from fitting state.
A \emph{rowwise} node implements \texttt{transform(\df)}: it must be
row-independent and cannot access the target. An \emph{aggregate} node
implements \texttt{transform(\dfit,\,\yfit,\,\df)}: it can fit encoders,
bin maps, or group statistics only on the frozen training frame \dfit,
excluding each fitting row's own target from its features. These interfaces
make the train/query boundary explicit for transformation and aggregation
operations \citep{kanter2015dfs}.

Contracts are checked statically and behaviorally. The schema validates
identifier uniqueness, type declarations, acyclicity, and that every referenced
column is raw or produced by a declared dependency; the code then runs under
probes before any model is fit --- declared inputs only, every declared output
assigned, split invariance, target-transform round-trip, and an own-target
leakage guard that perturbs \yfit{} at individual rows and asserts those rows'
features are unchanged. The guard fires in practice, on every arm (\S\ref{sec:results}).

\subsection{Mutation: a Schema-Constrained Graph Diff}
\label{sec:diff}

The LLM is asked for an edit rather than a program --- not itself new
(\S\ref{sec:related}); what differs is that the edit is not text to be parsed
but an instance of a schema regenerated for each parent. Fig.~\ref{fig:anatomy}b shows an accepted diff verbatim. Its fields:

\begin{itemize}\setlength\itemsep{1pt}
\item {\raggedright \texttt{structural\_intent} $\in$ \{\texttt{local\_edit},
  \texttt{extend\_chain}, \texttt{compose\_chain},
  \texttt{simplify\_graph}\} --- a declared plan that is \emph{enforced}, not
  advisory: \texttt{extend\_chain} must deepen the parent,
  \texttt{compose\_chain} must yield a chain of depth $\geq 2$.\par}
\item a node array that is the \emph{complete} child. Each entry either
  retains a parent node by reference --- optionally with edits replacing only
  the named fields --- or introduces a
  \texttt{new\_rowwise}\,/\,\texttt{new\_aggregate} node in full; any parent
  node not listed is deleted. Because the schema is rebuilt per parent, the
  only \texttt{id} values it admits for a kept node are the parent's own: a
  reference to a nonexistent node cannot be decoded.
\item \texttt{target\_change} (set / keep / drop the target transform), edits
  to the hidden-column set, and evidence fields --- the edit's archetype and
  justification.
\end{itemize}

\noindent A field the schema does not declare is rejected wherever it appears,
and node and dependency counts are capped. Applying a decoded diff assembles
the child, runs a few deterministic repairs (e.g., inferring a dependency the
code plainly reads), and re-validates in full before persisting: a
structurally invalid child is never materialized.

We distinguish rejected proposals from failed evaluations. \method{} checks
each proposed edit before evaluating the resulting program. An invalid edit
therefore costs an LLM call but does not consume the candidate-evaluation
budget. Across the 30 dataset--estimator combinations
(\S\ref{sec:setup}), 1{,}558 of 4{,}549 proposals (34.2\%) were rejected,
mainly for unavailable-column references (882) and unmet depth requirements
(384). The three \texttt{otto} combinations, with 93 raw columns, account for
775 rejections; the rate on the other datasets is 22.5\%.

The comparison is with two rewrite baselines: free-form Python
(\pythonarm) and whole-graph JSON (\jsonarm). They have no equivalent
pre-evaluation gate, so invalid programs can consume evaluations. Among
3{,}000 evaluated candidates per method, structural failures numbered
0 for \method{}, 4 for \pythonarm{}, and at least 86 for \jsonarm{}.
Including runtime and semantic failures, the totals were 602 (20.1\%),
838 (27.9\%), and 814 (27.1\%), respectively
(\S\ref{sec:results}). Thus the benefit is fewer failed evaluations,
not error-free generation or fewer LLM calls.

\subsection{Search Engine}
\label{sec:engine}

MAP-Elites \citep{mouret2015mapelites} retains high-scoring candidates with
different predictive behavior. Fitness is the estimator's metric under
$k$-fold cross-validation on the development split ($k{=}3$, shuffled,
fixed seed). Two measurements determine a candidate's archive location:
local smoothness, the 95th-percentile prediction slope under small input
perturbations; and covariate-shift sensitivity, the typical prediction
change after shifting every query feature by $+1.5$ standard deviations.

These measurements compare prediction responses, not code text. Programs
with different code but similar responses can occupy the same archive
region. The aim is behavioral diversity, not a large prediction change
from every mutation. Each run evaluates the identity founder and 100 mutant
children; parents are sampled from archive elites with weight increasing
in fitness. The representation and validators remain in the task plug-in,
separate from the search engine.

The engine is a general-purpose evolutionary framework, used without
tabular-specific modification and left unnamed for anonymity; an anonymized
implementation with every arm's complete run configuration accompanies this
submission.

\section{Experimental Setup}
\label{sec:setup}

\begin{table*}[t]
\centering
\scriptsize
\setlength{\tabcolsep}{4pt}
\renewcommand{\arraystretch}{0.95}
\begin{tabular}{llrrrrrr}
\toprule
 &  & \multicolumn{1}{c}{Reference} & \multicolumn{3}{c}{Baselines} & \multicolumn{2}{c}{Ours} \\
\cmidrule(lr){3-3}\cmidrule(lr){4-6}\cmidrule(lr){7-8}
Dataset & Est. & \multicolumn{1}{c}{\rawarm} & \multicolumn{1}{c}{\openfe} & \multicolumn{1}{c}{\llmfe} & \multicolumn{1}{c}{\llmfeof} & \multicolumn{1}{c}{\method} & \multicolumn{1}{c}{\methodof} \\
\midrule
\multicolumn{8}{l}{\textbf{Regression (R$^2$)}} \\
house & CatBoost & \cellcolor{cellmid}0.6729\,{\color{gray}\tiny$\pm$.0024}$^{\hphantom{\blacktriangle}}$ & \cellcolor{cellbest}\textbf{0.6821}\,{\color{gray}\tiny$\pm$.0033}$^{\blacktriangle}$ & \cellcolor{cellmid}0.6722\,{\color{gray}\tiny$\pm$.0037}$^{\hphantom{\blacktriangle}}$ & \cellcolor{cellbest}\textbf{0.6802}\,{\color{gray}\tiny$\pm$.0038}$^{\blacktriangle}$ & \cellcolor{cellmid}0.6697\,{\color{gray}\tiny$\pm$.0054}$^{\hphantom{\blacktriangle}}$ & \cellcolor{cellbest}\textbf{0.6829}\,{\color{gray}\tiny$\pm$.0028}$^{\blacktriangle}$ \\
 & TabM & \cellcolor{cellfar}0.6564\,{\color{gray}\tiny$\pm$.0079}$^{\hphantom{\blacktriangle}}$ & \cellcolor{cellbest}\textbf{0.6747}\,{\color{gray}\tiny$\pm$.0060}$^{\blacktriangle}$ & \cellcolor{cellmid}0.6612\,{\color{gray}\tiny$\pm$.0105}$^{\hphantom{\blacktriangle}}$ & \cellcolor{cellclose}0.6692\,{\color{gray}\tiny$\pm$.0127}$^{\blacktriangle}$ & \cellcolor{cellmid}0.6626\,{\color{gray}\tiny$\pm$.0076}$^{\hphantom{\blacktriangle}}$ & \cellcolor{cellclose}0.6706\,{\color{gray}\tiny$\pm$.0095}$^{\blacktriangle}$ \\
 & TabPFN & \cellcolor{cellnear}0.7496\,{\color{gray}\tiny$\pm$.0018}$^{\hphantom{\blacktriangle}}$ & \cellcolor{cellnear}0.7486\,{\color{gray}\tiny$\pm$.0021}$^{\hphantom{\blacktriangle}}$ & \cellcolor{cellnear}0.7474\,{\color{gray}\tiny$\pm$.0019}$^{\blacktriangledown}$ & \cellcolor{cellnear}0.7478\,{\color{gray}\tiny$\pm$.0014}$^{\blacktriangledown}$ & \cellcolor{cellbest}\textbf{0.7515}\,{\color{gray}\tiny$\pm$.0011}$^{\blacktriangle}$ & \cellcolor{cellclose}0.7437\,{\color{gray}\tiny$\pm$.0018}$^{\blacktriangledown}$ \\
airfoil & CatBoost & \cellcolor{cellfar}0.9460\,{\color{gray}\tiny$\pm$.0047}$^{\hphantom{\blacktriangle}}$ & \cellcolor{cellnear}0.9696\,{\color{gray}\tiny$\pm$.0013}$^{\blacktriangle}$ & \cellcolor{cellnear}0.9710\,{\color{gray}\tiny$\pm$.0015}$^{\blacktriangle}$ & \cellcolor{cellbest}\textbf{0.9732}\,{\color{gray}\tiny$\pm$.0013}$^{\blacktriangle}$ & \cellcolor{cellbest}\textbf{0.9736}\,{\color{gray}\tiny$\pm$.0007}$^{\blacktriangle}$ & \cellcolor{cellbest}\textbf{0.9732}\,{\color{gray}\tiny$\pm$.0011}$^{\blacktriangle}$ \\
 & TabM & \cellcolor{cellclose}0.9740\,{\color{gray}\tiny$\pm$.0025}$^{\hphantom{\blacktriangle}}$ & \cellcolor{cellclose}0.9742\,{\color{gray}\tiny$\pm$.0027}$^{\hphantom{\blacktriangle}}$ & \cellcolor{cellbest}\textbf{0.9788}\,{\color{gray}\tiny$\pm$.0015}$^{\blacktriangle}$ & \cellcolor{cellbest}\textbf{0.9771}\,{\color{gray}\tiny$\pm$.0020}$^{\blacktriangle}$ & \cellcolor{cellclose}0.9740\,{\color{gray}\tiny$\pm$.0014}$^{\hphantom{\blacktriangle}}$ & \cellcolor{cellnear}0.9754\,{\color{gray}\tiny$\pm$.0017}$^{\hphantom{\blacktriangle}}$ \\
 & TabPFN & \cellcolor{cellnear}0.9860\,{\color{gray}\tiny$\pm$.0003}$^{\hphantom{\blacktriangle}}$ & \cellcolor{cellnear}0.9867\,{\color{gray}\tiny$\pm$.0003}$^{\blacktriangle}$ & \cellcolor{cellbest}\textbf{0.9873}\,{\color{gray}\tiny$\pm$.0001}$^{\blacktriangle}$ & \cellcolor{cellbest}\textbf{0.9872}\,{\color{gray}\tiny$\pm$.0002}$^{\blacktriangle}$ & \cellcolor{cellnear}0.9860\,{\color{gray}\tiny$\pm$.0003}$^{\hphantom{\blacktriangle}}$ & \cellcolor{cellnear}0.9868\,{\color{gray}\tiny$\pm$.0002}$^{\blacktriangle}$ \\
protein & CatBoost & \cellcolor{cellworst}0.6148\,{\color{gray}\tiny$\pm$.0005}$^{\hphantom{\blacktriangle}}$ & \cellcolor{cellclose}0.6516\,{\color{gray}\tiny$\pm$.0008}$^{\blacktriangle}$ & \cellcolor{cellclose}0.6513\,{\color{gray}\tiny$\pm$.0008}$^{\blacktriangle}$ & \cellcolor{cellbest}\textbf{0.6592}\,{\color{gray}\tiny$\pm$.0007}$^{\blacktriangle}$ & \cellcolor{cellmid}0.6498\,{\color{gray}\tiny$\pm$.0004}$^{\blacktriangle}$ & \cellcolor{cellnear}0.6556\,{\color{gray}\tiny$\pm$.0009}$^{\blacktriangle}$ \\
 & TabM & \cellcolor{cellworst}0.6766\,{\color{gray}\tiny$\pm$.0046}$^{\hphantom{\blacktriangle}}$ & \cellcolor{cellmid}0.7076\,{\color{gray}\tiny$\pm$.0033}$^{\blacktriangle}$ & \cellcolor{cellclose}0.7120\,{\color{gray}\tiny$\pm$.0068}$^{\blacktriangle}$ & \cellcolor{cellbest}\textbf{0.7168}\,{\color{gray}\tiny$\pm$.0050}$^{\blacktriangle}$ & \cellcolor{cellmid}0.7048\,{\color{gray}\tiny$\pm$.0060}$^{\blacktriangle}$ & \cellcolor{cellclose}0.7104\,{\color{gray}\tiny$\pm$.0046}$^{\blacktriangle}$ \\
 & TabPFN & \cellcolor{cellnear}0.7693\,{\color{gray}\tiny$\pm$.0006}$^{\hphantom{\blacktriangle}}$ & \cellcolor{cellnear}0.7675\,{\color{gray}\tiny$\pm$.0010}$^{\blacktriangledown}$ & \cellcolor{cellnear}0.7690\,{\color{gray}\tiny$\pm$.0009}$^{\hphantom{\blacktriangle}}$ & \cellcolor{cellnear}0.7665\,{\color{gray}\tiny$\pm$.0011}$^{\blacktriangledown}$ & \cellcolor{cellbest}\textbf{0.7700}\,{\color{gray}\tiny$\pm$.0006}$^{\blacktriangle}$ & \cellcolor{cellnear}0.7684\,{\color{gray}\tiny$\pm$.0008}$^{\blacktriangledown}$ \\
concrete & CatBoost & \cellcolor{cellmid}0.9303\,{\color{gray}\tiny$\pm$.0024}$^{\hphantom{\blacktriangle}}$ & \cellcolor{cellclose}0.9334\,{\color{gray}\tiny$\pm$.0018}$^{\blacktriangle}$ & \cellcolor{cellnear}0.9377\,{\color{gray}\tiny$\pm$.0019}$^{\blacktriangle}$ & \cellcolor{cellnear}0.9364\,{\color{gray}\tiny$\pm$.0012}$^{\blacktriangle}$ & \cellcolor{cellbest}\textbf{0.9397}\,{\color{gray}\tiny$\pm$.0014}$^{\blacktriangle}$ & \cellcolor{cellclose}0.9350\,{\color{gray}\tiny$\pm$.0018}$^{\blacktriangle}$ \\
 & TabM & \cellcolor{cellfar}0.9207\,{\color{gray}\tiny$\pm$.0055}$^{\hphantom{\blacktriangle}}$ & \cellcolor{cellbest}\textbf{0.9412}\,{\color{gray}\tiny$\pm$.0034}$^{\blacktriangle}$ & \cellcolor{cellbest}\textbf{0.9433}\,{\color{gray}\tiny$\pm$.0038}$^{\blacktriangle}$ & \cellcolor{cellbest}\textbf{0.9443}\,{\color{gray}\tiny$\pm$.0021}$^{\blacktriangle}$ & \cellcolor{cellbest}\textbf{0.9416}\,{\color{gray}\tiny$\pm$.0044}$^{\blacktriangle}$ & \cellcolor{cellbest}\textbf{0.9452}\,{\color{gray}\tiny$\pm$.0041}$^{\blacktriangle}$ \\
 & TabPFN & \cellcolor{cellbest}\textbf{0.9545}\,{\color{gray}\tiny$\pm$.0005}$^{\hphantom{\blacktriangle}}$ & \cellcolor{cellclose}0.9508\,{\color{gray}\tiny$\pm$.0004}$^{\blacktriangledown}$ & \cellcolor{cellbest}\textbf{0.9545}\,{\color{gray}\tiny$\pm$.0005}$^{\hphantom{\blacktriangle}}$ & \cellcolor{cellclose}0.9508\,{\color{gray}\tiny$\pm$.0004}$^{\blacktriangledown}$ & \cellcolor{cellbest}\textbf{0.9550}\,{\color{gray}\tiny$\pm$.0005}$^{\hphantom{\blacktriangle}}$ & \cellcolor{cellnear}0.9527\,{\color{gray}\tiny$\pm$.0004}$^{\blacktriangledown}$ \\
miami & CatBoost & \cellcolor{cellnear}0.9347\,{\color{gray}\tiny$\pm$.0010}$^{\hphantom{\blacktriangle}}$ & \cellcolor{cellclose}0.9331\,{\color{gray}\tiny$\pm$.0015}$^{\blacktriangledown}$ & \cellcolor{cellnear}0.9347\,{\color{gray}\tiny$\pm$.0010}$^{\hphantom{\blacktriangle}}$ & \cellcolor{cellclose}0.9331\,{\color{gray}\tiny$\pm$.0015}$^{\blacktriangledown}$ & \cellcolor{cellbest}\textbf{0.9381}\,{\color{gray}\tiny$\pm$.0009}$^{\blacktriangle}$ & \cellcolor{cellnear}0.9355\,{\color{gray}\tiny$\pm$.0004}$^{\blacktriangle}$ \\
 & TabM & \cellcolor{cellclose}0.9303\,{\color{gray}\tiny$\pm$.0048}$^{\hphantom{\blacktriangle}}$ & \cellcolor{cellbest}\textbf{0.9338}\,{\color{gray}\tiny$\pm$.0027}$^{\hphantom{\blacktriangle}}$ & \cellcolor{cellnear}0.9311\,{\color{gray}\tiny$\pm$.0017}$^{\hphantom{\blacktriangle}}$ & \cellcolor{cellnear}0.9321\,{\color{gray}\tiny$\pm$.0015}$^{\hphantom{\blacktriangle}}$ & \cellcolor{cellbest}\textbf{0.9320}\,{\color{gray}\tiny$\pm$.0030}$^{\hphantom{\blacktriangle}}$ & \cellcolor{cellbest}\textbf{0.9348}\,{\color{gray}\tiny$\pm$.0039}$^{\blacktriangle}$ \\
 & TabPFN & \cellcolor{cellclose}0.9397\,{\color{gray}\tiny$\pm$.0003}$^{\hphantom{\blacktriangle}}$ & \cellcolor{cellnear}0.9402\,{\color{gray}\tiny$\pm$.0004}$^{\blacktriangle}$ & \cellcolor{cellnear}0.9435\,{\color{gray}\tiny$\pm$.0003}$^{\blacktriangle}$ & \cellcolor{cellnear}0.9412\,{\color{gray}\tiny$\pm$.0003}$^{\blacktriangle}$ & \cellcolor{cellbest}\textbf{0.9440}\,{\color{gray}\tiny$\pm$.0003}$^{\blacktriangle}$ & \cellcolor{cellnear}0.9423\,{\color{gray}\tiny$\pm$.0004}$^{\blacktriangle}$ \\
\midrule
\multicolumn{8}{l}{\textbf{Binary (accuracy)}} \\
churn & CatBoost & \cellcolor{cellclose}0.8608\,{\color{gray}\tiny$\pm$.0014}$^{\hphantom{\blacktriangle}}$ & \cellcolor{cellbest}\textbf{0.8639}\,{\color{gray}\tiny$\pm$.0019}$^{\blacktriangle}$ & \cellcolor{cellbest}\textbf{0.8640}\,{\color{gray}\tiny$\pm$.0027}$^{\blacktriangle}$ & \cellcolor{cellbest}\textbf{0.8652}\,{\color{gray}\tiny$\pm$.0025}$^{\blacktriangle}$ & \cellcolor{cellbest}\textbf{0.8651}\,{\color{gray}\tiny$\pm$.0013}$^{\blacktriangle}$ & \cellcolor{cellbest}\textbf{0.8655}\,{\color{gray}\tiny$\pm$.0022}$^{\blacktriangle}$ \\
 & TabM & \cellcolor{cellbest}\textbf{0.8619}\,{\color{gray}\tiny$\pm$.0020}$^{\hphantom{\blacktriangle}}$ & \cellcolor{cellbest}\textbf{0.8600}\,{\color{gray}\tiny$\pm$.0049}$^{\hphantom{\blacktriangle}}$ & \cellcolor{cellnear}0.8599\,{\color{gray}\tiny$\pm$.0032}$^{\blacktriangledown}$ & \cellcolor{cellbest}\textbf{0.8609}\,{\color{gray}\tiny$\pm$.0033}$^{\hphantom{\blacktriangle}}$ & \cellcolor{cellbest}\textbf{0.8620}\,{\color{gray}\tiny$\pm$.0023}$^{\hphantom{\blacktriangle}}$ & \cellcolor{cellbest}\textbf{0.8612}\,{\color{gray}\tiny$\pm$.0039}$^{\hphantom{\blacktriangle}}$ \\
 & TabPFN & \cellcolor{cellbest}\textbf{0.8658}\,{\color{gray}\tiny$\pm$.0006}$^{\hphantom{\blacktriangle}}$ & \cellcolor{cellbest}\textbf{0.8653}\,{\color{gray}\tiny$\pm$.0008}$^{\hphantom{\blacktriangle}}$ & \cellcolor{cellnear}0.8642\,{\color{gray}\tiny$\pm$.0008}$^{\blacktriangledown}$ & \cellcolor{cellnear}0.8640\,{\color{gray}\tiny$\pm$.0010}$^{\blacktriangledown}$ & \cellcolor{cellnear}0.8637\,{\color{gray}\tiny$\pm$.0008}$^{\blacktriangledown}$ & \cellcolor{cellnear}0.8633\,{\color{gray}\tiny$\pm$.0011}$^{\blacktriangledown}$ \\
adult & CatBoost & \cellcolor{cellnear}0.8740\,{\color{gray}\tiny$\pm$.0005}$^{\hphantom{\blacktriangle}}$ & \cellcolor{cellbest}\textbf{0.8749}\,{\color{gray}\tiny$\pm$.0008}$^{\blacktriangle}$ & \cellcolor{cellnear}0.8738\,{\color{gray}\tiny$\pm$.0006}$^{\hphantom{\blacktriangle}}$ & \cellcolor{cellnear}0.8739\,{\color{gray}\tiny$\pm$.0004}$^{\hphantom{\blacktriangle}}$ & \cellcolor{cellbest}\textbf{0.8744}\,{\color{gray}\tiny$\pm$.0004}$^{\blacktriangle}$ & \cellcolor{cellbest}\textbf{0.8746}\,{\color{gray}\tiny$\pm$.0003}$^{\blacktriangle}$ \\
 & TabM & \cellcolor{cellmid}0.8626\,{\color{gray}\tiny$\pm$.0010}$^{\hphantom{\blacktriangle}}$ & \cellcolor{cellbest}\textbf{0.8705}\,{\color{gray}\tiny$\pm$.0009}$^{\blacktriangle}$ & \cellcolor{cellmid}0.8630\,{\color{gray}\tiny$\pm$.0008}$^{\hphantom{\blacktriangle}}$ & \cellcolor{cellnear}0.8697\,{\color{gray}\tiny$\pm$.0013}$^{\blacktriangle}$ & \cellcolor{cellmid}0.8626\,{\color{gray}\tiny$\pm$.0010}$^{\hphantom{\blacktriangle}}$ & \cellcolor{cellbest}\textbf{0.8705}\,{\color{gray}\tiny$\pm$.0009}$^{\blacktriangle}$ \\
 & TabPFN & \cellcolor{cellmid}0.8623\,{\color{gray}\tiny$\pm$.0003}$^{\hphantom{\blacktriangle}}$ & \cellcolor{cellbest}\textbf{0.8719}\,{\color{gray}\tiny$\pm$.0004}$^{\blacktriangle}$ & \cellcolor{cellmid}0.8647\,{\color{gray}\tiny$\pm$.0005}$^{\blacktriangle}$ & \cellcolor{cellnear}0.8706\,{\color{gray}\tiny$\pm$.0007}$^{\blacktriangle}$ & \cellcolor{cellclose}0.8681\,{\color{gray}\tiny$\pm$.0005}$^{\blacktriangle}$ & \cellcolor{cellbest}\textbf{0.8719}\,{\color{gray}\tiny$\pm$.0005}$^{\blacktriangle}$ \\
higgs & CatBoost & \cellcolor{cellfar}0.7271\,{\color{gray}\tiny$\pm$.0010}$^{\hphantom{\blacktriangle}}$ & \cellcolor{cellmid}0.7295\,{\color{gray}\tiny$\pm$.0014}$^{\blacktriangle}$ & \cellcolor{cellclose}0.7343\,{\color{gray}\tiny$\pm$.0012}$^{\blacktriangle}$ & \cellcolor{cellclose}0.7349\,{\color{gray}\tiny$\pm$.0015}$^{\blacktriangle}$ & \cellcolor{cellnear}0.7371\,{\color{gray}\tiny$\pm$.0016}$^{\blacktriangle}$ & \cellcolor{cellbest}\textbf{0.7392}\,{\color{gray}\tiny$\pm$.0013}$^{\blacktriangle}$ \\
 & TabM & \cellcolor{cellfar}0.7334\,{\color{gray}\tiny$\pm$.0014}$^{\hphantom{\blacktriangle}}$ & \cellcolor{cellmid}0.7355\,{\color{gray}\tiny$\pm$.0008}$^{\blacktriangle}$ & \cellcolor{cellbest}\textbf{0.7438}\,{\color{gray}\tiny$\pm$.0013}$^{\blacktriangle}$ & \cellcolor{cellbest}\textbf{0.7437}\,{\color{gray}\tiny$\pm$.0014}$^{\blacktriangle}$ & \cellcolor{cellbest}\textbf{0.7440}\,{\color{gray}\tiny$\pm$.0015}$^{\blacktriangle}$ & \cellcolor{cellbest}\textbf{0.7447}\,{\color{gray}\tiny$\pm$.0013}$^{\blacktriangle}$ \\
 & TabPFN & \cellcolor{cellclose}0.7396\,{\color{gray}\tiny$\pm$.0005}$^{\hphantom{\blacktriangle}}$ & \cellcolor{cellmid}0.7358\,{\color{gray}\tiny$\pm$.0008}$^{\blacktriangledown}$ & \cellcolor{cellbest}\textbf{0.7428}\,{\color{gray}\tiny$\pm$.0009}$^{\blacktriangle}$ & \cellcolor{cellnear}0.7419\,{\color{gray}\tiny$\pm$.0008}$^{\blacktriangle}$ & \cellcolor{cellbest}\textbf{0.7436}\,{\color{gray}\tiny$\pm$.0011}$^{\blacktriangle}$ & \cellcolor{cellnear}0.7420\,{\color{gray}\tiny$\pm$.0008}$^{\blacktriangle}$ \\
\midrule
\multicolumn{8}{l}{\textbf{Multiclass (accuracy)}} \\
students & CatBoost & \cellcolor{cellclose}0.7649\,{\color{gray}\tiny$\pm$.0042}$^{\hphantom{\blacktriangle}}$ & \cellcolor{cellbest}\textbf{0.7695}\,{\color{gray}\tiny$\pm$.0041}$^{\blacktriangle}$ & \cellcolor{cellbest}\textbf{0.7693}\,{\color{gray}\tiny$\pm$.0030}$^{\blacktriangle}$ & \cellcolor{cellbest}\textbf{0.7703}\,{\color{gray}\tiny$\pm$.0036}$^{\blacktriangle}$ & \cellcolor{cellbest}\textbf{0.7716}\,{\color{gray}\tiny$\pm$.0036}$^{\blacktriangle}$ & \cellcolor{cellbest}\textbf{0.7688}\,{\color{gray}\tiny$\pm$.0035}$^{\hphantom{\blacktriangle}}$ \\
 & TabM & \cellcolor{cellbest}\textbf{0.7704}\,{\color{gray}\tiny$\pm$.0042}$^{\hphantom{\blacktriangle}}$ & \cellcolor{cellbest}\textbf{0.7723}\,{\color{gray}\tiny$\pm$.0051}$^{\hphantom{\blacktriangle}}$ & \cellcolor{cellclose}0.7685\,{\color{gray}\tiny$\pm$.0044}$^{\hphantom{\blacktriangle}}$ & \cellcolor{cellbest}\textbf{0.7732}\,{\color{gray}\tiny$\pm$.0062}$^{\hphantom{\blacktriangle}}$ & \cellcolor{cellbest}\textbf{0.7721}\,{\color{gray}\tiny$\pm$.0070}$^{\hphantom{\blacktriangle}}$ & \cellcolor{cellbest}\textbf{0.7743}\,{\color{gray}\tiny$\pm$.0055}$^{\hphantom{\blacktriangle}}$ \\
 & TabPFN & \cellcolor{cellclose}0.7732\,{\color{gray}\tiny$\pm$.0015}$^{\hphantom{\blacktriangle}}$ & \cellcolor{cellnear}0.7770\,{\color{gray}\tiny$\pm$.0017}$^{\blacktriangle}$ & \cellcolor{cellclose}0.7731\,{\color{gray}\tiny$\pm$.0018}$^{\hphantom{\blacktriangle}}$ & \cellcolor{cellnear}0.7767\,{\color{gray}\tiny$\pm$.0034}$^{\blacktriangle}$ & \cellcolor{cellbest}\textbf{0.7810}\,{\color{gray}\tiny$\pm$.0022}$^{\blacktriangle}$ & \cellcolor{cellnear}0.7780\,{\color{gray}\tiny$\pm$.0020}$^{\blacktriangle}$ \\
otto & CatBoost & \cellcolor{cellnear}0.8104\,{\color{gray}\tiny$\pm$.0006}$^{\hphantom{\blacktriangle}}$ & \cellcolor{cellnear}0.8118\,{\color{gray}\tiny$\pm$.0009}$^{\blacktriangle}$ & \cellcolor{cellbest}\textbf{0.8141}\,{\color{gray}\tiny$\pm$.0012}$^{\blacktriangle}$ & \cellcolor{cellbest}\textbf{0.8140}\,{\color{gray}\tiny$\pm$.0006}$^{\blacktriangle}$ & \cellcolor{cellnear}0.8130\,{\color{gray}\tiny$\pm$.0010}$^{\blacktriangle}$ & \cellcolor{cellbest}\textbf{0.8133}\,{\color{gray}\tiny$\pm$.0012}$^{\blacktriangle}$ \\
 & TabM & \cellcolor{cellbest}\textbf{0.8351}\,{\color{gray}\tiny$\pm$.0017}$^{\hphantom{\blacktriangle}}$ & \cellcolor{cellbest}\textbf{0.8351}\,{\color{gray}\tiny$\pm$.0013}$^{\hphantom{\blacktriangle}}$ & \cellcolor{cellbest}\textbf{0.8346}\,{\color{gray}\tiny$\pm$.0017}$^{\hphantom{\blacktriangle}}$ & \cellcolor{cellbest}\textbf{0.8347}\,{\color{gray}\tiny$\pm$.0013}$^{\hphantom{\blacktriangle}}$ & \cellcolor{cellbest}\textbf{0.8347}\,{\color{gray}\tiny$\pm$.0011}$^{\hphantom{\blacktriangle}}$ & \cellcolor{cellbest}\textbf{0.8352}\,{\color{gray}\tiny$\pm$.0014}$^{\hphantom{\blacktriangle}}$ \\
 & TabPFN & \cellcolor{cellnear}0.8382\,{\color{gray}\tiny$\pm$.0011}$^{\hphantom{\blacktriangle}}$ & \cellcolor{cellnear}0.8369\,{\color{gray}\tiny$\pm$.0015}$^{\hphantom{\blacktriangle}}$ & \cellcolor{cellbest}\textbf{0.8386}\,{\color{gray}\tiny$\pm$.0012}$^{\hphantom{\blacktriangle}}$ & \cellcolor{cellnear}0.8367\,{\color{gray}\tiny$\pm$.0008}$^{\blacktriangledown}$ & \cellcolor{cellbest}\textbf{0.8397}\,{\color{gray}\tiny$\pm$.0010}$^{\blacktriangle}$ & \cellcolor{cellnear}0.8375\,{\color{gray}\tiny$\pm$.0011}$^{\hphantom{\blacktriangle}}$ \\
\midrule
\multicolumn{2}{l}{\textit{rows above \rawarm{} ($\blacktriangle$)}} & {\color{gray}--}$^{\hphantom{\blacktriangle}}$ & 18$^{\hphantom{\blacktriangle}}$ & 15$^{\hphantom{\blacktriangle}}$ & 19$^{\hphantom{\blacktriangle}}$ & 19$^{\hphantom{\blacktriangle}}$ & 20$^{\hphantom{\blacktriangle}}$ \\
\multicolumn{2}{l}{\textit{rows in the best group}} & 5/30$^{\hphantom{\blacktriangle}}$ & 13/30$^{\hphantom{\blacktriangle}}$ & 11/30$^{\hphantom{\blacktriangle}}$ & 14/30$^{\hphantom{\blacktriangle}}$ & 19/30$^{\hphantom{\blacktriangle}}$ & 15/30$^{\hphantom{\blacktriangle}}$ \\
\bottomrule
\end{tabular}
\caption{Absolute test metrics (10 datasets $\times$ 3 estimators): R$^2$ for regression, accuracy for classification; higher is better. Cells: mean\,$\pm$\,std over 10 seeded re-scorings of the frozen selected program, each re-fitting transform and estimator (\S\ref{sec:setup}). \colorbox{cellbest}{\strut\textbf{Bold}}~= tied with the row's best arm; shading deepens with the cell's gap to it in tie widths $w$ (\S\ref{sec:results}), from \colorbox{cellnear}{\strut one} through \colorbox{cellclose}{\strut two}, \colorbox{cellmid}{\strut four} and \colorbox{cellfar}{\strut eight} to \colorbox{cellworst}{\strut beyond}. Superscripts place a cell above ($^{\blacktriangle}$) or below ($^{\blacktriangledown}$) \rawarm. $\cup$OF is a \emph{union}: the arm's selected program and \openfe{} are each fitted on the raw frame and their output columns concatenated.}
\label{tab:main}
\end{table*}

\paragraph{Datasets and estimators.}
We evaluate on 10 tabular datasets (sizes, tasks and metrics in the
supplementary material) crossed with three downstream estimators ---
\catboost{} \citep{prokhorenkova2018catboost} (gradient-boosted trees),
\tabm{} \citep{gorishniy2025tabm} (a parameter-efficient deep tabular
ensemble) and \tabpfn~v3
\citep{hollmann2025tabpfn,grinsztajn2026tabpfn3} (a prior-fitted transformer)
--- for 30 cells. A cell, one dataset with one estimator, is the unit of every
comparison. The datasets span 343--62{,}751 training rows and
5--93 raw columns; the score is R$^2$ for regression and accuracy for
classification, and every search optimizes it.

\paragraph{Arms.}
Table~\ref{tab:main} compares \emph{\rawarm} (untransformed columns),
\emph{\openfe} \citep{zhang2023openfe} with its full operator palette --- fit
once on the training split and applied inductively, not jointly with the test
frame --- \emph{\llmfe}
\citep{abhyankar2025llmfe} under a matched protocol, and \method. \llmfe{}
runs on the same backbone LLM as \method{} --- Gemini~3 Flash
\citep{googledeepmind2025gemini3flash}, snapshot
\texttt{gemini-3-flash-preview}, at high reasoning effort --- for the same
budget of 100 evaluated candidates per cell. Its released code scores every
candidate with a fixed four-fold XGBoost \citep{chen2016xgboost} surrogate
regardless of the downstream model; here the cell's own estimator and metric,
under the same validation protocol, guide search and program selection:
the baseline optimizes the objective it is evaluated on. Both baselines run their
authors' released code, pinned at \texttt{3e3bdf9} \citep{zhang2023openfecode} and
\texttt{e9a58ca} \citep{abhyankar2025llmfecode}; the estimators are the
released \catboost{} 1.2.10 \citep{catboost2018code}, \tabm{} 0.0.3
\citep{gorishniy2025tabmcode} and \tabpfn{} 8.1.0 with the official v3
checkpoints \citep{priorlabs2026tabpfncode}. The $\cup$OF arms
concatenate an arm's selected program features with \openfe's, quantifying
complementarity. The representation ablation
(cell-level scores in the supplementary material) swaps only what the
LLM rewrites --- raw Python source (\pythonarm), the whole graph JSON
(\jsonarm), or \method's schema-constrained diff --- with engine,
backbone LLM and budget held identical; the schema gate of \S\ref{sec:diff}
travels with the diff representation and is part of what the comparison
varies. Wall-clock was not matched across arms. Every
search, ablation and probe reported here cost approximately \$800 in LLM API
spend; the supplementary material breaks the search arms down per cell.

\paragraph{Protocol.}
No test value enters the search: fitness is development-side
cross-validation (\S\ref{sec:engine}). We select each cell's best program by
that score alone and freeze its structure and code before test evaluation.
The selected program is then scored on the
held-out test split as the mean\,$\pm$\,std of 10 seeded re-scorings, each
re-fitting transform and estimator. Within a cell, arms are compared by the
paired Wilcoxon signed-rank test over seeds;
claims aggregated across cells are Holm-corrected within families, and arms
compared jointly across cells by the Friedman
test \citep{demsar2006statistical}. Intervals are
bias-corrected accelerated (BCa) bootstrap. We
fixed claims R1--R4 of \S\ref{sec:results}, the families they are corrected in
and the checks each has to survive, before reading a single test score; R5's
representation comparisons came afterwards and are reported as exploratory.
Beyond the representation-level guards of \S\ref{sec:genome}, the final
selected programs of \emph{every} arm, baselines included, are checked by a
program-level label-leakage screen (\S\ref{sec:results}).

\section{Main Results}
\label{sec:results}

\begin{figure*}[t]
\centering
\includegraphics[width=\textwidth]{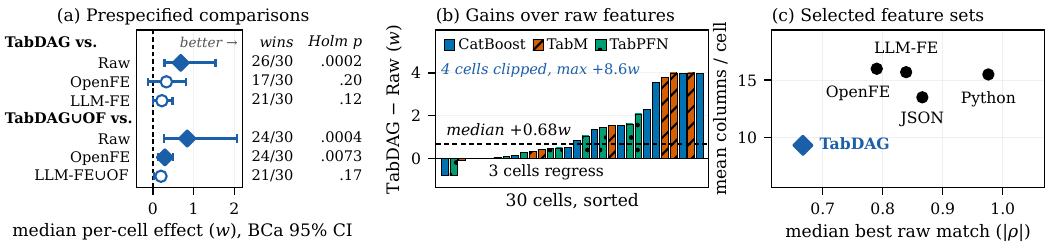}
\caption{The contrasts of \S\ref{sec:results}, on the same test scores as
Table~\ref{tab:main}. \textbf{(a)}~One claim per row, group heading versus the
row's arm, positive effects favoring the heading: median per-cell difference
in tie widths $w$, BCa 95\% interval, cells won, and Holm-corrected Wilcoxon
$p$ within its family; filled markers cleared all four checks set in advance.
\textbf{(b)}~Per-cell differences behind R1 (\method{} $-$ \rawarm), sorted,
colors and hatching distinguish estimators; four cells exceed the axis and are clipped.
\textbf{(c)}~What each arm's selected programs add, from a column-by-column replay of
all 30 cells: mean added columns per cell against the median best $|\rho|$
to any raw column; \method{} builds the smallest and least raw-redundant
sets.}
\label{fig:claims}
\end{figure*}

Differences below are read in tie widths. A tie width $w$ is the smallest gap
this experiment treats as a difference, fixed by how far a cell's score moves when
nothing but the seed changes: 0.0028 in binary accuracy, 0.0041 in R$^2$,
0.0048 in multiclass accuracy. Every
comparison runs over all thirty cells, and Fig.~\ref{fig:claims}a carries the
size, uncertainty and corrected $p$ of each claim fixed before a test score was
read.

\paragraph{R1. Evolved features improve on the raw feature set.}
\method{} scores above \rawarm{} on 26 of the 30 cells, and the typical cell
improves by about two thirds of a tie width; adding \openfe's features
alongside the
evolved ones raises that to about 0.85 tie widths. Neither result depends on one dataset --- both keep sign and significance
under leave-one-dataset-out and with the thirty cells collapsed to one value
per dataset --- nor on one fortunate run, as far as that can be tested without
repeating every search: redrawing every searched arm's scores ten thousand
times, with noise sized to the gap between two independent searches of the
same cell, leaves R1 standing in 99\% of the draws and its stacked counterpart
in 98\%. Those 26 count every cell that improves at all;
Table~\ref{tab:main} marks the stricter set, where the ten-seed gap also
survives the paired test: 19 of \method's 26 clear it and 20 of \methodof's
24, against 19 for \llmfeof, 18 for \openfe{} and 15 for \llmfe.

\paragraph{R2. The gains have a heavy tail, and the estimator sets their scale.}
Most cells move very little and a few move a great deal
(Fig.~\ref{fig:claims}b): against the typical two thirds of a tie width, the
four largest
gain five to nine, and three of the thirty cells finish below \rawarm. The estimator matters
more than the pooled median suggests: the median cell gains about
$1.5w$ under \catboost, $0.5w$ under \tabpfn{} and $0.4w$ under \tabm.

\paragraph{R3. Evolved features add to \openfe{} rather than duplicating it.}
Giving the estimator \openfe's features and the evolved ones together beats
\openfe{} alone on 24 of the 30 cells, by about three tenths of a tie width in
the typical cell --- a margin that holds up under correction ($p=.007$).
The margin is smaller than R1's --- about as large as the movement a
rerun of the search produces --- so the same redraw test blurs it: R3
survives three draws in ten, the outcome expected when an effect is about
the same size as the noise it is tested with.

\paragraph{R4. \method{} leads the ordering.} Ranking the six arms within
each cell orders them \methodof, \method, \llmfeof, \llmfe, \openfe, \rawarm,
and the arms are not interchangeable ($n=30$, Friedman $p=1\times10^{-4}$).
\method{} holds the top two places and the largest best-group count in
Table~\ref{tab:main} --- 19 of 30 rows, where no baseline arm exceeds 14. The typical cell puts \method{} ahead of both
\openfe{} and \llmfe{} by a fifth to a third of a tie width, and both
contrasts against \llmfe{} carry bootstrap intervals clear of zero
(Fig.~\ref{fig:claims}a); margins this size sit at the edge of what ten
datasets resolve, and the pairwise rank tests do not individually clear
correction. The experiment settles the gap between raw features and the
engineered arms, and the one contrast among the engineered arms it resolves
after correction --- the union against the \openfe{} it builds on (R3) ---
falls to \method.

\paragraph{R5. Constrained edits have the best mean rank and fewer failed
evaluations.} With engine, LLM and budget held fixed,
\method{} takes the best mean rank of the three encodings and the most rows in
the best group ---
18 of 30, against 17 for the \pythonarm{} rewrite and 15 for the whole-graph
\jsonarm{} one. Paired within cell it leads both, by about a quarter of a tie
width over \pythonarm{} and an eighth over \jsonarm{}, margins the discrete
accuracy cells dilute; on the fifteen regression cells,
where the fitness signal is continuous, the margin over \jsonarm{} widens to
$0.78w$ with 12 of 15 cells improving ($p=.015$).
In this ablation, whole-graph \jsonarm{} rewriting does not improve on
\pythonarm{} rewriting, whereas the constrained diff has the best mean
rank. The comparison supports the combination of structured edits and
pre-evaluation validation, rather than graph encoding alone.
We did not fix these three comparisons in advance, so we read
them as exploratory. What the diff
produces also differs in kind: 11\% of its selected program columns are monotone
recodings of a single raw column, against 17\% for the \jsonarm{} rewrite and
30\% for the \pythonarm{} one, and on the inspectable failure subsets the
own-target guard of \S\ref{sec:genome} rejected 16 of its candidates against 59
and 89 for the two rewrites. The failure counts in \S\ref{sec:diff}
separate rejected proposals from structural and runtime failures:
\method{} loses 20.1\% of evaluations, versus 27.1\% for \jsonarm{} and
27.9\% for \pythonarm. On the \texttt{students} dataset the
\pythonarm{} selected programs were re-selected under the program-level leakage
screen of \S\ref{sec:setup}; no other cell in the ablation was affected.

\section{Analysis}
\label{sec:analysis}

\paragraph{The arms build different kinds of columns.}
\openfe{} produces arithmetic combinations of raw-column pairs in quantity:
about sixteen per cell, each close to a single raw column and unnamed by
construction. The LLM arms produce named,
dataset-referring transformations instead: under a blind labeling of every
selected program column against a closed taxonomy, 85\% of \method's columns were
judged domain-specific against 5\% of \openfe's, and 36\% were multi-step
composites, functions of previously generated columns. \method{} also builds the smallest feature sets in the study, a median
of seven added columns per selected program (Fig.~\ref{fig:claims}c). Depth does not mean information the
estimator could not recover: a histogram-based gradient-boosting probe
\citep{pedregosa2011sklearn} rebuilds almost every engineered column from the
raw frame for all arms alike, and \openfe's columns take a larger share of
feature importance where \method{} concentrates it in one or two.
Fewer, deeper columns are a different set, not a uniformly stronger
one; the union arms of Table~\ref{tab:main} draw on that complementarity.

\paragraph{The two LLM arms build different feature sets.}
At the feature level the informative comparison is with \llmfe: the same
backbone, budget and program-selection rule, but features written as
free-standing assignment lines instead of a typed graph. For each added
column we record its strongest correlation with any single raw column:
$|\rho|$ near one means the column recodes an existing one. Free-form code
leans that way: \llmfe{} adds about sixteen columns per cell, the typical
one at $|\rho|=0.84$. The graph arm builds fewer and
reaches further --- nine columns per cell at $|\rho|=0.67$, combinations of
the frame rather than variations on it (Fig.~\ref{fig:claims}c). Cell by
cell, \method{} adds fewer columns in 24 of 30 cells (Wilcoxon $p<.001$),
and its columns are the less raw-correlated ones in 16 of the 26 cells
where both arms add any, a majority short of significance ($p=.22$). The
arms differ in their search loop as well as their encoding, so the contrast
belongs to the pair; R5 isolates the encoding.

\paragraph{A large part of the gain does not require search.}
The rival explanation is that the model already knows the answer and search
only collects it. Prompted once with the same task description and no
evaluation feedback, the backbone LLM proposes feature sets recovering a median
79\% of the searched gain when the best of 16 samples is kept, and matching or
exceeding the program selected by search on three of the ten datasets.
The prior is not uniformly reliable --- a single unselected proposal falls
below the identity baseline on half the datasets --- and best-of-16 is a
diagnostic, not a selection rule. With that caveat, end-of-search numbers,
ours included, credit search with gain already present in the
backbone's first proposals. The probe, scored on validation under
\catboost{}, is tabulated in the supplementary material alongside a semantic
control: with column names anonymized or shuffled, the search still improves
on raw features, though correctly aligned names reach a higher frontier
sooner.

\paragraph{Selected programs are inspectable.}
Declared inputs, outputs, and dependencies expose how features are computed.
The supplementary material reproduces every cell's selected program,
including the baselines.

\paragraph{The estimator changes the ranking.}
Ranked within one estimator, Table~\ref{tab:main}'s six arms are
distinguishable under \catboost{} (Friedman $p=.004$) and \tabm{}
($p=.002$) but not \tabpfn{} ($p=.14$), and the orderings differ in kind: raw features
come last under \catboost{} and \tabm,
whereas under \tabpfn{} they sit mid-table, ahead of \openfe{} and \llmfeof,
with plain \method{} still first. \tabpfn{} appears to carry in its prior
much of what the trees gain from engineering, bounding how far a feature set
selected against one estimator transfers.

\begin{table}[t]
\centering
\scriptsize
\setlength{\tabcolsep}{4.5pt}
\renewcommand{\arraystretch}{0.98}
\begin{tabular}{@{}lrrrrrrr@{}}
\toprule
selector & \rotatebox{60}{\catboost} & \rotatebox{60}{XGBoost} &
\rotatebox{60}{TabICL} & \rotatebox{60}{\tabpfn} &
\rotatebox{60}{RealMLP} & \rotatebox{60}{\tabm} & \rotatebox{60}{as source} \\
\midrule
\catboost &
\cellcolor{black!8}\textit{self} &
\cellcolor{cellfar}3 &
\cellcolor{cellmid}4 &
\cellcolor{cellmid}4 &
\cellcolor{cellnear}5 &
\cellcolor{cellmid}4 & 20 \\
XGBoost &
\cellcolor{cellfar}2 &
\cellcolor{black!8}\textit{self} &
\cellcolor{cellmid}4 &
\cellcolor{cellmid}4 &
\cellcolor{cellbest}7 &
\cellcolor{cellmid}4 & 21 \\
TabICL &
\cellcolor{cellfar}2 &
\cellcolor{cellfar}2 &
\cellcolor{black!8}\textit{self} &
\cellcolor{cellfar}2 &
\cellcolor{cellbest}6 &
\cellcolor{cellfar}2 & 14 \\
\tabpfn &
\cellcolor{cellworst}0 &
\cellcolor{cellfar}2 &
\cellcolor{cellbest}6 &
\cellcolor{black!8}\textit{self} &
\cellcolor{cellmid}4 &
\cellcolor{cellfar}2 & 14 \\
RealMLP &
\cellcolor{cellworst}0 &
\cellcolor{cellworst}0 &
\cellcolor{cellworst}1 &
\cellcolor{cellworst}1 &
\cellcolor{black!8}\textit{self} &
\cellcolor{cellworst}1 & 3 \\
\tabm &
\cellcolor{cellfar}2 &
\cellcolor{cellmid}4 &
\cellcolor{cellfar}3 &
\cellcolor{cellfar}3 &
\cellcolor{cellnear}5 &
\cellcolor{black!8}\textit{self} & 17 \\
\midrule
against & 6 & 11 & 18 & 14 & 27 & 13 & 89 \\
\bottomrule
\end{tabular}
\caption{Can a graph be borrowed? Each cell counts the datasets, of eight,
where the five-best mean held-out score of programs selected against the row
estimator beats the receiving column's own, over three paired evaluator
refits; no cell tied. A zero denotes no wins over the receiver's own
selection, not a failure to execute the transferred graphs.
Margins: each estimator's wins as source and foreign
wins against it, of forty each. Colors: \colorbox{cellbest}{6--8},
\colorbox{cellnear}{5}, \colorbox{cellmid}{4}, \colorbox{cellfar}{2--3},
\colorbox{cellworst}{0--1} wins.}
\label{tab:transfer}
\end{table}

\paragraph{Can a graph be borrowed?}
A transfer campaign, detailed in the supplementary material, ran two
independent \method{} searches for each of eight datasets crossed with six
estimators --- the experiment's three plus a second member of each family
(XGBoost \citep{chen2016xgboost}, RealMLP \citep{holzmuller2024realmlp},
TabICL \citep{qu2025tabicl}) --- and
re-scored all 96 frozen selected programs under every estimator: a graph selected by
a different model beats the receiver's own in 164 of 480 dataset-level
comparisons and is the best source in 66 of 96 receiver columns,
yet most selector--receiver pairs win on fewer than half the datasets. A
finer replay of one search per cell froze its five best programs by
development fitness, averaging held-out scores over three evaluator
refits; Table~\ref{tab:transfer} counts, per selector--receiver pair, the
datasets where the foreign five-program mean beats the receiver's own.
A zero means that foreign selection never gives a higher mean score on
these eight datasets; it does not mean that the graphs cannot be executed
with the receiving model. Conversely, it does not establish equivalent
performance or improvement over raw features: this table counts strict wins
against the receiver's own selected programs, not effect sizes or wins
against Raw.

Foreign selection wins 89 of the 240 off-diagonal cells, with clear
directional differences: graphs selected
against RealMLP win 3 of their 40 transfers, yet foreign selections
beat RealMLP's own in 27 of 40.
\catboost{} is the opposite pole, its own selection beaten in only 6 of 40,
and single pairs span 7 of 8 (XGBoost to RealMLP) to 0 of 8 (\tabpfn{} to
\catboost). The same replay puts the rank-one selected program above its own
search's five-program mean in 186 of 288 cells, from 40 of 48 under RealMLP
to 25 of 48 under \tabm: the development ranking carries signal at rank one;
averaging the top five usually lowers the score. These results show that
graphs can be reused across estimators, but that reuse does not reliably
outperform selection for the receiving estimator.

\paragraph{Repeated-search checks.}
Repeated searches on a subset of the primary experiment give a median
test-score gap of 0.0024, below one tie width. In three quarters of the
cells, the paired seeded evaluations do not detect a difference. These
observed gaps determine the search-sized noise used in R1 and R3's
sensitivity analyses. Repeating the transfer experiment in all 48 cells,
with one auxiliary framework setting changed, gives a median gap of 0.002
and a single 0.065 R$^2$ outlier, without a systematic effect of that
setting. These checks suggest similar performance across many observed
runs, but do not replace controlled replication over independent search seeds.

\paragraph{A limited backbone check.}
Re-running the \method{} search under a newer backbone (Gemini~3.5 Flash) on
four \catboost{} cells --- one search per cell, scored like the experiment over
10 seeds --- gives a consistent picture: three selected programs land within
one seed standard deviation of the reported ones (\texttt{airfoil} 0.9739
vs.\ 0.9736, \texttt{concrete} 0.9400 vs.\ 0.9397, \texttt{students} 0.7721
vs.\ 0.7716), and in the fourth a single search stops at raw-feature level
rather than below it (\texttt{churn} 0.8605, raw 0.8608, reported 0.8651).
This four-cell check is consistent with applicability to a second backbone,
but is too small to establish backbone-independent performance.

\section{Limitations}
\label{sec:limitations}

The main limitation is the evaluation scope: ten datasets and three
estimators, with one primary search per method and dataset--estimator
combination. Ten seeded re-scorings refit the selected program and estimator,
not the search. Intervals and $p$-values therefore describe fitting
variability conditional on those programs. Repeated-search checks
(\S\ref{sec:analysis}) and noise-based sensitivity analyses (R1, R3)
provide supporting evidence, but are not controlled search-seed replication:
the repeated runs also differed in an auxiliary setting.

The results support gains over raw features and complementarity with
\openfe{}, but not a reliable pairwise ranking of all feature-engineering
methods after multiple-testing correction. Broader datasets, independent
search seeds, and additional backbones are needed. Wall-clock and API
spend were not matched, so fewer failed
evaluations do not establish lower end-to-end cost. Possible memorization
of public datasets by the proposing LLM \citep{bordt2024memorization}
also remains unresolved; the search-free control measures existing useful
proposals but does not rule out contamination.

\section{Conclusion}
\label{sec:conclusion}

TabDAG combines typed feature DAGs with schema-constrained edits and
pre-evaluation checks. Across ten datasets and three estimator families,
evolved features improve on raw inputs and complement OpenFE. TabDAG and
its union achieve the two best mean ranks, but pairwise rankings among
feature-engineering methods remain uncertain. Constrained edits fail fewer
evaluations than Python or whole-graph JSON rewrites, without establishing
lower API cost. The search-free control finds much of the gain in the
LLM's initial proposals. Explicit graphs support inspection of the
selected features and their dependencies.

\bibliography{refs}

\begin{thebibliography}{48}
\providecommand{\natexlab}[1]{#1}

\bibitem[{Abhyankar, Shojaee, and Reddy(2025)}]{abhyankar2025llmfecode}
Abhyankar, N.; Shojaee, P.; and Reddy, C.~K. 2025.
\newblock {LLM-FE}: Reference Implementation.
\newblock \url{https://github.com/nikhilsab/LLMFE}.

\bibitem[{Abhyankar, Shojaee, and Reddy(2026)}]{abhyankar2025llmfe}
Abhyankar, N.; Shojaee, P.; and Reddy, C.~K. 2026.
\newblock {LLM-FE}: Automated Feature Engineering for Tabular Data with {LLMs}
  as Evolutionary Optimizers.
\newblock \emph{Transactions on Machine Learning Research}.
\newblock ArXiv:2503.14434.

\bibitem[{Baldi, Sadowski, and Whiteson(2014)}]{baldi2014higgs}
Baldi, P.; Sadowski, P.; and Whiteson, D. 2014.
\newblock Searching for exotic particles in high-energy physics with deep
  learning.
\newblock \emph{Nature Communications}, 5: 4308.

\bibitem[{Bordt et~al.(2024)Bordt, Nori, Rodrigues, Nushi, and
  Caruana}]{bordt2024memorization}
Bordt, S.; Nori, H.; Rodrigues, V.; Nushi, B.; and Caruana, R. 2024.
\newblock Elephants Never Forget: Memorization and Learning of Tabular Data in
  Large Language Models.
\newblock In \emph{Conference on Language Modeling (COLM)}.

\bibitem[{Bourassa(2021)}]{bourassa2021miami}
Bourassa, S.~C. 2021.
\newblock MiamiHousing2016.
\newblock OpenML dataset 43093.

\bibitem[{Brooks, Pope, and Marcolini(1989)}]{brooks1989airfoil}
Brooks, T.~F.; Pope, D.~S.; and Marcolini, M.~A. 1989.
\newblock Airfoil Self-Noise and Prediction.
\newblock Technical Report NASA-RP-1218, NASA.

\bibitem[{Chen and Guestrin(2016)}]{chen2016xgboost}
Chen, T.; and Guestrin, C. 2016.
\newblock {XGBoost}: A Scalable Tree Boosting System.
\newblock In \emph{Proceedings of the 22nd ACM SIGKDD International Conference
  on Knowledge Discovery and Data Mining (KDD)}, 785--794.

\bibitem[{Chen et~al.(2019)Chen, Lin, Luo, Li, Zhang, Xu, Dang, Sui, Zhang,
  Qiao, Zhang, Wu, Chintalapati, and Zhang}]{chen2019nfs}
Chen, X.; Lin, Q.; Luo, C.; Li, X.; Zhang, H.; Xu, Y.; Dang, Y.; Sui, K.;
  Zhang, X.; Qiao, B.; Zhang, W.; Wu, W.; Chintalapati, M.; and Zhang, D. 2019.
\newblock Neural Feature Search: A Neural Architecture for Automated Feature
  Engineering.
\newblock In \emph{IEEE International Conference on Data Mining (ICDM)},
  71--80.

\bibitem[{Dem{\v{s}}ar(2006)}]{demsar2006statistical}
Dem{\v{s}}ar, J. 2006.
\newblock Statistical Comparisons of Classifiers over Multiple Data Sets.
\newblock \emph{Journal of Machine Learning Research}, 7: 1--30.

\bibitem[{Dong et~al.(2026)Dong, Zheng, Han, Chen, Ruan, Xu, Chen, and
  Chen}]{dong2026malmas}
Dong, F.; Zheng, Z.; Han, X.; Chen, W.; Ruan, J.; Xu, T.; Chen, Y.; and Chen,
  E. 2026.
\newblock Memory-Augmented {LLM}-based Multi-Agent System for Automated Feature
  Generation on Tabular Data.
\newblock In \emph{Proceedings of the 64th Annual Meeting of the Association
  for Computational Linguistics (Volume 1: Long Papers)}, 10988--11007.

\bibitem[{Geng et~al.(2023)Geng, Josifoski, Peyrard, and
  West}]{geng2023grammar}
Geng, S.; Josifoski, M.; Peyrard, M.; and West, R. 2023.
\newblock Grammar-Constrained Decoding for Structured {NLP} Tasks without
  Finetuning.
\newblock In \emph{Proceedings of the 2023 Conference on Empirical Methods in
  Natural Language Processing}, 10932--10952.

\bibitem[{{Google DeepMind}(2025)}]{googledeepmind2025gemini3flash}
{Google DeepMind}. 2025.
\newblock Gemini 3 Flash Model Card.
\newblock
  \url{https://storage.googleapis.com/deepmind-media/Model-Cards/Gemini-3-Flash-Model-Card.pdf}.
\newblock Published December 2025.

\bibitem[{Gorishniy, Kotelnikov, and
  Babenko(2025{\natexlab{a}})}]{gorishniy2025tabm}
Gorishniy, Y.; Kotelnikov, A.; and Babenko, A. 2025{\natexlab{a}}.
\newblock {TabM}: Advancing Tabular Deep Learning with Parameter-Efficient
  Ensembling.
\newblock In \emph{International Conference on Learning Representations
  (ICLR)}.

\bibitem[{Gorishniy, Kotelnikov, and
  Babenko(2025{\natexlab{b}})}]{gorishniy2025tabmcode}
Gorishniy, Y.; Kotelnikov, A.; and Babenko, A. 2025{\natexlab{b}}.
\newblock {TabM}: Reference Implementation, Version 0.0.3.
\newblock \url{https://github.com/yandex-research/tabm}.

\bibitem[{Grinsztajn et~al.(2026)Grinsztajn, Fl{\"o}ge, Key, Birkel, Jund
  et~al.}]{grinsztajn2026tabpfn3}
Grinsztajn, L.; Fl{\"o}ge, K.; Key, O.; Birkel, F.; Jund, P.; et~al. 2026.
\newblock {TabPFN-3}: Technical Report.
\newblock \url{https://arxiv.org/abs/2605.13986}.
\newblock arXiv:2605.13986.

\bibitem[{Grinsztajn, Oyallon, and Varoquaux(2022)}]{grinsztajn2022why}
Grinsztajn, L.; Oyallon, E.; and Varoquaux, G. 2022.
\newblock Why Do Tree-Based Models Still Outperform Deep Learning on Typical
  Tabular Data?
\newblock In \emph{Advances in Neural Information Processing Systems (NeurIPS),
  Datasets and Benchmarks Track}.

\bibitem[{Han et~al.(2024)Han, Yoon, Ar{\i}k, and Pfister}]{han2024featllm}
Han, S.; Yoon, J.; Ar{\i}k, S.~{\"O}.; and Pfister, T. 2024.
\newblock Large Language Models Can Automatically Engineer Features for
  Few-Shot Tabular Learning.
\newblock In \emph{Proceedings of the 41st International Conference on Machine
  Learning (ICML)}.

\bibitem[{Hollmann, M{\"u}ller, and Hutter(2023)}]{hollmann2023caafe}
Hollmann, N.; M{\"u}ller, S.; and Hutter, F. 2023.
\newblock Large Language Models for Automated Data Science: Introducing {CAAFE}
  for Context-Aware Automated Feature Engineering.
\newblock In \emph{Advances in Neural Information Processing Systems
  (NeurIPS)}.

\bibitem[{Hollmann et~al.(2025)Hollmann, M{\"u}ller, Purucker, Krishnakumar,
  K{\"o}rfer, Hoo, Schirrmeister, and Hutter}]{hollmann2025tabpfn}
Hollmann, N.; M{\"u}ller, S.; Purucker, L.; Krishnakumar, A.; K{\"o}rfer, M.;
  Hoo, S.~B.; Schirrmeister, R.~T.; and Hutter, F. 2025.
\newblock Accurate Predictions on Small Data with a Tabular Foundation Model.
\newblock \emph{Nature}, 637: 319--326.

\bibitem[{Holzm{\"u}ller, Grinsztajn, and
  Steinwart(2024)}]{holzmuller2024realmlp}
Holzm{\"u}ller, D.; Grinsztajn, L.; and Steinwart, I. 2024.
\newblock Better by Default: Strong Pre-Tuned {MLPs} and Boosted Trees on
  Tabular Data.
\newblock In \emph{Advances in Neural Information Processing Systems 37
  (NeurIPS)}.

\bibitem[{Kaggle({\natexlab{b}})}]{kaggle2015otto}
Kaggle. 2015{\natexlab{b}}.
\newblock Otto Group Product Classification Challenge.
\newblock Kaggle competition.

\bibitem[{Kaggle({\natexlab{a}})}]{kaggle2019churn}
Kaggle. 2019{\natexlab{a}}.
\newblock Churn Modelling.
\newblock Kaggle dataset.

\bibitem[{Kanter and Veeramachaneni(2015)}]{kanter2015dfs}
Kanter, J.~M.; and Veeramachaneni, K. 2015.
\newblock Deep Feature Synthesis: Towards Automating Data Science Endeavors.
\newblock In \emph{IEEE International Conference on Data Science and Advanced
  Analytics (DSAA)}.

\bibitem[{Katz, Shin, and Song(2016)}]{katz2016explorekit}
Katz, G.; Shin, E. C.~R.; and Song, D. 2016.
\newblock {ExploreKit}: Automatic Feature Generation and Selection.
\newblock In \emph{IEEE International Conference on Data Mining (ICDM)},
  979--984.

\bibitem[{Khurana, Samulowitz, and Turaga(2018)}]{khurana2018transgraph}
Khurana, U.; Samulowitz, H.; and Turaga, D. 2018.
\newblock Feature Engineering for Predictive Modeling Using Reinforcement
  Learning.
\newblock In \emph{Proceedings of the AAAI Conference on Artificial
  Intelligence}, volume~32, 3407--3414.

\bibitem[{Kohavi(1996)}]{kohavi1996adult}
Kohavi, R. 1996.
\newblock Scaling Up the Accuracy of Naive-Bayes Classifiers: A Decision-Tree
  Hybrid.
\newblock In \emph{Proceedings of the Second International Conference on
  Knowledge Discovery and Data Mining (KDD)}, 202--207.

\bibitem[{K{\"u}ken, Purucker, and Hutter(2024)}]{kueken2024operatorbias}
K{\"u}ken, J.; Purucker, L.; and Hutter, F. 2024.
\newblock Large Language Models Engineer Too Many Simple Features for Tabular
  Data.
\newblock In \emph{Table Representation Learning Workshop at NeurIPS}.
\newblock ArXiv:2410.17787.

\bibitem[{Lehman et~al.(2022)Lehman, Gordon, Jain, Ndousse, Yeh, and
  Stanley}]{lehman2022elm}
Lehman, J.; Gordon, J.; Jain, S.; Ndousse, K.; Yeh, C.; and Stanley, K.~O.
  2022.
\newblock Evolution through Large Models.
\newblock \emph{arXiv preprint arXiv:2206.08896}.

\bibitem[{Liu et~al.(2024)Liu, Tong, Yuan, Lin, Luo, Wang, Lu, and
  Zhang}]{liu2024eoh}
Liu, F.; Tong, X.; Yuan, M.; Lin, X.; Luo, F.; Wang, Z.; Lu, Z.; and Zhang, Q.
  2024.
\newblock Evolution of Heuristics: Towards Efficient Automatic Algorithm Design
  Using Large Language Model.
\newblock In \emph{Proceedings of the 41st International Conference on Machine
  Learning (ICML)}.

\bibitem[{McKinney(2010)}]{mckinney2010pandas}
McKinney, W. 2010.
\newblock Data Structures for Statistical Computing in {P}ython.
\newblock In \emph{Proceedings of the 9th Python in Science Conference
  (SciPy)}, 56--61.

\bibitem[{Mouret and Clune(2015)}]{mouret2015mapelites}
Mouret, J.-B.; and Clune, J. 2015.
\newblock Illuminating Search Spaces by Mapping Elites.
\newblock \emph{arXiv preprint arXiv:1504.04909}.

\bibitem[{Nam et~al.(2024)Nam, Kim, Oh, Tack, Kim, and Shin}]{nam2024octree}
Nam, J.; Kim, K.; Oh, S.; Tack, J.; Kim, J.; and Shin, J. 2024.
\newblock Optimized Feature Generation for Tabular Data via {LLMs} with
  Decision Tree Reasoning.
\newblock In \emph{Advances in Neural Information Processing Systems
  (NeurIPS)}.

\bibitem[{Nargesian et~al.(2017)Nargesian, Samulowitz, Khurana, Khalil, and
  Turaga}]{nargesian2017lfe}
Nargesian, F.; Samulowitz, H.; Khurana, U.; Khalil, E.~B.; and Turaga, D.~S.
  2017.
\newblock Learning Feature Engineering for Classification.
\newblock In \emph{Proceedings of the 26th International Joint Conference on
  Artificial Intelligence (IJCAI)}, 2529--2535.

\bibitem[{Novikov et~al.(2025)Novikov, V{\~u}, Eisenberger, Dupont, Huang,
  Wagner, Shirobokov, Kozlovskii, Ruiz, Mehrabian, Kumar, See, Chaudhuri,
  Holland, Davies, Nowozin, Kohli, and Balog}]{novikov2025alphaevolve}
Novikov, A.; V{\~u}, N.; Eisenberger, M.; Dupont, E.; Huang, P.-S.; Wagner,
  A.~Z.; Shirobokov, S.; Kozlovskii, B.; Ruiz, F. J.~R.; Mehrabian, A.; Kumar,
  M.~P.; See, A.; Chaudhuri, S.; Holland, G.; Davies, A.; Nowozin, S.; Kohli,
  P.; and Balog, M. 2025.
\newblock {AlphaEvolve}: A Coding Agent for Scientific and Algorithmic
  Discovery.
\newblock \emph{arXiv preprint arXiv:2506.13131}.

\bibitem[{OpenML()}]{openml_house16h}
OpenML. 2014.
\newblock house\_16H.
\newblock OpenML dataset 574.
\newblock Originates from the DELVE repository; based on data from the 1990
  U.S. Census.

\bibitem[{Pedregosa et~al.(2011)Pedregosa, Varoquaux, Gramfort, Michel,
  Thirion, Grisel, Blondel, Prettenhofer, Weiss, Dubourg, Vanderplas, Passos,
  Cournapeau, Brucher, Perrot, and Duchesnay}]{pedregosa2011sklearn}
Pedregosa, F.; Varoquaux, G.; Gramfort, A.; Michel, V.; Thirion, B.; Grisel,
  O.; Blondel, M.; Prettenhofer, P.; Weiss, R.; Dubourg, V.; Vanderplas, J.;
  Passos, A.; Cournapeau, D.; Brucher, M.; Perrot, M.; and Duchesnay, {\'E}.
  2011.
\newblock Scikit-learn: Machine Learning in {P}ython.
\newblock \emph{Journal of Machine Learning Research}, 12: 2825--2830.

\bibitem[{{Prior Labs}(2026)}]{priorlabs2026tabpfncode}
{Prior Labs}. 2026.
\newblock {TabPFN}: Reference Implementation, Version 8.1.0, and {TabPFN-3}
  Model Weights.
\newblock \url{https://github.com/PriorLabs/TabPFN}, weights at
  \url{https://huggingface.co/Prior-Labs/tabpfn_3}.

\bibitem[{Prokhorenkova et~al.(2018)Prokhorenkova, Gusev, Vorobev, Dorogush,
  and Gulin}]{prokhorenkova2018catboost}
Prokhorenkova, L.; Gusev, G.; Vorobev, A.; Dorogush, A.~V.; and Gulin, A. 2018.
\newblock {CatBoost}: Unbiased Boosting with Categorical Features.
\newblock In \emph{Advances in Neural Information Processing Systems
  (NeurIPS)}.

\bibitem[{Qu et~al.(2025)Qu, Holzm{\"u}ller, Varoquaux, and
  Le~Morvan}]{qu2025tabicl}
Qu, J.; Holzm{\"u}ller, D.; Varoquaux, G.; and Le~Morvan, M. 2025.
\newblock {TabICL}: A Tabular Foundation Model for In-Context Learning on Large
  Data.
\newblock In \emph{Proceedings of the 42nd International Conference on Machine
  Learning (ICML)}.

\bibitem[{Rana(2013)}]{rana2013protein}
Rana, P. 2013.
\newblock Physicochemical Properties of Protein Tertiary Structure.
\newblock UCI Machine Learning Repository.

\bibitem[{Realinho et~al.(2022)Realinho, Machado, Baptista, and
  Martins}]{realinho2022students}
Realinho, V.; Machado, J.; Baptista, L.; and Martins, M.~V. 2022.
\newblock Predicting Student Dropout and Academic Success.
\newblock \emph{Data}, 7(11): 146.

\bibitem[{Romera-Paredes et~al.(2024)Romera-Paredes, Barekatain, Novikov,
  Balog, Kumar, Dupont, Ruiz, Ellenberg, Wang, Fawzi, Kohli, and
  Fawzi}]{romeraparedes2024funsearch}
Romera-Paredes, B.; Barekatain, M.; Novikov, A.; Balog, M.; Kumar, M.~P.;
  Dupont, E.; Ruiz, F. J.~R.; Ellenberg, J.~S.; Wang, P.; Fawzi, O.; Kohli, P.;
  and Fawzi, A. 2024.
\newblock Mathematical Discoveries from Program Search with Large Language
  Models.
\newblock \emph{Nature}, 625: 468--475.

\bibitem[{Willard and Louf(2023)}]{willard2023efficient}
Willard, B.~T.; and Louf, R. 2023.
\newblock Efficient Guided Generation for Large Language Models.
\newblock \emph{arXiv preprint arXiv:2307.09702}.

\bibitem[{{Yandex}(2018)}]{catboost2018code}
{Yandex}. 2018.
\newblock {CatBoost}: Reference Implementation, Version 1.2.10.
\newblock \url{https://github.com/catboost/catboost}.

\bibitem[{Yeh(1998)}]{yeh1998concrete}
Yeh, I.-C. 1998.
\newblock Modeling of strength of high-performance concrete using artificial
  neural networks.
\newblock \emph{Cement and Concrete Research}, 28(12): 1797--1808.

\bibitem[{Zhang et~al.(2023{\natexlab{a}})Zhang, Zhang, Fan, Luo, Liu, Liu,
  Cao, and Li}]{zhang2023openfe}
Zhang, T.; Zhang, Z.; Fan, Z.; Luo, H.; Liu, F.; Liu, Q.; Cao, W.; and Li, J.
  2023{\natexlab{a}}.
\newblock {OpenFE}: Automated Feature Generation with Expert-level Performance.
\newblock In \emph{Proceedings of the 40th International Conference on Machine
  Learning (ICML)}.

\bibitem[{Zhang et~al.(2023{\natexlab{b}})Zhang, Zhang, Fan, Luo, Liu, Liu,
  Cao, and Li}]{zhang2023openfecode}
Zhang, T.; Zhang, Z.; Fan, Z.; Luo, H.; Liu, F.; Liu, Q.; Cao, W.; and Li, J.
  2023{\natexlab{b}}.
\newblock {OpenFE}: Reference Implementation.
\newblock \url{https://github.com/IIIS-Li-Group/OpenFE}.

\bibitem[{Zhu et~al.(2022)Zhu, Xu, Yuan, and Huang}]{zhu2022difer}
Zhu, G.; Xu, Z.; Yuan, C.; and Huang, Y. 2022.
\newblock {DIFER}: Differentiable Automated Feature Engineering.
\newblock In \emph{Proceedings of the First International Conference on
  Automated Machine Learning (AutoML)}.

\end{thebibliography}

\clearpage
\raggedbottom 
\twocolumn[
\begin{center}
{\Large\bfseries Supplementary Material}
\end{center}
\vspace{0.5em}
]
\appendix

\section{Dataset Panel}
\label{app:datasets}

\begin{table}[t]
\centering
\small
\setlength{\tabcolsep}{5pt}
\begin{tabular}{@{}llrr@{}}
\toprule
Dataset & Task (metric) & Train rows & Raw cols \\
\midrule
house    & regression (R$^2$)  & 14{,}581 & 16 \\
airfoil  & regression (R$^2$)  &     501  &  5 \\
protein  & regression (R$^2$)  & 15{,}243 &  9 \\
concrete & regression (R$^2$)  &     343  &  8 \\
miami    & regression (R$^2$)  &  4{,}592 & 15 \\
churn    & binary (acc.)       &  6{,}400 & 11 \\
adult    & binary (acc.)       & 26{,}048 & 14 \\
higgs    & binary (acc.)       & 62{,}751 & 28 \\
students & multiclass (acc.)   &  1{,}474 & 36 \\
otto     & multiclass (acc.)   & 39{,}601 & 93 \\
\bottomrule
\end{tabular}
\caption{The 10-dataset panel. ``Train rows'' is the training-split size ---
the only split any search reads; the test split is used solely for the final
scoring of frozen selected programs.}
\label{tab:datasets}
\end{table}

Table~\ref{tab:datasets} lists the ten datasets, their task families, their
training-split sizes and raw column counts, and the metric each search
optimizes. Row counts are training-split sizes: the training split is the only
split any search reads, and the test split is used solely to score a frozen
selected program. All ten are public: \texttt{house} is the Delve \texttt{house\_16H}
task \citep{openml_house16h}; \texttt{airfoil} \citep{brooks1989airfoil},
\texttt{protein} \citep{rana2013protein}, \texttt{concrete}
\citep{yeh1998concrete}, \texttt{adult} \citep{kohavi1996adult},
\texttt{higgs} \citep{baldi2014higgs} and \texttt{students}
\citep{realinho2022students} are distributed by the UCI repository;
\texttt{miami} by OpenML \citep{bourassa2021miami}; \texttt{churn}
\citep{kaggle2019churn} and \texttt{otto} \citep{kaggle2015otto} by Kaggle.

\section{Memory}
\label{app:memory}

This section states in full the framework's memory mechanism. It was active in
every search reported here: the thirty \method{} cells of
Table~\ref{tab:main} and every arm of the representation ablation in
Table~\ref{tab:ablation}.
Framework-maintained memory is run-local: its card bank starts empty in every
cell. With the parent and run statistics, the mutation prompt receives a
summary of up to 24 recently evaluated direct children of that parent and a
separate ancestor trail bounded at depth 8. A post-step pass asks a separate
LLM for at most one conditional mechanism card per valid mutation that was
archived or improved on its parent, and merges equivalents. A reader excludes
cards already proposed or applied on that branch, assesses applicability, and
delivers at most one, chosen from a posterior over the outcomes those cards
have been associated with, under a randomized 70/30 treatment--control split.
A suggestion stage reads the delivered card; only the suggestion enters the
mutation prompt, so card text never reaches the mutator directly. Credit is
intention-to-treat over all randomized deliveries, cited or not, and includes
a separate nonnegative best-accepted-descendant residual bounded at depth~3.

\section{Interpretability, Semantics, and Transfer on Held-Out Domains}
\label{sec:interp}

\paragraph{Inspectable frozen graphs and protocol.}
Here interpretability means that the transferred object is an explicit frozen
FeatureGraph: its selecting model, receiving model, and graph hash are
inspectable.  We call the model used during search the
\emph{selector} (source), the model used for held-out evaluation the
\emph{receiver} (target), a graph selected for its receiver
\emph{target-self}, and one selected by another model \emph{foreign}.  The
same terminology applies throughout all three experiments.  The
transfer study crosses Adult, Churn, HIGGS-small, House, Airfoil, Protein,
Otto, and Students with \catboost, XGBoost, TabICL, \tabpfn, RealMLP, and
\tabm.  Its held-out metrics are ROC AUC for binary tasks, \(R^2\) for
regression, and macro-F1 for multiclass tasks.  Two independent 100-mutant
searches were run for each of the 48 dataset--selector cells; each of the 96
resulting graphs was selected without test labels, then held fixed while the
receiving estimator was refit with seeds 0--9.

\paragraph{Cross-model transfer.}
The primary contrast is the paired held-out score of a foreign graph minus
the score of the receiver's target-self graph.  All three metrics in this
experiment are higher-is-better, so positive values always favor the foreign
graph.  Table~\ref{tab:section7-transfer} first averages the 10 paired refits
within each dataset--search--selector--receiver cell, then counts the
dataset--search combinations with a positive mean.  The bundle contains no Raw arm evaluated under this
protocol; the experiment therefore asks whether foreign selection can match
or outperform target-specific selection, not whether either graph improves
on Raw.

\begin{table*}[t!]
\centering
\scriptsize
\setlength{\tabcolsep}{7.0pt}
\renewcommand{\arraystretch}{1.05}
\begin{tabular}{@{}lrrrrrr@{}}
\toprule
Selector $\backslash$ receiver &
\catboost & XGBoost & TabICL & \tabpfn & RealMLP & \tabm \\
\midrule
\catboost &
\cellcolor{black!8}\textit{self} &
\cellcolor{cellmid}8/16 &
\cellcolor{cellfar}6/16 &
\cellcolor{cellfar}6/16 &
\cellcolor{cellnear}11/16 &
\cellcolor{cellfar}6/16 \\
XGBoost &
\cellcolor{cellfar}6/16 &
\cellcolor{black!8}\textit{self} &
\cellcolor{cellnear}9/16 &
\cellcolor{cellfar}5/16 &
\cellcolor{cellmid}8/16 &
\cellcolor{cellmid}8/16 \\
TabICL &
\cellcolor{cellworst}3/16 &
\cellcolor{cellfar}6/16 &
\cellcolor{black!8}\textit{self} &
\cellcolor{cellfar}6/16 &
\cellcolor{cellfar}7/16 &
\cellcolor{cellfar}6/16 \\
\tabpfn &
\cellcolor{cellworst}1/16 &
\cellcolor{cellfar}4/16 &
\cellcolor{cellfar}5/16 &
\cellcolor{black!8}\textit{self} &
\cellcolor{cellfar}5/16 &
\cellcolor{cellfar}5/16 \\
RealMLP &
\cellcolor{cellworst}2/16 &
\cellcolor{cellfar}4/16 &
\cellcolor{cellfar}5/16 &
\cellcolor{cellfar}4/16 &
\cellcolor{black!8}\textit{self} &
\cellcolor{cellworst}3/16 \\
\tabm &
\cellcolor{cellworst}2/16 &
\cellcolor{cellnear}9/16 &
\cellcolor{cellfar}4/16 &
\cellcolor{cellworst}2/16 &
\cellcolor{cellmid}8/16 &
\cellcolor{black!8}\textit{self} \\
\bottomrule
\end{tabular}
\caption{Cross-model transfer on eight held-out datasets.  Each off-diagonal entry is the number of dataset--search combinations (8 datasets $\times$ 2 independent searches $=$ 16) on which the foreign graph's mean score exceeds the receiver's target-self graph after averaging 10 paired evaluator seeds; AUC, \(R^2\), and macro-F1 are all higher-is-better.  Colors encode 12--16 wins (\colorbox{cellbest}{best}), 9--11 (\colorbox{cellnear}{near}), 8 (\colorbox{cellmid}{half}), 4--7 (\colorbox{cellfar}{few}), and 0--3 (\colorbox{cellworst}{rare}); gray is self-selection.  There is no matched Raw baseline in this bundle, so the table measures conditional portability relative to target-specific selection, not improvement over Raw.}
\label{tab:section7-transfer}
\end{table*}

Foreign graphs exceed target-self in 164 of the 480 off-diagonal
dataset--search cells and provide the best source in 66 of the 96
dataset--search--receiver columns.  However, most selector--receiver pairs win
on fewer than half of those combinations.  A graph selected with another
model can therefore be competitive, but portability is strongly
model-dependent rather than universal.

\paragraph{Top-five program means.}
A finer evaluation of one search per cell of
Table~\ref{tab:section7-transfer} goes beyond that search's single
selected program.  For every dataset--selector pair the five highest development-CV,
unique, non-empty graphs were frozen and re-fit under all six receivers with
three evaluator seeds (240 graphs, 4{,}320 fits); the \emph{top-five mean} is
the arithmetic mean of the five native-metric scores within one
dataset/selector/receiver/seed cell --- five separately fitted graph
conditions, not a prediction ensemble or a single deployable graph.  Rank~1
reproduced the original frozen selected program in 47 of 48 searches; House/RealMLP's
original selected program was a zero-node no-op graph, excluded by the non-empty
rule, and that search was re-selected and evaluated afresh.  On this basis a
foreign five-program mean beats the receiver's own in 89 of the 240
off-diagonal cells, with no ties
(Table~\ref{tab:section7-top5-transfer}), and the asymmetry is directional:
graphs selected against RealMLP win 3 of their 40 transfers, while foreign
selections beat RealMLP's own in 27 of 40 --- the weakest source is also the
easiest target.  Table~\ref{tab:section7-top5-rank1} pairs each top-five mean
with the same search's rank-1 graph in matched cells: rank~1 wins 186 of 288,
most heavily under RealMLP (40 of 48) and most narrowly under \tabm{} (25 of
48), so the development-CV pick at rank~1 carries signal beyond membership in
the top five.  The three evaluator seeds vary receiving-estimator fitting
only and do not repeat evolutionary search; the five programs are dependent
candidates from one search, not five search replications; this is a post-hoc
analysis of an existing campaign; and cross-dataset summaries are counts only
--- AUC, \(R^2\), and macro-F1 are never averaged together.

\begin{table*}[t!]
\centering
\scriptsize
\setlength{\tabcolsep}{7.0pt}
\renewcommand{\arraystretch}{1.05}
\begin{tabular}{@{}lrrrrrr@{}}
\toprule
Selector $\backslash$ receiver &
\catboost & XGBoost & TabICL & \tabpfn & RealMLP & \tabm \\
\midrule
\catboost &
\cellcolor{black!8}\textit{self} &
\cellcolor{cellfar}3/8 &
\cellcolor{cellmid}4/8 &
\cellcolor{cellmid}4/8 &
\cellcolor{cellnear}5/8 &
\cellcolor{cellmid}4/8 \\
XGBoost &
\cellcolor{cellfar}2/8 &
\cellcolor{black!8}\textit{self} &
\cellcolor{cellmid}4/8 &
\cellcolor{cellmid}4/8 &
\cellcolor{cellbest}7/8 &
\cellcolor{cellmid}4/8 \\
TabICL &
\cellcolor{cellfar}2/8 &
\cellcolor{cellfar}2/8 &
\cellcolor{black!8}\textit{self} &
\cellcolor{cellfar}2/8 &
\cellcolor{cellbest}6/8 &
\cellcolor{cellfar}2/8 \\
\tabpfn &
\cellcolor{cellworst}0/8 &
\cellcolor{cellfar}2/8 &
\cellcolor{cellbest}6/8 &
\cellcolor{black!8}\textit{self} &
\cellcolor{cellmid}4/8 &
\cellcolor{cellfar}2/8 \\
RealMLP &
\cellcolor{cellworst}0/8 &
\cellcolor{cellworst}0/8 &
\cellcolor{cellworst}1/8 &
\cellcolor{cellworst}1/8 &
\cellcolor{black!8}\textit{self} &
\cellcolor{cellworst}1/8 \\
\tabm &
\cellcolor{cellfar}2/8 &
\cellcolor{cellmid}4/8 &
\cellcolor{cellfar}3/8 &
\cellcolor{cellfar}3/8 &
\cellcolor{cellnear}5/8 &
\cellcolor{black!8}\textit{self} \\
\bottomrule
\end{tabular}
\caption{Cross-model transfer on the top-five-mean basis, for the searches of
Table~\ref{tab:section7-transfer}.  Each off-diagonal entry counts
datasets (of eight) on which the
foreign selector's five-program mean exceeds the receiving estimator's own
five-program mean after averaging three paired evaluator seeds; no cell tied
at machine precision.  Color coding follows
Table~\ref{tab:section7-transfer}.  There is one search realization per
dataset--selector pair, and no Raw arm is evaluated under this protocol, so
the table measures conditional portability relative to target-specific
selection, not improvement over Raw.}
\label{tab:section7-top5-transfer}
\end{table*}

\begin{table}[t!]
\centering
\scriptsize
\setlength{\tabcolsep}{6.0pt}
\renewcommand{\arraystretch}{1.05}
\begin{tabular}{@{}lrrr@{}}
\toprule
Selector & Top-5 mean better & Rank 1 better & Cells \\
\midrule
\catboost & 20 & 28 & 48 \\
XGBoost   & 16 & 32 & 48 \\
TabICL    & 14 & 34 & 48 \\
\tabpfn   & 21 & 27 & 48 \\
RealMLP   &  8 & 40 & 48 \\
\tabm     & 23 & 25 & 48 \\
\midrule
All       & 102 & 186 & 288 \\
\bottomrule
\end{tabular}
\caption{Top-five mean versus the same search's rank-1 graph, paired within
dataset/selector/receiver/seed cells of the searches of
Table~\ref{tab:section7-top5-transfer}; no cell tied at machine precision.  Rank~1 is the development-CV selected program; since it is one
of the five, the comparison reads as rank~1 against the mean of ranks 2--5.}
\label{tab:section7-top5-rank1}
\end{table}

\paragraph{Semantic metadata controls.}
The matched controls cover Gallstone Disease, 5G Energy Consumption,
SDSS-17, HIGGS-small, and Airfoil Self-Noise with \catboost, the
same 100-mutant budget, and development-CV selection.  All conditions retain
identical canonical \(x_j\) values, dtypes, column order, target, and
train/validation/test split: Semantic presents the correct feature metadata;
Anonymous replaces names, descriptions, units, and category meanings with
neutral packets; Shuffled attaches a compatible feature's deranged metadata
while leaving the receiving column's data unchanged.  Airfoil's singleton
categorical compatibility class instead receives the neutral packet.

\begin{table*}[t!]
\centering
\scriptsize
\setlength{\tabcolsep}{3.1pt}
\renewcommand{\arraystretch}{1.08}
\begin{tabular*}{\textwidth}{@{\extracolsep{\fill}}llrrrrrrrr@{}}
\toprule
Dataset & Metric & Raw & Semantic & Anonymous & Shuffled &
\(S-A\) & Holm \(p\) & \(S-\mathit{Sh}\) & Holm \(p\) \\
\midrule
\multicolumn{10}{@{}l}{\textbf{A. Ten paired evaluator refits}} \\
Gallstone &
AUC\(\uparrow\) &
\cellcolor{cellbest}\textbf{.8744{\tiny\(\pm\).0068}} &
\cellcolor{cellbest}\textbf{.8819{\tiny\(\pm\).0061}} &
\cellcolor{cellbest}\textbf{.8717{\tiny\(\pm\).0151}} &
\cellcolor{cellbest}\textbf{.8753{\tiny\(\pm\).0072}} &
\(+.0102\) & .1699 & \(+.0066\) & .1699 \\
5G Energy &
MAPE\(\downarrow\) &
\cellcolor{cellnear}.1067{\tiny\(\pm\).0020} &
\cellcolor{cellbest}\textbf{.1026{\tiny\(\pm\).0020}} &
\cellcolor{cellclose}.1076{\tiny\(\pm\).0011} &
\cellcolor{cellbest}\textbf{.1041{\tiny\(\pm\).0016}} &
\(\mathbf{+.0051}\) & \(\mathbf{.0117}\) &
\(+.0015\) & .1699 \\
SDSS-17 &
log loss\(\downarrow\) &
\cellcolor{cellfar}.0740{\tiny\(\pm\).0004} &
\cellcolor{cellnear}.0662{\tiny\(\pm\).0005} &
\cellcolor{cellbest}\textbf{.0652{\tiny\(\pm\).0006}} &
\cellcolor{cellmid}.0696{\tiny\(\pm\).0003} &
\(\mathbf{-.0010}\) & \(\mathbf{.0234}\) &
\(\mathbf{+.0034}\) & \(\mathbf{.0117}\) \\
\addlinespace[2pt]
\multicolumn{10}{@{}l}{\textbf{B. One evaluator fit (descriptive extension)}} \\
HIGGS-small &
AUC\(\uparrow\) &
.8054 &
.8136 &
.8084 &
.8074 &
\(+.0052\) & -- & \(+.0061\) & -- \\
Airfoil &
\(R^2\)\(\uparrow\) &
.9480 &
.9747 &
.9663 &
.9669 &
\(+.0084\) & -- & \(+.0078\) & -- \\
\bottomrule
\end{tabular*}
\caption{Matched schema controls on held-out test folds; every evolved
condition has one search seed.  Panel A reports
mean\(\,\pm\,\)sample SD over 10 paired \catboost refits.  Its positive
contrasts favor Semantic (subtraction reverses for lower-is-better metrics);
\(p\)-values are exact two-sided sign-flip tests, Holm-adjusted over its six
contrasts, with adjusted \(p<.05\) in bold.  For Panel A score colors, \(w\)
is the median sample SD among the six paired condition differences
(\(.01076,.002272,.0006886\), by row).  The oriented gap \(g\) from the
best mean maps to \colorbox{cellbest}{\(\leq w\), bold},
\colorbox{cellnear}{\(\leq2w\)}, \colorbox{cellclose}{\(\leq4w\)},
\colorbox{cellmid}{\(\leq8w\)}, \colorbox{cellfar}{\(\leq16w\)}, then
\colorbox{cellworst}{\(>16w\)}.  Panel B has one evaluator fit, so reports
neither uncertainty nor tests and is deliberately uncolored.  All inference
is conditional on frozen selected programs and fixed splits.}
\label{tab:section7-semantic}
\end{table*}

Correct metadata helps conditionally on some frozen selected programs, but not
monotonically.  Semantic beats Anonymous on 5G Energy (Holm
\(p=.0117\)); on SDSS-17, Anonymous beats Semantic (\(p=.0234\)), while
Semantic beats Shuffled (\(p=.0117\)).  Gallstone and the 5G
Semantic--Shuffled contrast are not statistically separated.  Where a
control condition does not exceed Raw in Panel A --- Anonymous on Gallstone
and on 5G Energy --- the shortfall is smaller than the refit spread of the
two conditions compared, so those rows tie with Raw rather than fall below
it.  In the
one-refit extension, Semantic is numerically best on both HIGGS-small and
Airfoil, while Anonymous and Shuffled also exceed Raw; these rows are
descriptive and carry neither uncertainty nor \(p\)-values.  Thus familiar
feature semantics are neither necessary nor sufficient for selecting the
best graph, although mismatched metadata can be harmful.

\paragraph{Evidence boundary.}
For Gallstone, 5G, and SDSS-17 we repeated model fitting, but not evolutionary
search; HIGGS-small and Airfoil repeat neither.  The uncertainty and
Holm-adjusted tests therefore describe estimator-fitting randomness only for
the first three datasets, conditional on their observed frozen selected programs and
fixed test splits, and do not estimate search-to-search variation.  Within
that boundary, explicit graph identity makes origin, transfer direction, and
schema condition auditable, while the results support only partial
portability and dataset-dependent semantic effects.

\paragraph{Absolute target-self scores.}
Table~\ref{tab:section7-selfscores} reports the absolute held-out
target-self score of every dataset and selector model, for the same one
search per cell that the top-five replay uses.  Each cell therefore
evaluates a graph with the same model that selected it; foreign-graph
comparisons remain in Table~\ref{tab:section7-transfer}.  Reporting the
absolute scores exposes the underlying numbers without pooling unlike AUC,
\(R^2\), and macro-F1 scales.

\begin{table*}[t!]
\centering
\scriptsize
\setlength{\tabcolsep}{4pt}
\renewcommand{\arraystretch}{0.95}
\begin{tabular*}{\textwidth}{@{\extracolsep{\fill}}lrrrrrr@{}}
\toprule
Dataset & \catboost & XGBoost & TabICL & \tabpfn & RealMLP & \tabm \\
\midrule
\multicolumn{7}{l}{\textbf{Binary (AUC)}} \\
Adult & 0.9282\,{\color{gray}\tiny$\pm$.0002} & 0.9279\,{\color{gray}\tiny$\pm$.0000} & 0.9283\,{\color{gray}\tiny$\pm$.0020} & 0.9196\,{\color{gray}\tiny$\pm$.0004} & 0.9125\,{\color{gray}\tiny$\pm$.0015} & 0.9170\,{\color{gray}\tiny$\pm$.0009} \\
Churn & 0.8666\,{\color{gray}\tiny$\pm$.0006} & 0.8590\,{\color{gray}\tiny$\pm$.0000} & 0.8670\,{\color{gray}\tiny$\pm$.0004} & 0.8708\,{\color{gray}\tiny$\pm$.0003} & 0.8605\,{\color{gray}\tiny$\pm$.0042} & 0.8647\,{\color{gray}\tiny$\pm$.0026} \\
HIGGS-small & 0.8140\,{\color{gray}\tiny$\pm$.0005} & 0.8118\,{\color{gray}\tiny$\pm$.0000} & 0.8227\,{\color{gray}\tiny$\pm$.0004} & 0.8256\,{\color{gray}\tiny$\pm$.0003} & 0.8141\,{\color{gray}\tiny$\pm$.0010} & 0.8310\,{\color{gray}\tiny$\pm$.0008} \\
\midrule
\multicolumn{7}{l}{\textbf{Regression (R$^2$)}} \\
House & 0.6737\,{\color{gray}\tiny$\pm$.0045} & 0.6542\,{\color{gray}\tiny$\pm$.0000} & 0.7249\,{\color{gray}\tiny$\pm$.0015} & 0.7506\,{\color{gray}\tiny$\pm$.0017} & 0.6539\,{\color{gray}\tiny$\pm$.0072} & 0.6676\,{\color{gray}\tiny$\pm$.0054} \\
Airfoil & 0.9736\,{\color{gray}\tiny$\pm$.0010} & 0.9627\,{\color{gray}\tiny$\pm$.0000} & 0.9831\,{\color{gray}\tiny$\pm$.0003} & 0.9859\,{\color{gray}\tiny$\pm$.0002} & 0.9748\,{\color{gray}\tiny$\pm$.0014} & 0.9752\,{\color{gray}\tiny$\pm$.0015} \\
Protein & 0.6467\,{\color{gray}\tiny$\pm$.0006} & 0.6767\,{\color{gray}\tiny$\pm$.0000} & 0.7611\,{\color{gray}\tiny$\pm$.0006} & 0.7689\,{\color{gray}\tiny$\pm$.0008} & 0.6821\,{\color{gray}\tiny$\pm$.0071} & 0.7079\,{\color{gray}\tiny$\pm$.0036} \\
\midrule
\multicolumn{7}{l}{\textbf{Multiclass (macro-F1)}} \\
Otto & 0.7687\,{\color{gray}\tiny$\pm$.0012} & 0.7831\,{\color{gray}\tiny$\pm$.0000} & 0.8179\,{\color{gray}\tiny$\pm$.0009} & 0.8093\,{\color{gray}\tiny$\pm$.0015} & 0.7815\,{\color{gray}\tiny$\pm$.0012} & 0.8028\,{\color{gray}\tiny$\pm$.0018} \\
Students & 0.7064\,{\color{gray}\tiny$\pm$.0048} & 0.6983\,{\color{gray}\tiny$\pm$.0000} & 0.7291\,{\color{gray}\tiny$\pm$.0020} & 0.7043\,{\color{gray}\tiny$\pm$.0020} & 0.7019\,{\color{gray}\tiny$\pm$.0118} & 0.7099\,{\color{gray}\tiny$\pm$.0072} \\
\bottomrule
\end{tabular*}
\caption{Held-out target-self scores of the frozen selected programs on the eight transfer datasets. Each cell is mean\(\,\pm\,\)sample SD over 10 evaluator refits of the graph selected and received by the column model; all metrics are higher-is-better. Each graph comes from one search, so uncertainty is conditional on the observed selected program and not search-to-search variation.}
\label{tab:section7-selfscores}
\end{table*}

\section{Search Engine}
\label{app:engine}

The archive is nominally a $15\times10$ grid over the two behavior
descriptors, with hard limits $[0,4]$ and $[0,2]$, but its bounds are dynamic:
they contract around the current elites after an accepted insertion and the
archive is rebinned, so a cell label is not comparable across the run. The
archive holds at most 75 elites, so at most half the nominal cells can be
occupied at once; final occupancy is a median of 12.5 elites per run over
the fifty panel searches whose archive statistics were extracted. Parent
selection is two-stage: elites are
softmax-weighted on normalized fitness at an auto-set temperature, a pool of
five times the requested number of parents is sampled without replacement, and
the parents are drawn uniformly from that pool. Two parents are requested per
proposal, a base and a donor: the first is drawn as above, and the second is
drawn the same way but only from elites whose fitness lies within
$1.5\times2\,\mathrm{MAD}$ of the first --- a same-stratum coupler that stops
a pair from spanning the archive's fitness range, falling back to all
remaining elites when too few candidates survive. The child is a diff against
the base parent; the donor is a source of citable nodes and not a second thing
to edit (Appendix~\ref{app:genome}).
Fitness is the estimator's metric under 3-fold
cross-validation, shuffled at a fixed seed, within the development split.

\section{Cost Accounting}
\label{app:cost}

\begin{table}[t]
\centering
\small
\setlength{\tabcolsep}{5pt}
\begin{tabular}{lrrrr}
\toprule
Arm & priced cells & total (\$) & \$/cell & to best (\$) \\
\midrule
\llmfe{} & 24 & 44.70 & 1.86 & 25.80 \\
\pythonarm{} & 21 & 102.17 & 4.87 & 51.99 \\
\jsonarm{} & 24 & 145.53 & 6.06 & 74.03 \\
\method{} & 17 & 107.28 & 6.31 & 60.56 \\
\midrule
total & 86 & 399.69 & & 212.39 \\
\bottomrule
\end{tabular}
\caption{LLM API spend per search arm, at OpenRouter rates (fetched 2026-07-25). ``Priced cells'' counts the cells with complete token accounting; the remainder are excluded from the totals rather than estimated, and Table~\ref{tab:costcell} shows which. ``To best'' prorates each cell's spend up to the iteration that produced its selected program. \llmfe{} is priced on its original selection wave. Memory-system tokens were served by a self-hosted model and carry no API price; \rawarm{} and \openfe{} use no LLM. Wall-clock was not matched across arms.}
\label{tab:costs}
\end{table}

\begin{table*}[t]
\centering
\footnotesize
\setlength{\tabcolsep}{3.5pt}
\begin{tabular}{llrrrr}
\toprule
dataset & est. & \llmfe{} & \pythonarm{} & \jsonarm{} & \method{} \\
\midrule
house & \catboost & 1.81\,{\color{gray}\scriptsize(0.81)} & 4.77\,{\color{gray}\scriptsize(4.35)} & 6.08\,{\color{gray}\scriptsize(5.39)} & 5.99\,{\color{gray}\scriptsize(3.74)} \\
 & \tabm & 2.00\,{\color{gray}\scriptsize(1.28)} & 5.41\,{\color{gray}\scriptsize(4.88)} & 6.31\,{\color{gray}\scriptsize(3.69)} & 5.87\,{\color{gray}\scriptsize(0.60)} \\
 & \tabpfn & 1.92\,{\color{gray}\scriptsize(0.64)} & 5.73\,{\color{gray}\scriptsize(2.09)} & 6.07\,{\color{gray}\scriptsize(0.72)} & 6.40\,{\color{gray}\scriptsize(6.32)} \\
\addlinespace[1pt]
airfoil & \catboost & 1.89\,{\color{gray}\scriptsize(0.76)} & 4.50\,{\color{gray}\scriptsize(0.99)} & 5.68\,{\color{gray}\scriptsize(5.46)} & 7.14\,{\color{gray}\scriptsize(6.93)} \\
 & \tabm & 2.04\,{\color{gray}\scriptsize(1.12)} & 4.59\,{\color{gray}\scriptsize(4.54)} & 5.65\,{\color{gray}\scriptsize(2.36)} & 6.52\,{\color{gray}\scriptsize(3.63)} \\
 & \tabpfn & 1.83\,{\color{gray}\scriptsize(0.93)} & 4.42\,{\color{gray}\scriptsize(3.68)} & 5.89\,{\color{gray}\scriptsize(2.92)} & 6.10\,{\color{gray}\scriptsize(1.86)} \\
\addlinespace[1pt]
protein & \catboost & 1.86\,{\color{gray}\scriptsize(1.30)} & 4.51\,{\color{gray}\scriptsize(0.97)} & 5.23\,{\color{gray}\scriptsize(3.45)} & 5.77\,{\color{gray}\scriptsize(5.64)} \\
 & \tabm & 1.79\,{\color{gray}\scriptsize(1.68)} & 4.33\,{\color{gray}\scriptsize(0.79)} & 5.46\,{\color{gray}\scriptsize(0.52)} & 5.04\,{\color{gray}\scriptsize(1.56)} \\
 & \tabpfn & 1.62\,{\color{gray}\scriptsize(1.26)} & 4.48\,{\color{gray}\scriptsize(0.07)} & 5.24\,{\color{gray}\scriptsize(5.24)} & 6.00\,{\color{gray}\scriptsize(0.59)} \\
\addlinespace[1pt]
concrete & \catboost & {\color{gray}--} & {\color{gray}--} & {\color{gray}--} & {\color{gray}--} \\
 & \tabm & {\color{gray}--} & {\color{gray}--} & {\color{gray}--} & {\color{gray}--} \\
 & \tabpfn & {\color{gray}--} & {\color{gray}--} & {\color{gray}--} & {\color{gray}--} \\
\addlinespace[1pt]
miami & \catboost & {\color{gray}--} & {\color{gray}--} & {\color{gray}--} & {\color{gray}--} \\
 & \tabm & {\color{gray}--} & {\color{gray}--} & {\color{gray}--} & {\color{gray}--} \\
 & \tabpfn & {\color{gray}--} & {\color{gray}--} & {\color{gray}--} & {\color{gray}--} \\
\addlinespace[1pt]
churn & \catboost & 1.42\,{\color{gray}\scriptsize(1.42)} & 4.37\,{\color{gray}\scriptsize(3.56)} & 5.31\,{\color{gray}\scriptsize(1.81)} & 6.77\,{\color{gray}\scriptsize(4.12)} \\
 & \tabm & 1.46\,{\color{gray}\scriptsize(1.23)} & 4.67\,{\color{gray}\scriptsize(0.88)} & 5.68\,{\color{gray}\scriptsize(0.15)} & 5.80\,{\color{gray}\scriptsize(1.35)} \\
 & \tabpfn & 1.44\,{\color{gray}\scriptsize(0.01)} & 4.58\,{\color{gray}\scriptsize(3.13)} & 5.19\,{\color{gray}\scriptsize(0.59)} & 5.93\,{\color{gray}\scriptsize(1.75)} \\
\addlinespace[1pt]
adult & \catboost & 1.70\,{\color{gray}\scriptsize(1.58)} & 5.07\,{\color{gray}\scriptsize(0.08)} & 6.68\,{\color{gray}\scriptsize(0.37)} & 6.06\,{\color{gray}\scriptsize(3.90)} \\
 & \tabm & 1.68\,{\color{gray}\scriptsize(0.13)} & 4.94\,{\color{gray}\scriptsize(0.00)} & 6.09\,{\color{gray}\scriptsize(3.72)} & {\color{gray}--} \\
 & \tabpfn & 1.51\,{\color{gray}\scriptsize(1.36)} & 4.62\,{\color{gray}\scriptsize(2.80)} & 5.87\,{\color{gray}\scriptsize(2.54)} & 6.15\,{\color{gray}\scriptsize(5.87)} \\
\addlinespace[1pt]
higgs & \catboost & 2.40\,{\color{gray}\scriptsize(2.18)} & 4.94\,{\color{gray}\scriptsize(4.02)} & 6.07\,{\color{gray}\scriptsize(4.62)} & {\color{gray}--} \\
 & \tabm & 2.29\,{\color{gray}\scriptsize(2.18)} & {\color{gray}--} & 5.94\,{\color{gray}\scriptsize(3.59)} & {\color{gray}--} \\
 & \tabpfn & 2.29\,{\color{gray}\scriptsize(0.76)} & 5.05\,{\color{gray}\scriptsize(4.50)} & 6.47\,{\color{gray}\scriptsize(2.03)} & {\color{gray}--} \\
\addlinespace[1pt]
students & \catboost & 1.76\,{\color{gray}\scriptsize(0.39)} & 5.28\,{\color{gray}\scriptsize(3.32)} & 6.48\,{\color{gray}\scriptsize(5.99)} & 6.84\,{\color{gray}\scriptsize(4.27)} \\
 & \tabm & 2.02\,{\color{gray}\scriptsize(1.65)} & 5.48\,{\color{gray}\scriptsize(3.46)} & 6.59\,{\color{gray}\scriptsize(2.92)} & 7.40\,{\color{gray}\scriptsize(3.42)} \\
 & \tabpfn & 1.86\,{\color{gray}\scriptsize(0.95)} & 4.92\,{\color{gray}\scriptsize(0.90)} & 6.29\,{\color{gray}\scriptsize(5.67)} & 7.51\,{\color{gray}\scriptsize(5.03)} \\
\addlinespace[1pt]
otto & \catboost & 2.11\,{\color{gray}\scriptsize(0.97)} & {\color{gray}--} & 7.84\,{\color{gray}\scriptsize(0.49)} & {\color{gray}--} \\
 & \tabm & 2.01\,{\color{gray}\scriptsize(1.19)} & {\color{gray}--} & 6.80\,{\color{gray}\scriptsize(4.12)} & {\color{gray}--} \\
 & \tabpfn & 1.97\,{\color{gray}\scriptsize(0.00)} & 5.52\,{\color{gray}\scriptsize(2.97)} & 6.63\,{\color{gray}\scriptsize(5.68)} & {\color{gray}--} \\
\midrule
\textbf{total} & & \textbf{44.70}\,{\color{gray}\scriptsize(25.80)} & \textbf{102.17}\,{\color{gray}\scriptsize(51.99)} & \textbf{145.53}\,{\color{gray}\scriptsize(74.03)} & \textbf{107.28}\,{\color{gray}\scriptsize(60.56)} \\
\bottomrule
\end{tabular}
\caption{Search spend per cell in US dollars, with the spend already incurred when that cell's selected program was first created in gray parentheses. Dashes are cells whose token accounting is incomplete: they are excluded from the totals rather than estimated, which is why the priced-cell counts of Table~\ref{tab:costs} fall short of thirty. Rates as in Table~\ref{tab:costs}.}
\label{tab:costcell}
\end{table*}

Table~\ref{tab:costs} reports LLM API spend per search arm under a shared
price sheet, and Table~\ref{tab:costcell} the cell-by-cell breakdown behind
it. Both are exact sums of logged token counts at posted rates ---
accounting identities, not sampled measurements, so they carry no error bars
by construction. Rejected diffs still consume their LLM calls, so the
decode-time gate of the mutation schema reduces wasted evaluations rather
than wasted proposals. Per priced cell, spend runs from \$1.86 for \llmfe{}
to \$6.31 for \method{}, with the \jsonarm{} rewrite at \$6.06 and
\pythonarm{} at \$4.87. Once gain is divided by spend, no difference between
the LLM arms survives correction; the closest, \llmfe{} above \method{} on
both gain per dollar and gain per token, reaches only $p\approx.15$. The arms
were not matched on wall-clock time, so these figures order the arms by
expenditure and not by efficiency.

The ledger prices the campaign over the cells whose token accounting is
complete: \$399.69 across 86 cells, a subset of the study's total spend,
which also covers the unpriced cells, the re-run cells, the
$\cup$\,OpenFE recompositions and the probes. Nothing here is estimated to
fill a gap. A re-run cell's billable cost is comparable to a first run's,
because the budget of 100 evaluated mutants and the backbone model are
unchanged; the memory mechanism's own calls carry no API price, as the
caption notes.

\clearpage
\section{Representation Ablation}
\label{app:ablation}

\begin{table*}[t]
\centering
\footnotesize
\setlength{\tabcolsep}{4pt}
\renewcommand{\arraystretch}{0.95}
\begin{tabular}{llrrrr}
\toprule
 &  & \multicolumn{1}{c}{Reference} & \multicolumn{3}{c}{Evolved representations} \\
\cmidrule(lr){3-3}\cmidrule(lr){4-6}
Dataset & Est. & \rawarm & \jsonarm & \method & \pythonarm \\
\midrule
\multicolumn{6}{l}{\textbf{Regression (R$^2$)}} \\
house & CatBoost & \cellcolor{cellbest}\textbf{0.6729}\,{\color{gray}\tiny$\pm$.0024} & \cellcolor{cellclose}0.6654\,{\color{gray}\tiny$\pm$.0034}$^{\blacktriangledown}$ & \cellcolor{cellnear}0.6713\,{\color{gray}\tiny$\pm$.0046}$^{\blacktriangledown}$ & \cellcolor{cellmid}0.6642\,{\color{gray}\tiny$\pm$.0026}$^{\blacktriangledown}$ \\
 & TabM & \cellcolor{cellbest}\textbf{0.6564}\,{\color{gray}\tiny$\pm$.0079} & \cellcolor{cellclose}0.6547\,{\color{gray}\tiny$\pm$.0045} & \cellcolor{cellbest}\textbf{0.6603}\,{\color{gray}\tiny$\pm$.0121} & \cellcolor{cellbest}\textbf{0.6567}\,{\color{gray}\tiny$\pm$.0156} \\
 & TabPFN & \cellcolor{cellnear}0.7496\,{\color{gray}\tiny$\pm$.0018} & \cellcolor{cellnear}0.7478\,{\color{gray}\tiny$\pm$.0017}$^{\blacktriangledown}$ & \cellcolor{cellbest}\textbf{0.7515}\,{\color{gray}\tiny$\pm$.0011}$^{\blacktriangle}$ & \cellcolor{cellnear}0.7498\,{\color{gray}\tiny$\pm$.0013} \\
airfoil & CatBoost & \cellcolor{cellfar}0.9460\,{\color{gray}\tiny$\pm$.0047} & \cellcolor{cellbest}\textbf{0.9717}\,{\color{gray}\tiny$\pm$.0007}$^{\blacktriangle}$ & \cellcolor{cellbest}\textbf{0.9729}\,{\color{gray}\tiny$\pm$.0010}$^{\blacktriangle}$ & \cellcolor{cellnear}0.9694\,{\color{gray}\tiny$\pm$.0010}$^{\blacktriangle}$ \\
 & TabM & \cellcolor{cellnear}0.9740\,{\color{gray}\tiny$\pm$.0025} & \cellcolor{cellbest}\textbf{0.9764}\,{\color{gray}\tiny$\pm$.0013}$^{\blacktriangle}$ & \cellcolor{cellbest}\textbf{0.9758}\,{\color{gray}\tiny$\pm$.0015}$^{\blacktriangle}$ & \cellcolor{cellbest}\textbf{0.9765}\,{\color{gray}\tiny$\pm$.0031}$^{\blacktriangle}$ \\
 & TabPFN & \cellcolor{cellnear}0.9860\,{\color{gray}\tiny$\pm$.0003} & \cellcolor{cellbest}\textbf{0.9869}\,{\color{gray}\tiny$\pm$.0002}$^{\blacktriangle}$ & \cellcolor{cellnear}0.9861\,{\color{gray}\tiny$\pm$.0003}$^{\blacktriangle}$ & \cellcolor{cellnear}0.9866\,{\color{gray}\tiny$\pm$.0002}$^{\blacktriangle}$ \\
protein & CatBoost & \cellcolor{cellworst}0.6148\,{\color{gray}\tiny$\pm$.0005} & \cellcolor{cellnear}0.6470\,{\color{gray}\tiny$\pm$.0009}$^{\blacktriangle}$ & \cellcolor{cellnear}0.6458\,{\color{gray}\tiny$\pm$.0012}$^{\blacktriangle}$ & \cellcolor{cellbest}\textbf{0.6489}\,{\color{gray}\tiny$\pm$.0008}$^{\blacktriangle}$ \\
 & TabM & \cellcolor{cellfar}0.6766\,{\color{gray}\tiny$\pm$.0046} & \cellcolor{cellclose}0.7002\,{\color{gray}\tiny$\pm$.0062}$^{\blacktriangle}$ & \cellcolor{cellbest}\textbf{0.7079}\,{\color{gray}\tiny$\pm$.0045}$^{\blacktriangle}$ & \cellcolor{cellbest}\textbf{0.7076}\,{\color{gray}\tiny$\pm$.0073}$^{\blacktriangle}$ \\
 & TabPFN & \cellcolor{cellnear}0.7693\,{\color{gray}\tiny$\pm$.0006} & \cellcolor{cellnear}0.7697\,{\color{gray}\tiny$\pm$.0005}$^{\blacktriangle}$ & \cellcolor{cellbest}\textbf{0.7700}\,{\color{gray}\tiny$\pm$.0006}$^{\blacktriangle}$ & \cellcolor{cellnear}0.7696\,{\color{gray}\tiny$\pm$.0008} \\
concrete & CatBoost & \cellcolor{cellmid}0.9303\,{\color{gray}\tiny$\pm$.0024} & \cellcolor{cellmid}0.9349\,{\color{gray}\tiny$\pm$.0023}$^{\blacktriangle}$ & \cellcolor{cellbest}\textbf{0.9443}\,{\color{gray}\tiny$\pm$.0014}$^{\blacktriangle}$ & \cellcolor{cellnear}0.9417\,{\color{gray}\tiny$\pm$.0012}$^{\blacktriangle}$ \\
 & TabM & \cellcolor{cellfar}0.9207\,{\color{gray}\tiny$\pm$.0055} & \cellcolor{cellclose}0.9324\,{\color{gray}\tiny$\pm$.0023}$^{\blacktriangle}$ & \cellcolor{cellbest}\textbf{0.9394}\,{\color{gray}\tiny$\pm$.0028}$^{\blacktriangle}$ & \cellcolor{cellbest}\textbf{0.9374}\,{\color{gray}\tiny$\pm$.0024}$^{\blacktriangle}$ \\
 & TabPFN & \cellcolor{cellbest}\textbf{0.9545}\,{\color{gray}\tiny$\pm$.0005} & \cellcolor{cellnear}0.9542\,{\color{gray}\tiny$\pm$.0005}$^{\blacktriangledown}$ & \cellcolor{cellbest}\textbf{0.9550}\,{\color{gray}\tiny$\pm$.0005} & \cellcolor{cellnear}0.9537\,{\color{gray}\tiny$\pm$.0004}$^{\blacktriangledown}$ \\
miami & CatBoost & \cellcolor{cellnear}0.9347\,{\color{gray}\tiny$\pm$.0010} & \cellcolor{cellbest}\textbf{0.9365}\,{\color{gray}\tiny$\pm$.0008}$^{\blacktriangle}$ & \cellcolor{cellnear}0.9345\,{\color{gray}\tiny$\pm$.0012} & \cellcolor{cellbest}\textbf{0.9366}\,{\color{gray}\tiny$\pm$.0009}$^{\blacktriangle}$ \\
 & TabM & \cellcolor{cellbest}\textbf{0.9303}\,{\color{gray}\tiny$\pm$.0048} & \cellcolor{cellmid}0.9206\,{\color{gray}\tiny$\pm$.0078}$^{\blacktriangledown}$ & \cellcolor{cellbest}\textbf{0.9338}\,{\color{gray}\tiny$\pm$.0016} & \cellcolor{cellnear}0.9302\,{\color{gray}\tiny$\pm$.0015} \\
 & TabPFN & \cellcolor{cellclose}0.9397\,{\color{gray}\tiny$\pm$.0003} & \cellcolor{cellnear}0.9438\,{\color{gray}\tiny$\pm$.0002}$^{\blacktriangle}$ & \cellcolor{cellnear}0.9440\,{\color{gray}\tiny$\pm$.0003}$^{\blacktriangle}$ & \cellcolor{cellbest}\textbf{0.9449}\,{\color{gray}\tiny$\pm$.0004}$^{\blacktriangle}$ \\
\midrule
\multicolumn{6}{l}{\textbf{Binary (accuracy)}} \\
churn & CatBoost & \cellcolor{cellnear}0.8608\,{\color{gray}\tiny$\pm$.0014} & \cellcolor{cellbest}\textbf{0.8617}\,{\color{gray}\tiny$\pm$.0020} & \cellcolor{cellbest}\textbf{0.8624}\,{\color{gray}\tiny$\pm$.0023}$^{\blacktriangle}$ & \cellcolor{cellbest}\textbf{0.8622}\,{\color{gray}\tiny$\pm$.0007}$^{\blacktriangle}$ \\
 & TabM & \cellcolor{cellbest}\textbf{0.8619}\,{\color{gray}\tiny$\pm$.0020} & \cellcolor{cellbest}\textbf{0.8610}\,{\color{gray}\tiny$\pm$.0018} & \cellcolor{cellbest}\textbf{0.8627}\,{\color{gray}\tiny$\pm$.0038} & \cellcolor{cellbest}\textbf{0.8619}\,{\color{gray}\tiny$\pm$.0028} \\
 & TabPFN & \cellcolor{cellbest}\textbf{0.8658}\,{\color{gray}\tiny$\pm$.0006} & \cellcolor{cellnear}0.8646\,{\color{gray}\tiny$\pm$.0010}$^{\blacktriangledown}$ & \cellcolor{cellnear}0.8637\,{\color{gray}\tiny$\pm$.0007}$^{\blacktriangledown}$ & \cellcolor{cellbest}\textbf{0.8663}\,{\color{gray}\tiny$\pm$.0010} \\
adult & CatBoost & \cellcolor{cellnear}0.8740\,{\color{gray}\tiny$\pm$.0005} & \cellcolor{cellnear}0.8726\,{\color{gray}\tiny$\pm$.0005}$^{\blacktriangledown}$ & \cellcolor{cellnear}0.8737\,{\color{gray}\tiny$\pm$.0005} & \cellcolor{cellbest}\textbf{0.8744}\,{\color{gray}\tiny$\pm$.0007}$^{\blacktriangle}$ \\
 & TabM & \cellcolor{cellnear}0.8626\,{\color{gray}\tiny$\pm$.0010} & \cellcolor{cellbest}\textbf{0.8645}\,{\color{gray}\tiny$\pm$.0010}$^{\blacktriangle}$ & \cellcolor{cellnear}0.8627\,{\color{gray}\tiny$\pm$.0009} & \cellcolor{cellnear}0.8626\,{\color{gray}\tiny$\pm$.0010} \\
 & TabPFN & \cellcolor{cellmid}0.8623\,{\color{gray}\tiny$\pm$.0003} & \cellcolor{cellclose}0.8648\,{\color{gray}\tiny$\pm$.0002}$^{\blacktriangle}$ & \cellcolor{cellbest}\textbf{0.8681}\,{\color{gray}\tiny$\pm$.0005}$^{\blacktriangle}$ & \cellcolor{cellnear}0.8662\,{\color{gray}\tiny$\pm$.0003}$^{\blacktriangle}$ \\
higgs & CatBoost & \cellcolor{cellmid}0.7271\,{\color{gray}\tiny$\pm$.0010} & \cellcolor{cellbest}\textbf{0.7343}\,{\color{gray}\tiny$\pm$.0016}$^{\blacktriangle}$ & \cellcolor{cellnear}0.7320\,{\color{gray}\tiny$\pm$.0010}$^{\blacktriangle}$ & \cellcolor{cellnear}0.7325\,{\color{gray}\tiny$\pm$.0012}$^{\blacktriangle}$ \\
 & TabM & \cellcolor{cellmid}0.7334\,{\color{gray}\tiny$\pm$.0014} & \cellcolor{cellbest}\textbf{0.7429}\,{\color{gray}\tiny$\pm$.0009}$^{\blacktriangle}$ & \cellcolor{cellclose}0.7391\,{\color{gray}\tiny$\pm$.0020}$^{\blacktriangle}$ & \cellcolor{cellbest}\textbf{0.7436}\,{\color{gray}\tiny$\pm$.0012}$^{\blacktriangle}$ \\
 & TabPFN & \cellcolor{cellclose}0.7396\,{\color{gray}\tiny$\pm$.0005} & \cellcolor{cellbest}\textbf{0.7444}\,{\color{gray}\tiny$\pm$.0006}$^{\blacktriangle}$ & \cellcolor{cellnear}0.7436\,{\color{gray}\tiny$\pm$.0011}$^{\blacktriangle}$ & \cellcolor{cellnear}0.7424\,{\color{gray}\tiny$\pm$.0005}$^{\blacktriangle}$ \\
\midrule
\multicolumn{6}{l}{\textbf{Multiclass (accuracy)}} \\
students & CatBoost & \cellcolor{cellmid}0.7649\,{\color{gray}\tiny$\pm$.0042} & \cellcolor{cellbest}\textbf{0.7713}\,{\color{gray}\tiny$\pm$.0029}$^{\blacktriangle}$ & \cellcolor{cellbest}\textbf{0.7753}\,{\color{gray}\tiny$\pm$.0033}$^{\blacktriangle}$ & \cellcolor{cellbest}\textbf{0.7722}\,{\color{gray}\tiny$\pm$.0031}$^{\blacktriangle}$ \\
 & TabM & \cellcolor{cellbest}\textbf{0.7704}\,{\color{gray}\tiny$\pm$.0042} & \cellcolor{cellclose}0.7679\,{\color{gray}\tiny$\pm$.0066} & \cellcolor{cellbest}\textbf{0.7719}\,{\color{gray}\tiny$\pm$.0031} & \cellcolor{cellbest}\textbf{0.7742}\,{\color{gray}\tiny$\pm$.0049} \\
 & TabPFN & \cellcolor{cellnear}0.7732\,{\color{gray}\tiny$\pm$.0015} & \cellcolor{cellbest}\textbf{0.7759}\,{\color{gray}\tiny$\pm$.0017}$^{\blacktriangle}$ & \cellcolor{cellnear}0.7734\,{\color{gray}\tiny$\pm$.0018} & \cellcolor{cellbest}\textbf{0.7761}\,{\color{gray}\tiny$\pm$.0027}$^{\blacktriangle}$ \\
otto & CatBoost & \cellcolor{cellnear}0.8104\,{\color{gray}\tiny$\pm$.0006} & \cellcolor{cellbest}\textbf{0.8136}\,{\color{gray}\tiny$\pm$.0013}$^{\blacktriangle}$ & \cellcolor{cellbest}\textbf{0.8130}\,{\color{gray}\tiny$\pm$.0010}$^{\blacktriangle}$ & \cellcolor{cellnear}0.8121\,{\color{gray}\tiny$\pm$.0012}$^{\blacktriangle}$ \\
 & TabM & \cellcolor{cellbest}\textbf{0.8351}\,{\color{gray}\tiny$\pm$.0017} & \cellcolor{cellbest}\textbf{0.8344}\,{\color{gray}\tiny$\pm$.0015} & \cellcolor{cellbest}\textbf{0.8347}\,{\color{gray}\tiny$\pm$.0011} & \cellcolor{cellbest}\textbf{0.8341}\,{\color{gray}\tiny$\pm$.0013} \\
 & TabPFN & \cellcolor{cellnear}0.8382\,{\color{gray}\tiny$\pm$.0011} & \cellcolor{cellbest}\textbf{0.8394}\,{\color{gray}\tiny$\pm$.0013}$^{\blacktriangle}$ & \cellcolor{cellbest}\textbf{0.8397}\,{\color{gray}\tiny$\pm$.0010}$^{\blacktriangle}$ & \cellcolor{cellbest}\textbf{0.8397}\,{\color{gray}\tiny$\pm$.0005}$^{\blacktriangle}$ \\
\midrule
\multicolumn{2}{l}{\textit{rows above \rawarm{} ($\blacktriangle$)}} & {\color{gray}--} & 19 & 18 & 19 \\
\multicolumn{2}{l}{\textit{rows in the best group}} & 8/30 & 15/30 & 18/30 & 17/30 \\
\midrule
\multicolumn{2}{l}{\textit{structural failures / 3{,}000 candidates}} & {\color{gray}--} & $\geq$86 {\color{gray}\tiny(2.9\%)} & \textbf{0} {\color{gray}\tiny(0\%)} & 4 {\color{gray}\tiny(0.1\%)} \\
\multicolumn{2}{l}{\textit{runtime failures / 3{,}000 candidates}} & {\color{gray}--} & $\leq$728 {\color{gray}\tiny(24.3\%)} & 602 {\color{gray}\tiny(20.1\%)} & 834 {\color{gray}\tiny(27.8\%)} \\
\bottomrule
\end{tabular}
\caption{Representation ablation: the three evolved encodings --- raw-source \pythonarm{} rewrite, whole-graph \jsonarm{} rewrite, and \method's schema-constrained graph diff --- under identical launch conditions and budgets, so that the encoding the LLM mutates is the only variable. Cell format, shading and superscripts as in Table~\ref{tab:main}, recomputed within these columns; the \method{} column keeps this series' own selected programs, so its entries can differ from the programs reported of Table~\ref{tab:main}. In the counter rows (pooled over all 30 cells, 3{,}000 evaluated candidates per arm), \emph{structural} failures are candidates whose encoding is ill-formed (unparsable JSON, Python syntax errors, graph-schema violations) and \emph{runtime} failures parse but fail at evaluation. The two rows partition each arm's lost evaluations, so their sum is its total failure count --- 602 for \method{} against 814 and 838, the figures quoted in \S\ref{sec:results}; the \jsonarm{} bounds move together, so that arm's total is exact even though its split is not. \method{} rejects malformed proposals at the schema gate of \S\ref{sec:diff}, before candidacy; the other two arms have no such gate. The \jsonarm{} count is a lower bound --- among its 161 validator-rejected programs still inspectable on disk, 23 more are graph-schema violations --- and the classification is conservative \emph{against} \method: unavailable-column references, which its diff schema rejects structurally, surface as runtime errors elsewhere.}
\label{tab:ablation}
\end{table*}

Table~\ref{tab:ablation} holds the cell-level scores behind R5, with per-arm
counters for rows above \rawarm, best-group membership, and structural and
runtime failures. The ladder varies exactly one thing --- what the mutation
LLM is asked to emit --- and holds the rest fixed. \pythonarm{} asks for the
whole feature-engineering program as raw Python source, rewritten from the
parent's source, which is the form the \llmfe{} baseline uses. \jsonarm{} asks
for the whole FeatureGraph of Appendix~\ref{app:genome} as JSON, rewritten
from the parent's graph. \method{} asks for a diff against the parent graph
under the schema of that same appendix. Archive and selection rule, memory
configuration, backbone LLM, the budget of 100 evaluated mutants, the thirty
dataset--estimator cells, the ten evaluator seeds and the development-side
selection signal are identical across the three. What is not identical, and
cannot be, is the decode-time gate: \jsonarm{} writes graphs in the same
vocabulary, but its node identifiers are not retyped against the parent and
nothing is rejected before evaluation. The ladder's lower step therefore
isolates the graph vocabulary and its upper step the constrained edit that
rides on it.

All three arms ran with eight mutation proposals in flight, one search per arm
per cell, no pooling of searches inside a cell. Eight in flight was chosen for
wall-clock efficiency: it keeps the evaluators busy and is markedly faster in
wall-clock than issuing proposals one at a time. It is not free. Concurrent
proposals branch from parents that do not yet reflect each other's outcomes,
so at a budget as small as 100 evaluated mutants the search sees a staler
picture of the archive than a strictly sequential loop would, and it may
therefore search less well. Holding that setting across all three arms is what
makes the ladder internally matched, and it is also why this table's \method{}
column is a separate series of searches from the one in Table~\ref{tab:main},
which reports the sequential protocol of Appendix~\ref{app:config}. Every
comparison below is read within Table~\ref{tab:ablation}.

Read jointly rather than pairwise, over the thirty cells and including
\rawarm{} as a fourth block, the mean ranks are 1.92 for \method, 2.12 for
\pythonarm, 2.63 for \jsonarm{} and 3.33 for \rawarm{} (Friedman
$\chi^2(3)=21.81$, $p=7.2\times10^{-5}$); the Nemenyi critical difference at
$\alpha=.05$ is 0.86, which separates each encoding from \rawarm{} but not the
encodings from one another. The same ordering shows in the table's own
counters: \method{} takes 18 of the 30 best-group rows, \pythonarm{} 17 and
\jsonarm{} 15. The paired step-by-step margins behind that ordering are quoted
in the results section and are not restated here. R5 claims the ordering and
those margins, not a corrected pairwise separation, and none of the pairwise
comparisons was fixed before test scores were read.

What the encodings waste differs more sharply than what they score. The
failure counters of Table~\ref{tab:ablation} are pooled over all thirty cells,
3{,}000 evaluated candidates per arm, and split into candidates whose encoding
never parsed into a runnable program and candidates that parsed and failed at
evaluation; the caption gives the classification rule and the one lower bound
in it. Two further contrasts are read off the candidates rather than the
counters. First, the own-target guard of Appendix~\ref{app:genome}, which
rejects an aggregate node that consumes its own target column, can be counted
only where a run store still holds its discarded programs --- the \catboost{}
cell of each dataset; the \tabm{} and \tabpfn{} stores evict them. On that
ten-cell subset per arm the guard accounts for 16 of \method's 121 classifiable
rejections, 59 of \jsonarm's 161 and 89 of \pythonarm's 239, which is the
order the results section states them in. Second, a produced column counts as a
\emph{monotone recoding} when its Spearman correlation against some single raw
column is $\pm 1$ to a tolerance of $10^{-4}$ --- an order-preserving rewrite
of one input, carrying nothing a tree-based estimator could not already
extract from that input. Across the arms' selected program feature sets --- thirty
cells each for \jsonarm{} and \method, twenty-nine for \pythonarm, which
contributed no usable added column in one cell --- the monotone share is 0.11
for \method, 0.17 for \jsonarm{} and 0.30 for \pythonarm.

Table~\ref{tab:structure} takes the two graph arms apart node by node. Under
the matched ladder setting the \jsonarm{} rewrite builds wider selected programs ---
$3.34$ nodes against $2.97$, $8.66$ distinct operator families against $6.53$,
$705$ characters of code per node against $521$, and a larger share of
aggregate nodes ($0.39$ against $0.28$) --- while \method{} builds deeper ones:
$1.47$ dependency levels against $1.28$, and $0.15$ of its nodes consume
another node's output against $0.08$. The whole-graph rewrite adds material at
the same level; the diff, which always starts from a parent that already
works, more often stacks on top of it.

\begin{table*}[t]
\centering
\small
\setlength{\tabcolsep}{5pt}
\begin{tabular}{lrrrrrr}
\toprule
 & \multicolumn{2}{c}{\jsonarm{}} & \multicolumn{2}{c}{\method{} (ladder)} & \multicolumn{2}{c}{\method{} sequential} \\
\cmidrule(lr){2-3}\cmidrule(lr){4-5}\cmidrule(lr){6-7}
 & mean $\pm$ sd & median & mean $\pm$ sd & median & mean $\pm$ sd & median \\
\midrule
cells & \multicolumn{2}{c}{29} & \multicolumn{2}{c}{30} & \multicolumn{2}{c}{20} \\
\midrule
nodes per selected program & $3.34 \pm 1.95$ & 3.00 & $2.97 \pm 1.92$ & 3.00 & $3.60 \pm 2.19$ & 3.00 \\
graph depth (levels) & $1.28 \pm 0.45$ & 1.00 & $1.47 \pm 0.68$ & 1.00 & $1.50 \pm 0.69$ & 1.50 \\
nodes fed by another node & $0.08 \pm 0.15$ & 0.00 & $0.15 \pm 0.23$ & 0.00 & $0.17 \pm 0.19$ & 0.11 \\
aggregate nodes (share) & $0.39 \pm 0.25$ & 0.50 & $0.28 \pm 0.33$ & 0.19 & $0.25 \pm 0.26$ & 0.25 \\
distinct operator families & $8.66 \pm 3.71$ & 9.00 & $6.53 \pm 2.83$ & 6.00 & $7.20 \pm 3.38$ & 8.00 \\
input columns per node & $13.85 \pm 27.51$ & 4.33 & $14.66 \pm 26.95$ & 4.58 & $3.85 \pm 2.93$ & 2.50 \\
raw columns touched (share) & $0.77 \pm 0.26$ & 0.87 & $0.73 \pm 0.29$ & 0.84 & $0.60 \pm 0.38$ & 0.55 \\
code characters per node & $705 \pm 469$ & 552 & $521 \pm 476$ & 429 & $519 \pm 437$ & 320 \\
\bottomrule
\end{tabular}
\caption{Structure of the selected graphs; the \emph{cells} row gives how many selected programs each column summarizes. The \jsonarm{} column covers 29 of the 30 cells --- house\,/\,\catboost{} is not in this census --- so it is the one column whose counts fall short of the cells the arm was run on. The first two series are the \jsonarm{} and \method{} arms of the representation ladder (Appendix~\ref{app:ablation}), matched at eight proposals in flight; the third is the sequential (one proposal in flight) \method{} search. The \pythonarm{} rewrite has no column here: its selected programs are Python source, not graphs. A node is \emph{fed by another node} when it consumes a column produced upstream rather than a raw column, and \emph{aggregate} nodes are the ones that fit state on the frozen training frame. Depth is the number of levels in the dependency graph. Column counts per selected program are the analysis section's statistic and are not repeated here.}
\label{tab:structure}
\end{table*}

One further screen touches this table. Beyond the representation-level
guards, the shipped selected program of every arm is put through a program-level
label-leakage screen before its scores enter any table: its transform is run
twice on the same frame, once as delivered and once with the frame's row index
offset by 10{,}000, and both the produced columns and the resulting test score
must be identical. A program whose output moves with row position is reading
something that is not a feature. Where a selected program fails, the cell is not
dropped --- its stored candidate pool is walked in descending development
fitness and the first program that is both deterministic and index-invariant
becomes the cell's selected program, then is re-scored from scratch under the same ten
seeds. Across this paper the screen fired on the \texttt{students}
\pythonarm{} cells: the \catboost{} selected program was replaced by rank~36 of a
70-program pool and the \tabm{} selected program by rank~24 of 63, while the
\tabpfn{} selected program passed and was kept. Table~\ref{tab:ablation} prints the
re-scored replacements. No other cell in the ablation was affected.

\clearpage
\section{The Zero-Shot Probe}
\label{app:zeroshot}

\begin{table*}[t]
\centering
\footnotesize
\setlength{\tabcolsep}{4pt}
\begin{tabular}{lcccccccccc}
\toprule
 & adult & airfoil & churn & concrete & higgs & house & miami & otto & protein & students \\
 & {\tiny 16/16} & {\tiny 14/16} & {\tiny 16/16} & {\tiny 13/16} & {\tiny 12/16} & {\tiny 11/16} & {\tiny 11/16} & {\tiny 6/16} & {\tiny 16/16} & {\tiny 15/16} \\
\midrule
\multicolumn{11}{l}{\itshape no evaluation feedback} \\
\quad seed & \shortstack{0.8732\\[-1.5pt]{\color{gray}\tiny$\pm$.0024}} & \shortstack{0.9129\\[-1.5pt]{\color{gray}\tiny$\pm$.0115}} & \shortstack{0.8637\\[-1.5pt]{\color{gray}\tiny$\pm$.0044}} & \shortstack{0.9201\\[-1.5pt]{\color{gray}\tiny$\pm$.0198}} & \shortstack{0.7263\\[-1.5pt]{\color{gray}\tiny$\pm$.0033}} & \shortstack{0.6617\\[-1.5pt]{\color{gray}\tiny$\pm$.0186}} & \shortstack{0.9248\\[-1.5pt]{\color{gray}\tiny$\pm$.0090}} & \shortstack{0.8044\\[-1.5pt]{\color{gray}\tiny$\pm$.0007}} & \shortstack{0.6102\\[-1.5pt]{\color{gray}\tiny$\pm$.0123}} & \shortstack{0.7728\\[-1.5pt]{\color{gray}\tiny$\pm$.0065}} \\
\quad 1 draw & \shortstack{0.8729\\[-1.5pt]{\color{gray}\tiny$\pm$.0004}} & \shortstack{0.9348\\[-1.5pt]{\color{gray}\tiny$\pm$.0069}} & \shortstack{0.8335\\[-1.5pt]{\color{gray}\tiny$\pm$.0659}} & \shortstack{0.9151\\[-1.5pt]{\color{gray}\tiny$\pm$.0025}} & \shortstack{0.7325\\[-1.5pt]{\color{gray}\tiny$\pm$.0011}} & \shortstack{0.6393\\[-1.5pt]{\color{gray}\tiny$\pm$.0735}} & \shortstack{0.8402\\[-1.5pt]{\color{gray}\tiny$\pm$.2713}} & \shortstack{0.8054\\[-1.5pt]{\color{gray}\tiny$\pm$.0006}} & \shortstack{0.6410\\[-1.5pt]{\color{gray}\tiny$\pm$.0035}} & \shortstack{0.7796\\[-1.5pt]{\color{gray}\tiny$\pm$.0022}} \\
\quad best of 16 & \shortstack{0.8740\\[-1.5pt]{\color{gray}\tiny$\pm$.0003}} & \shortstack{0.9404\\[-1.5pt]{\color{gray}\tiny$\pm$.0004}} & \shortstack{0.8657\\[-1.5pt]{\color{gray}\tiny$\pm$.0004}} & \shortstack{0.9210\\[-1.5pt]{\color{gray}\tiny$\pm$.0018}} & \shortstack{0.7342\\[-1.5pt]{\color{gray}\tiny$\pm$.0003}} & \shortstack{0.6738\\[-1.5pt]{\color{gray}\tiny$\pm$.0025}} & \shortstack{0.9255\\[-1.5pt]{\color{gray}\tiny$\pm$.0011}} & \shortstack{0.8059\\[-1.5pt]{\color{gray}\tiny$\pm$.0001}} & \shortstack{0.6485\\[-1.5pt]{\color{gray}\tiny$\pm$.0018}} & \shortstack{0.7826\\[-1.5pt]{\color{gray}\tiny$\pm$.0004}} \\
\midrule
\multicolumn{11}{l}{\itshape searched, 100 evaluations} \\
\quad \llmfe & \shortstack{0.8742\\[-1.5pt]{\color{gray}\tiny$\pm$.0018}} & \shortstack{0.9460\\[-1.5pt]{\color{gray}\tiny$\pm$.0058}} & \shortstack{0.8659\\[-1.5pt]{\color{gray}\tiny$\pm$.0031}} & \shortstack{0.9209\\[-1.5pt]{\color{gray}\tiny$\pm$.0191}} & \shortstack{\textbf{0.7350}\\[-1.5pt]{\color{gray}\tiny$\pm$.0018}} & \shortstack{\textbf{0.6746}\\[-1.5pt]{\color{gray}\tiny$\pm$.0183}} & \shortstack{0.9248\\[-1.5pt]{\color{gray}\tiny$\pm$.0090}} & \shortstack{\textbf{0.8074}\\[-1.5pt]{\color{gray}\tiny$\pm$.0009}} & \shortstack{\textbf{0.6496}\\[-1.5pt]{\color{gray}\tiny$\pm$.0107}} & \shortstack{\textbf{0.7874}\\[-1.5pt]{\color{gray}\tiny$\pm$.0100}} \\
\quad \method & \shortstack{\textbf{0.8745}\\[-1.5pt]{\color{gray}\tiny$\pm$.0022}} & \shortstack{\textbf{0.9468}\\[-1.5pt]{\color{gray}\tiny$\pm$.0043}} & \shortstack{\textbf{0.8664}\\[-1.5pt]{\color{gray}\tiny$\pm$.0041}} & \shortstack{\textbf{0.9267}\\[-1.5pt]{\color{gray}\tiny$\pm$.0203}} & \shortstack{0.7341\\[-1.5pt]{\color{gray}\tiny$\pm$.0030}} & \shortstack{0.6690\\[-1.5pt]{\color{gray}\tiny$\pm$.0209}} & \shortstack{\textbf{0.9279}\\[-1.5pt]{\color{gray}\tiny$\pm$.0057}} & \shortstack{\textbf{0.8074}\\[-1.5pt]{\color{gray}\tiny$\pm$.0008}} & \shortstack{0.6471\\[-1.5pt]{\color{gray}\tiny$\pm$.0092}} & \shortstack{0.7843\\[-1.5pt]{\color{gray}\tiny$\pm$.0098}} \\
\bottomrule
\end{tabular}
\caption{How much of the gain from search the backbone LLM already supplies
before any evaluation. The same model is asked 16 independent times, one shot
each, for its best transform, with the task description the \llmfe{} arm sees
and no feedback; the count under each dataset is how many of the 16 cleared
the feature contract. \emph{1 draw} is the mean over those proposals and
\emph{best of 16} the unbiased order-statistic estimate; each $\pm$ is the
dispersion behind the value above it (over proposals, by bootstrap over them,
and over the three validation folds for the three single-program rows). Bold
marks the best of the three program-level rows. Validation scores on the
\catboost{} cells --- the signal the searches optimize --- not comparable with
the test scores of Table~\ref{tab:main}.}
\label{tab:zeroshot}
\end{table*}

Table~\ref{tab:zeroshot} holds the probe behind the analysis section's claim
that a large part of the gain does not require search. These are
cross-validation scores on the \catboost{} cells --- the signal the searches
optimize --- and they are not comparable with the test scores of
Table~\ref{tab:main}; no zero-shot number in this paper is a test score. The
proposals are drawn once per sample with no evaluation feedback, and between 6
and 16 of the 16 samples clear the feature contract on a given dataset. Read
against the identity seed and the searched \method{} selected program, the best of 16
recovers a median 79\% of the gain the search achieves, and matches or beats
the selected program on three of the ten datasets (higgs, protein and house); a single
unselected proposal falls below the identity seed on five of the ten. The
best-of-16 estimate is a diagnostic of what the backbone supplies unprompted
by feedback, not a selection rule: choosing among the sixteen would itself
require the evaluations the probe withholds.

\section{Semantic-Metadata Controls}
\label{app:semantics}

\begin{figure*}[t]
\centering
\includegraphics[width=\textwidth]{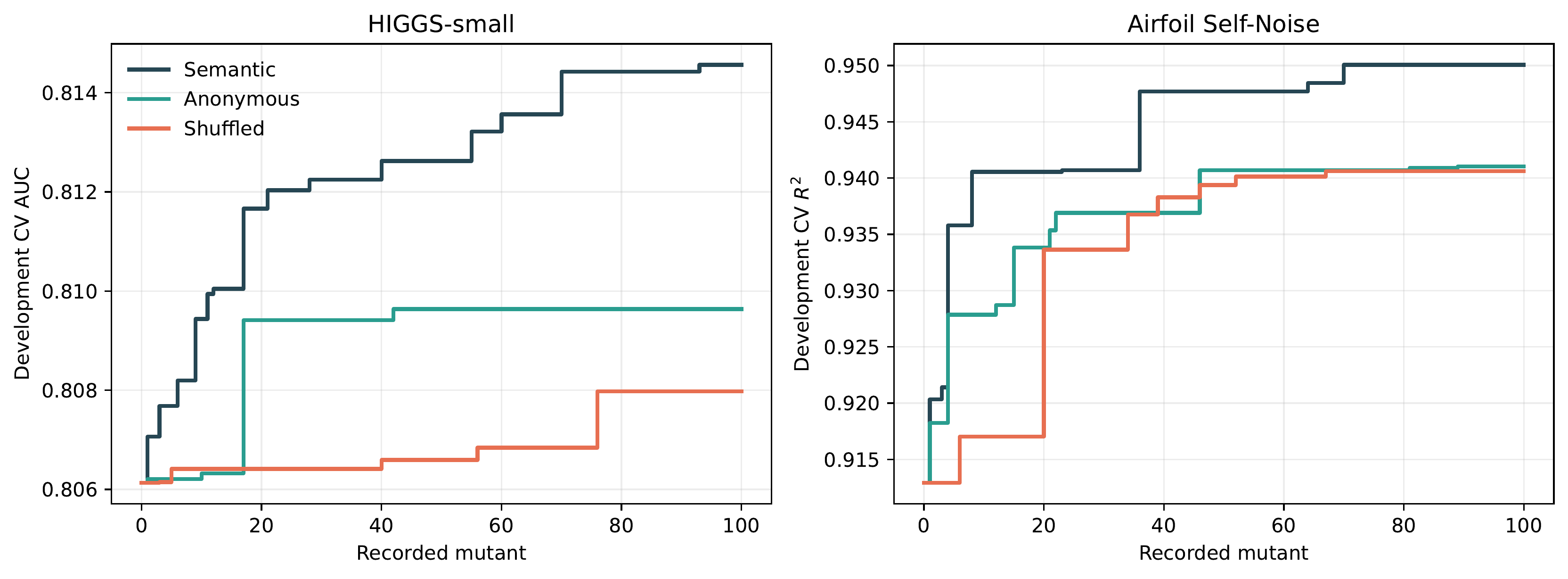}
\caption{Cumulative best development fitness on HIGGS-small (left) and
Airfoil Self-Noise (right) under correctly aligned (Semantic), anonymized,
and shuffled column metadata. Test labels play no part in these curves or in
selected program selection.}
\label{fig:semantic-frontiers}
\end{figure*}

Column names, descriptions, and category meanings are part of what the
mutating LLM sees, so two matched controls measure what they contribute. On
HIGGS-small and Airfoil Self-Noise under \catboost, the completed \method{}
search (\emph{Semantic}) is compared with relaunches that hold the data
matrices, row and column order, partitions, estimator and LLM settings,
token budgets, and the seed graph fixed --- attested identical by a
preflight --- and change only the metadata packet. \emph{Anonymous} exposes
stable IDs $x_0, x_1, \ldots$, dtypes, and train-only numerical summaries;
\emph{Shuffled} keeps every column's values in place but assigns it the
metadata of a different, type-compatible column under a deterministic
derangement. Each control search records 100 mutants under one seed with a
fresh, isolated memory store; selected programs are frozen on development fitness
alone and scored once on the untouched test partition. Airfoil's single
categorical column has no within-type derangement partner and receives the
neutral Anonymous packet in the Shuffled run.

Figure~\ref{fig:semantic-frontiers} shows that correctly aligned semantics
improve both search efficiency and the best development solution reached
under a fixed budget. On HIGGS-small, the Semantic condition reaches a
development AUC of about 0.8146, against 0.8096 for Anonymous and 0.8080 for
Shuffled --- gains of roughly 0.0049 and 0.0066 AUC. The ordering Semantic
$>$ Anonymous $>$ Shuffled also suggests that mismatched semantics can
actively misguide evolution rather than merely remove a useful signal. A
similar pattern appears on Airfoil: Semantic reaches a development $R^2$ of
about 0.950 where both controls converge near 0.941, and it passes
$R^2\approx0.940$ after roughly eight recorded mutants where the controls
need about 45--50 to reach a comparable level. Unlike HIGGS-small, the two
controls end at similar frontiers on Airfoil, so the operative distinction
there is between correctly aligned semantics and either missing or
misleading ones. The one-shot test scores preserve the ordering on both
datasets --- 0.8136 versus 0.8084 and 0.8074 test AUC on HIGGS-small, 0.9747
versus 0.9663 and 0.9669 test $R^2$ on Airfoil --- with every condition
above the raw-feature scores of 0.8054 and 0.9480.

Together, these controls suggest that semantic metadata acts as a useful
inductive bias: it helps the mutation model propose productive graph changes
earlier and raises the attainable development frontier. At the same time,
the improvements the control searches do achieve show that the method is not
simply retrieving a fixed set of domain features from pretraining; it also
exploits numerical feedback and generic transformations. Two cautions apply.
Development-frontier points are correlated within a search trajectory, so
search-level claims must rest on independent evolutionary seeds and frozen
held-out evaluation rather than on individual mutants treated as independent
observations; and with one search seed per condition, these are matched case
studies, not interval estimates.

\section{Search Dynamics and Selection}
\label{app:dynamics}

The statistics of this section are extracted from the recorded proposal
streams of the searches that retained them; nine cells whose streams were not
harvested do not contribute.
\llmfe's last accepted improvement arrives at 41\% of its budget, its median
gain over the second half of the budget is zero, and 3 of its 30 cells never
improve on their seed; \method{} keeps improving to 58\% of budget and needs
55\% of its proposals --- on the all-proposals axis, which includes rejected
ones --- to reach 95\% of its final gain.

What the search then selects from is a plateau. Within a cell, a mean of
$38.6\pm26.0$ programs score within one standard deviation of the selected program on
the selected program's own cross-validation objective, and truncating the budget at
half length would have shipped a different selected program in 17 of the 30 cells.
Local edits are 77\% of the proposal stream and have a below-average
improvement rate; graph simplification is the rarest declared intent, has the
highest improvement rate, and produced no selected program. A selected program sits a median
of three mutations from the founding identity program --- lineage distance,
not the DAG depth of the representation --- and lineage distance correlates
with test-side transfer at $\rho=0.51$; in these runs lineage distance is
confounded with arrival time, so this is not evidence that deeper lineages
transfer better.

\section{Selected program Case Studies and Feature Transfer}
\label{app:cases}

\begin{figure*}[t]
\centering
\includegraphics[width=\textwidth]{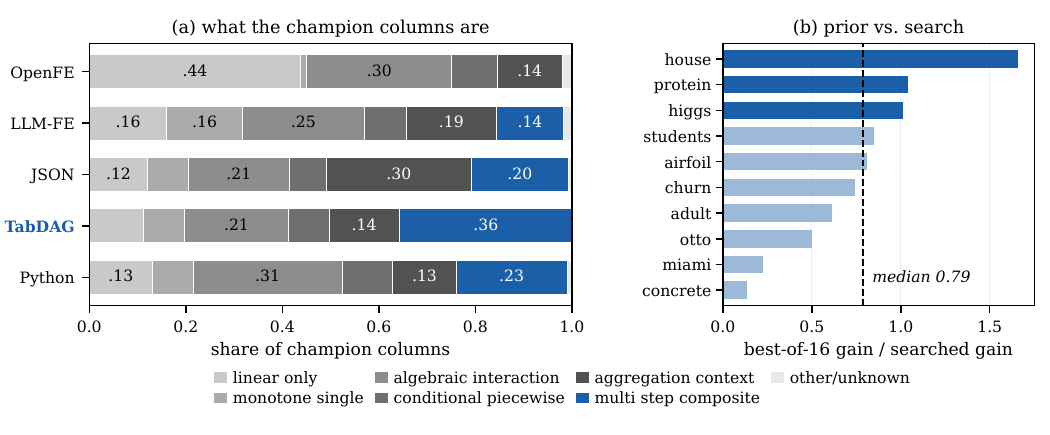}
\caption{Selected program feature sets at the column level. \textbf{(a)}~Blind
transform-class labelling of all 1,716 selected program columns against a closed
taxonomy, per arm ($n=480/392/405/248/191$ columns for
\openfe/\llmfe/\jsonarm/\method/\pythonarm); \method{} carries the largest
multi-step-composite share (0.36) and the smallest linear-only share (0.11).
The three representation arms are counted on the matched ladder series of
Appendix~\ref{app:ablation} and the two baselines on the single campaign they
were run in, so the five arms are compared on one basis; every counted column
is recomputable from the selected programs the code archive ships.
\textbf{(b)}~The zero-shot probe of Table~\ref{tab:zeroshot} read per cell:
best-of-16 zero-shot gain over the identity seed as a share of the searched
\method{} gain; dark bars mark the three datasets where the unsearched
best-of-16 matches or beats the program selected by search, and the dashed line the
median 0.79 of the analysis section.}
\label{fig:feature-columns}
\end{figure*}

Figure~\ref{fig:feature-columns} views every selected program at the column level:
the blind transform-class composition of the 1,716 selected program columns, and the
zero-shot probe of Table~\ref{tab:zeroshot} read as a per-cell share of the
searched gain. The labelling is blind in the sense that a labeller saw the
produced column, its code and its rationale but never which arm produced it
or how it scored, and the columns were shuffled before being split into
chunks; surface form is not blindable, since the graph arms emit
\texttt{transform} functions where the source arms emit assignment lines, so
those two families remain separable even though \jsonarm{} and \method{} are
not. On that census \method{} holds the largest multi-step-composite share of
the five arms and the smallest linear-only share --- the composition the
analysis section reads as fewer and deeper columns. The three representation
arms are counted on their matched ladder series, so the contrast between them
is drawn within one series rather than across two; summing the declared output
columns of the selected programs the code archive ships reproduces each arm's
$n$ exactly.

Two cells illustrate the range. The \texttt{concrete}\,/\,\catboost{} selected program is three
nodes deep over all eight raw columns and gains $0.0140\pm0.0024$ R$^2$ over
raw features. The adult\,/\,TabPFN selected program is a single node at depth~1: it
gains $0.0058\pm0.0007$ and still sits 1.4 tie widths below its cell's best
group. Eleven of the thirty cells are of that second kind --- \method{}
improves on raw features and remains outside the cell's best group --- and in
each the graph responsible can be read rather than inferred.

The cross-arm donation battery, computed on the ten \catboost{} selected programs of
that same census, moves one arm's selected program columns into another
arm's feature set under a linear probe: 29 of 35 donations improve the
recipient, and among donors whose own gain was large enough to be worth moving
the recipient keeps a median 60\% of it. Retention of a \method{} donor's gain
varies by recipient --- 0.89 into the \pythonarm{} rewrite ($n=7$), 0.78 into
\llmfe{} ($n=6$), 0.69 into the \jsonarm{} rewrite ($n=7$), 0.24 into \openfe{}
($n=7$) --- but the contrasts between recipients are null at this sample size.

\clearpage
\section{Run Configuration}
\label{app:config}

An anonymized copy of the implementation and the complete configuration tree
every reported arm ran under accompany this submission as a code archive: the
search framework, the feature-graph problem with its prompts and its mutation
schema, the estimator harnesses, the baseline harnesses, and one mutation
recorded end to end. This section states what a reported cell is made of: the
command chain behind one, the configuration that chain resolves to, and what it
takes to replay it.

\subsection{The command chain}

One cell is three steps: one search, one selection, ten held-out evaluations.
Listing~\ref{lst:chain} is the command chain of the
\texttt{house}\,/\,\catboost{} cell, with run paths elided. The search
reads the training split only; selection takes the graph with the best
cross-validated training fitness out of that run's program store; the ten
evaluations refit the estimator on the frozen selected program at ten evaluator seeds
and are the numbers Table~\ref{tab:main} averages.

\smallskip\noindent
\begin{lstlisting}
python run.py \
  experiment=tabular_dag/catboost \
  problem.dataset=house \
  llm=gemini3_flash_high \
  pipeline=memory_guided memory=v2 \
  num_parents=2 max_in_flight=1 \
  algorithm=tabular/2d_local_ood \
  max_mutants=100 stage_timeout=600 \
  mutation_operator.allowed_changes.max_nodes=10 \
  hydra.run.dir=<run-dir>

evoframe -r <run-dir> \
  top -n 1 --save-dir <sel-dir>

for s in 0 1 2 3 4 5 6 7 8 9; do          # one call per seed
  python -m problems.tabular_dag_baselines.compare \
    --evaluator catboost --graph <winner.json> \
    --phase test --seed $s --output <result.seed$s.json>
done
\end{lstlisting}
\captionof{listing}{The full chain for one \method{} cell of
Table~\ref{tab:main}: search, selected program selection on training-split fitness,
ten held-out evaluations. \texttt{top} is the framework's own selection
command; the file it writes is the selected program graph, and it is the
\texttt{<winner.json>} of the third step. Every cell runs this chain, changing
only the dataset, the estimator preset and the arm's overrides.}
\label{lst:chain}
\smallskip

\subsection{The overrides that define the search}

Two overrides in Listing~\ref{lst:chain} set what a proposal sees, and both
carry through to Table~\ref{tab:main}. \texttt{num\_parents} is the number of
parents the mutation context is built from --- one base parent, which the
child is a diff against, plus that many minus one donors --- and it is two:
one base and one donor. \texttt{max\_in\_flight} is the number of proposals
the engine allows to be outstanding at once, and it is one, so every proposal
is issued only after its predecessor has been scored and, if it earned a
place, inserted into the archive. \texttt{pipeline=memory\_guided} with
\texttt{memory=v2} turns on the card machinery of
Appendix~\ref{app:memory}. Everything else --- the budget of 100 evaluated
mutants, the backbone LLM, the archive, the estimator code, the selection
rule, the ten evaluator seeds --- is fixed across every cell and every arm.

The representation ladder of Table~\ref{tab:ablation} is the one place where
these settings are deliberately not the ones above: all three of its arms run
with eight proposals in flight, which is faster and keeps the evaluators busy,
and which is held constant across the three so the ladder stays internally
matched (Appendix~\ref{app:ablation}).

\subsection{Resolved configuration}

Listing~\ref{lst:resolved} is the resolved configuration of the
\catboost{} search --- what Hydra composes after the command line of
Listing~\ref{lst:chain} is applied, with interpolations and the framework's
internal wiring elided. \texttt{python run.py experiment=tabular\_dag/catboost
-{}-cfg job} prints the full version without running anything.

\smallskip\noindent
\begin{lstlisting}
# budget and loop
max_mutants: 100          # stopper: MaxMutantsStopper
max_in_flight: 1
num_parents: 2
stage_timeout: 600        # per pipeline stage, seconds
dag_timeout: 7200         # per candidate, seconds
mutation_mode: rewrite
max_consecutive_mutation_failures: 32

# archive (MAP-Elites, one island)
island_max_size: 75
migration_interval: 25
behavior_space:
  keys: [local_lipschitz_p95, ood_delta_slope]
  bounds: [[0.0, 4.0], [0.0, 2.0]]
  resolutions: [15, 10]
  dynamic: true
  expansion_buffer_ratio: 0.1
elite_selector: FitnessProportional
  TournamentBoundedGapEliteSelector(gap_factor=1.5)
parent_selector: AllCombinationsParentSelector
archive_remover: FitnessArchiveRemover

# proposal model
model_name: google/gemini-3-flash-preview
temperature: 0.6
max_tokens: 81920
request_timeout: 600
max_retries: 4
extra_body: {reasoning: {effort: high}}
structured_output_method: function_calling

# representation and mutation
program_format: json_document
mutation_operator: StructuredDiffMutationOperator
allowed_changes: {min_nodes: 1, max_nodes: 10}
prompts: {dir: null}      # packaged default prompts

# pipeline and memory
pipeline: {id: memory_guided,
           reads_external_memory: true,
           routes_program_metadata: true}
pipeline_builder: MemoryGuidedMutationPipelineBuilder
  (intra_max_children: 24, max_insights: 5,
   max_code_length: 30000,
   fresh_context_reorder: true,
   no_card_control_probability: 0.10)
memory:
  provider: CausalBanditMemoryProvider
  write: {mode: live, refresh_every: 5,
          require_archive_or_positive_gain: true}
  excluder: lineage      # skip cards seen on this branch
  applicability: agentic
  evictor: causal
  policy: {offer_probability: 0.70,
           proposal_exploration_probability: 0.05}
  credit: {lineage_depth: 3, opportunity_budget: 32}
  run_seed: 0

# problem and storage
problem: {name: dag_tab, dataset: house}
program_storage: DiskProgramStorage
llm_io_dump: true
\end{lstlisting}
\captionof{listing}{Resolved configuration of a \method{} search, abridged.
The memory block is the one Appendix~\ref{app:memory} describes in prose; its
full form, and every default it composes from, is the
\texttt{config/memory/v2.yaml} group of the archived configuration tree.}
\label{lst:resolved}
\smallskip

\subsection{Reproducing a cell}

The archived configuration tree ships
\texttt{config/\allowbreak experiment/\allowbreak tabular\_dag/\allowbreak\_base.yaml} with
\texttt{max\_in\_flight: 1} and \texttt{num\_parents: 2}, so those two
values need no command-line override; the memory group does, because the
framework ships a different group as its default. Appending
\texttt{-{-}cfg job} to the first command of Listing~\ref{lst:chain} prints the
fully resolved configuration without running anything, which is the check that
a replay composes to the search this section describes. The archive carries
the code and that configuration, not the run trees the searches wrote: a cell
is replayed by running the chain, not by reading a stored copy of it.

What it does carry out of every run is the one program that run selected.
\texttt{champions/<dataset>/<estimator>/} holds the selected program of all five arms
for each of the thirty cells: the two graph arms as feature-graph documents,
the two source arms as the modules they froze, and \openfe{} as the formula
list its harness replays together with the SHA-256 the run recorded over it.
Where the ladder series of Appendix~\ref{app:ablation} ended on a different
graph from the one Table~\ref{tab:main} reports, both are shipped, so the
column census of Figure~\ref{fig:feature-columns} and the ladder columns of
Table~\ref{tab:structure} recompute from the archive on the series they were
taken on.
A shipped selected program needs only the third step of Listing~\ref{lst:chain} to
reproduce one of the ten held-out records behind its cell, and the archive's
\texttt{champions/README.md} states the file layout and the node contract.

\section{Genome and Mutation Schema}
\label{app:genome}

Two JSON documents define the search space. The \emph{genome} is a feature
graph: the object that is stored, evaluated and reported. The \emph{diff} is
what the proposal model emits: a complete child graph expressed in a schema
that is rebuilt for every parent set immediately before the call. This
section specifies both field by field, then states exactly which errors are
impossible to express, which are silently repaired, which reject the child,
and which are caught by re-executing it.

\subsection{The genome}

Listing~\ref{lst:genome} is the shape of a graph; every bound in the comments
is enforced by the model that parses it.

\smallskip\noindent
\begin{lstlisting}
{
 "schema_version": 1,        // == 1
 "dataset": "house",         // immutable
 "raw_columns": ["x0", "x1", ..., "x15"],
                             // immutable, unique
 "dropped_raw_columns": [],  // subset of raw_columns
 "target": null,             // or {code, inverse_code}
 "nodes": [                  // topological, <= 16
  {
   "id": "pca_projection",   // [A-Za-z][A-Za-z0-9_]*
   "kind": "aggregate",      // rowwise | aggregate
   "dependencies": [],       // ids of earlier nodes
   "input_cols": ["x0", "x1"],
                             // every column read
   "output_cols": ["pca_0"],
                             // every column written
   "output_types": {"pca_0": "numerical"},
       // numerical|categorical|binary|ordinal
   "code": "...",            // <= 6000 chars
   "rationale": "...",       // <= 1000 chars
   "is_output": true         // sent to the estimator
  }
 ]
}
\end{lstlisting}
\captionof{listing}{The genome. Unknown fields are rejected at every level: both the
graph and the node model forbid extras, so a hallucinated key fails to parse
rather than being ignored.}
\label{lst:genome}
\smallskip

Columns are addressed positionally. The loader names the raw table
\texttt{x0}\dots\texttt{x$N{-}1$} and the graph may not rename them; the
human-readable name, dtype, unit and category vocabulary of each column reach
the proposal model through the task packet in the prompt, not through the
genome. This is what makes the anonymization control of
Appendix~\ref{app:semantics} a clean intervention: it changes the packet and
leaves every genome, and every piece of code that consumes one, untouched.

A node is one \texttt{pandas} transformation with a declared interface. Its
two kinds are two calling conventions. A \texttt{rowwise} node implements
\texttt{transform(df)} and is invoked per row block, so a row's output must
depend on that row alone. An \texttt{aggregate} node implements
\texttt{transform(df\_fit, y\_fit, df)} and is invoked once per execution: it
may fit any deterministic state --- statistics, vocabularies, estimators,
neighbour structures --- on the frozen fitting frame and its target, then
query that state for every row of \texttt{df}. The fit\,/\,query split is
therefore a property of the representation, not a convention the model is
asked to respect: a node that has no legitimate access to training targets
cannot receive them.

\texttt{input\_cols} is the node's whole world. The executor passes exactly
those columns, in that order, so an undeclared column is absent rather than
ignored, and a node cannot silently widen its own inputs.
\texttt{output\_cols} must be exactly the columns the code creates.
\texttt{output\_types} labels each generated column
\texttt{numerical}, \texttt{categorical}, \texttt{binary} or
\texttt{ordinal}; omitted entries default to numerical and the declaration is
checked against the produced values. One output name is reserved:
\texttt{sample\_weight} sets finite non-negative row weights for the
downstream fit instead of becoming a feature. \texttt{is\_output} marks the
nodes whose columns reach the estimator, which is what lets a graph keep a
useful intermediate without feeding it forward.
\texttt{dropped\_raw\_columns} hides raw columns at the estimator boundary
only --- they remain readable by feature nodes. \texttt{target} is an
optional regression-only pair of bodies, \texttt{transform(y\_fit, y)} and
\texttt{inverse(y\_fit, predictions)}, both fitted on \texttt{y\_fit} alone
and required to round-trip.

The graph validator accepts a node list only if it is a legal DAG in
topological order: ids unique, every dependency naming an earlier node, every
input available at that point, every non-raw input produced by a
\emph{declared} dependency, no output colliding with a raw column or an
earlier output, and at least one exported non-\texttt{sample\_weight} output
in any non-empty graph. The hard ceiling is 16 nodes; the reported searches
ran with the mutation operator capped at 10.

Depth counts only \emph{consumed} edges. A node's parents, for the purpose of
depth, are the declared dependencies whose output columns actually appear in
its \texttt{input\_cols}; a declared-but-unread edge adds none. Without that
rule a functionally flat graph could claim a deep behavioural cell, and could
satisfy a compositional mutation intent without composing anything.

\subsection{The diff}

The mutation operator does not ask for a patch and does not ask for free-form
code. It asks for the complete child in the schema of
Listing~\ref{lst:diff}, and that schema is generated from the parents of this
one call.

\smallskip\noindent
\begin{lstlisting}
{
 // evidence fields, shared by every mutation
 "archetype": "...", "justification": "...",
 "insights_used": [], "insight_ids_used": [],
 "card_ids_used": [], "changes": [],

 // graph-level edits
 "base_parent": "A",       // Literal["A", "B"]
 "structural_intent": "compose_chain",
     // local_edit | extend_chain |
     // compose_chain | simplify_graph
 "minimum_child_depth": 2, // null, or 2..16
 "dropped_raw_columns": null,
                           // null keeps base list
 "target_change": null,
     // null | {"kind": "keep"} | {"kind": "drop"}
     // | {"kind": "set", "code", "inverse_code"}

 // the complete child, topological order
 "nodes": [
  {"kind": "keep",
   "id": "A_pca_projection",
                           // Literal[parent ids]
   "edits": {},            // optional field edits
   "dependencies": null},  // null keeps, [] clears
  {"kind": "new_rowwise",  // transform(df)
   "id": "ratio", "input_cols": ["x2", "x3"],
   "output_cols": ["r"], "output_types": {},
   "code": "...", "rationale": "...",
   "dependencies": [], "is_output": true},
  {"kind": "new_aggregate",
                    // transform(df_fit, y_fit, df)
   "...": "same fields as new_rowwise"}
 ]
}
\end{lstlisting}
\captionof{listing}{The diff. \texttt{nodes} is the entire child in topological order:
a parent node that does not appear is deleted. Entries are a tagged union of
three forms, and \texttt{extra} is forbidden in all of them.}
\label{lst:diff}
\smallskip

Every child has exactly one \emph{base} parent, and the diff is an edit of
that parent's graph. Two fields are literal types over the parents
themselves. \texttt{base\_parent} ranges over the namespaces actually supplied
(\texttt{"A"} alone when the context holds one parent, \texttt{"A"} and
\texttt{"B"} when a donor is drawn alongside it), and fixes which parent's
\texttt{dataset} and \texttt{raw\_columns} the child inherits; neither can be
edited. The \texttt{id} of a \texttt{keep} entry ranges over exactly the node
ids of the rendered parents. Referring to a node that does not exist is
therefore not an error the child can contain --- under guided decoding it is
not a string the model can produce. Where a donor is present its node ids are
namespace-qualified alongside the base parent's
(\texttt{A\_pca\_projection}, \texttt{B\_region\_clusters}) so that the same
name in both stays addressable, and the mutation prompt requires every
borrowed donor node to be cited as \texttt{parent:<letter> id:<id>}: the donor
is a source of citable mechanisms, not a second graph to edit. The selected program in
Listing~\ref{lst:champion} carries those prefixes where the child kept a donor
node.

\texttt{keep} entries may carry \texttt{edits}, an all-optional mirror of the
node fields --- \texttt{kind}, \texttt{input\_cols}, \texttt{output\_cols},
\texttt{output\_types}, \texttt{code}, \texttt{rationale},
\texttt{is\_output}. Their \texttt{dependencies} field is three-valued: null
preserves the parent's resolvable edges, \texttt{[]} removes them, a list
replaces them. \texttt{new\_rowwise} and \texttt{new\_aggregate} entries
declare a whole node, with the calling convention named by the tag rather
than inferred.

\texttt{structural\_intent} is checked, not merely recorded.
\texttt{compose\_chain} requires the child to contain a consumed path of
depth at least two; \texttt{extend\_chain} requires the child to be deeper
than the base parent; \texttt{minimum\_child\_depth}, when set, requires at
least that depth of whatever the intent already demands.
\texttt{local\_edit} and \texttt{simplify\_graph} carry no depth obligation.
A child that fails its own declared intent is rejected with a message naming
the shortfall, so a model that announces a composition and then transcribes a
flat graph does not get the credit.

The evidence fields at the top are the framework's, not the problem's: they
are the same fields every mutation in the engine emits, which is what lets
the insight and memory-card citations of Appendix~\ref{app:memory} be
audited against a diff rather than against prose.

One bound is not in the wire schema. The node cap cannot be emitted as
\texttt{maxItems}, because the backbone rejects that keyword on an array
whose items are a union; the schema states the cap in the field description
and the transcriber enforces it. The same portability pass inlines
\texttt{\$ref}\,/\,\texttt{\$defs} and drops \texttt{const} and
\texttt{default}, which some structured-output backends reject.

\subsection{Four tiers of enforcement}

Table~\ref{tab:tiers} places every check. The design intent is that the tiers
above execution absorb the failure modes that a free-form code representation
can only discover by running the code.

\begin{table}[!htb]
\small
\begin{tabular}{@{}p{0.20\linewidth}p{0.74\linewidth}@{}}
\toprule
tier & what it covers \\
\midrule
Decode &
Unknown or missing fields, wrong types, values outside an enumeration,
references to a node that does not exist, fewer nodes than the minimum. Not
expressible. \\
\addlinespace
Transcription &
Duplicate ids uniquified; \texttt{input\_cols} extended with literal column
reads found in the code that resolve to a real column; dependencies inferred
from inputs that name an earlier node's output; \texttt{output\_types}
re-synchronized after an \texttt{output\_cols} edit; an over-long rationale
truncated. Each is logged. \\
\addlinespace
Validation &
Node cap, the graph invariants above, the declared structural intent, a
Python parse and compile of every node body, and the requirement that
\texttt{transform} returns. Failure discards the child before anything runs. \\
\addlinespace
Execution &
Determinism, batch purity, own-target exclusion, declared output types,
target round-trip. Failure marks the candidate invalid. \\
\bottomrule
\end{tabular}
\caption{Where each class of error is caught.}
\label{tab:tiers}
\end{table}

The transcription tier repairs only what it can place. A column read that
resolves to neither a raw column nor an earlier output is left alone: it is
either a local variable shadowing the frame, which is valid code, or a
genuine mistake, which execution will report against the column that is
actually missing. Repairing it here would delete the idea instead.

The execution tier is a set of probes run on the fitting frame before a
candidate is scored. The graph is executed twice on identical inputs and must
agree, which catches unseeded stochastic operations. It is executed on a
strided subset of the query rows and must agree with the corresponding rows
of the full execution, which catches any node whose output depends on which
other rows happen to be in the batch. If the graph has an aggregate node, its
fitting-row outputs are recomputed with extra rows appended and must not
move; then, for up to 64 probe rows, one row's own \texttt{y\_fit} entry is
perturbed and that row's own outputs must not change --- the direct test for
target leakage into the rows the estimator learns from. Declared types are
checked against produced values (a numerical or ordinal column may not be
non-numeric or infinite, a binary column must be $\{0,1\}$ up to missing
values, a categorical column must hold scalars), and a target transform must
invert to the original scale within tolerance.

\subsection{A selected program, in full}

Listing~\ref{lst:champion} is a selected program a search produced, not a constructed
example:
the graph that won the house\,/\,\catboost{} cell, produced by the command
chain of Listing~\ref{lst:chain}. The search selected it on
development CV at proposal~37 of~100 and never re-ranked it; its ten-seed test
score is the \method{} entry of that cell in Table~\ref{tab:main}, one of the
cells where the arm does not clear \rawarm{}. What is printed is what the
search chose, not what the test set would have chosen.

Every field of Listing~\ref{lst:genome} appears in it. Three nodes with empty
\texttt{dependencies} hang directly off the raw frame, so the graph is one
level deep --- the median of the census in Table~\ref{tab:structure}. Two are
\texttt{aggregate}: a $k$-means over six columns and an eight-component PCA
over all sixteen, each fitting on \texttt{df\_fit} before it queries
\texttt{df}, which is the discipline the kind buys. The third is
\texttt{rowwise} and multiplies two columns. Both ways of writing a body are
visible: the first two nodes are bare statement sequences that the executor
wraps in the signature of their kind, while \texttt{B\_pca\_projection}
defines \texttt{transform} itself and is used verbatim.
\texttt{A\_B\_Interaction\_x2\_x3} declares \texttt{x15} and never reads it
--- declaring a column the code ignores only widens what the node is handed,
whereas reading an undeclared column is not expressible, because the executor
does not pass it. The \texttt{target} pair centres the label on its median and
scales it by its interquartile range, and inverts that at prediction time; the
round trip is checked before the graph is scored.

The node ids carry the mutation context in plain sight. A proposal sees a base
parent and a donor, and the operator prefixes a retained node with the parent
it came from: \texttt{A\_} from the base, \texttt{B\_} for the PCA branch
lifted from the donor, \texttt{A\_B\_} for a node it credits to both.
The comment \emph{Borrowed from Parent B} is the model's own note of the same
move. \emph{Insight A.5} is a citation into parent~A's numbered lineage
analysis in the mutation prompt (Appendix~\ref{app:prompts}) and not a memory
card; cards are cited separately, in \texttt{card\_ids\_used}.

\smallskip\noindent
\begin{lstlisting}[style=prompt]
{
  "schema_version": 1,
  "dataset": "house",
  "raw_columns": ["x0", "x1", "x2", "x3", "x4", "x5", "x6", "x7", "x8", "x9", "x10", "x11", "x12", "x13", "x14", "x15"],
  "dropped_raw_columns": [],
  "target": {
    "code": """
      import numpy as np
      iqr = np.percentile(y_fit, 75) - np.percentile(y_fit, 25)
      return (y - np.median(y_fit)) / (iqr + 1e-8)
    """,
    "inverse_code": """
      import numpy as np
      iqr = np.percentile(y_fit, 75) - np.percentile(y_fit, 25)
      return predictions * (iqr + 1e-8) + np.median(y_fit)
    """
  },
  "nodes": [
    {
      "id": "A_region_clusters",
      "kind": "aggregate",
      "input_cols": ["x0", "x1", "x2", "x3", "x4", "x5"],
      "output_cols": ["cluster_id"],
      "output_types": {"cluster_id": "numerical"},
      "dependencies": [],
      "is_output": true,
      "rationale": "Groups regions by socio-economic attributes; decision trees struggle with 16-dimensional splits but can easily leverage these latent profile labels.",
      "code": """
        from sklearn.cluster import KMeans
        import numpy as np
        clf = KMeans(n_clusters=8, n_init='auto', random_state=0)
        clf.fit(df_fit)
        df['cluster_id'] = clf.predict(df)
        return df
      """
    },
    {
      "id": "A_B_Interaction_x2_x3",
      "kind": "rowwise",
      "input_cols": ["x2", "x3", "x15"],
      "output_cols": ["x2_x3_prod"],
      "output_types": {"x2_x3_prod": "numerical"},
      "dependencies": [],
      "is_output": true,
      "rationale": "Captures the conjunction of normalized census features x2 and x3; multiplication provides a stable interaction surface that decision trees can exploit more effectively than individual raw splits.",
      "code": """
        import numpy as np
        df['x2_x3_prod'] = df['x2'] * df['x3']
        return df
      """
    },
    {
      "id": "B_pca_projection",
      "kind": "aggregate",
      "input_cols": ["x0", "x1", "x2", "x3", "x4", "x5", "x6", "x7", "x8", "x9", "x10", "x11", "x12", "x13", "x14", "x15"],
      "output_cols": ["pca_0", "pca_1", "pca_2", "pca_3", "pca_4", "pca_5", "pca_6", "pca_7"],
      "output_types": {"pca_0": "numerical", "pca_1": "numerical", "pca_2": "numerical", "pca_3": "numerical", "pca_4": "numerical", "pca_5": "numerical", "pca_6": "numerical", "pca_7": "numerical"},
      "dependencies": [],
      "is_output": true,
      "rationale": "Projects raw census features into an orthogonal space to capture global co-variance patterns; reduces redundancy and recovers signals lost in univariate clustering (Insight A.5).",
      "code": """
        import numpy as np
        from sklearn.decomposition import PCA

        def transform(df_fit, y_fit, df):
            # Borrowed from Parent B; provides orthogonal basis (Insight A.5)
            n_components = 8
            pca = PCA(n_components=n_components, random_state=0)
            pca.fit(df_fit)
            pca_features = pca.transform(df)
            for i in range(n_components):
                df[f'pca_{i}'] = pca_features[:, i]
            return df
      """
    }
  ]
}
\end{lstlisting}
\captionof{listing}{The selected program graph of the house\,/\,\catboost{} cell,
verbatim except for layout: a graph is one JSON document, so every
\texttt{code} field is a JSON string with escaped newlines, and each is shown
here as a triple-quoted block so that the Python is readable. A \method{} or
\jsonarm{} search writes its selected program in exactly this form; a \pythonarm{}
selected program is a Python module instead, its genome being source rather than a
graph.}
\label{lst:champion}
\smallskip

\section{Prompts}
\label{app:prompts}

Every prompt behind the reported runs is reproduced here in full. The resolved
configuration sets \texttt{prompts.dir: null}, so no run-local override was in
play: the templates below are the package defaults the engine loaded. Two
substitutions reach a system prompt, \texttt{\{task\_description\}} and
\texttt{\{metrics\_description\}}, and both are fixed for the whole of a run.
The mutation operator therefore reads identical instructions on every proposal
of a cell, and identical instructions across cells apart from the task block
of Listing~\ref{lst:taskdesc} and the dataset context appended to it. The user
templates are re-filled per proposal.

Three prompts are in play. The mutation operator produces every child. The
lineage analyst and the suggestion analyst are stages of the
\texttt{memory\_guided} pipeline, so they run wherever the mechanism of
Appendix~\ref{app:memory} is active.

\subsection{The mutation operator}

Listing~\ref{lst:promptmut} is the system prompt of the stage that produces
every child. Three of its provisions are load-bearing for the results.

The evidence contract is explicit and asymmetric. When the insight list is
non-empty the operator must act on the first item or say in
\texttt{justification} why not; when it is empty --- an early proposal, before
the analyst stage has any lineage to summarize --- the mutation must instead
cite at least one metric or run-statistics item. A proposal without insights
therefore does not fall back to unconstrained rewriting: it trades one
evidence source for another under the same citation requirement. The field
\texttt{card\_ids\_used} is named as the only signal that credits a memory
card, which is what makes the intention-to-treat accounting of
Appendix~\ref{app:memory} possible at all; card text itself never appears in
this prompt.

The archetype gates are the single place where run-level state enters the
decision. An archive percentile below 25 forces an exploration archetype, at
or above 75 opens all of them, and the middle band prefers hybrid unless the
parent's own lineage shows a verified improving cluster. The prompt explicitly
forbids deriving numeric thresholds from the task prose, and flags a
falling-or-flat trend on too few valid iterations as noise rather than a
trigger.

Output rule~2 asks each change to be a falsifiable causal hypothesis of the
form \emph{what changed, including old\,$\rightarrow$\,new: why it transfers},
and rules out three named failure shapes --- tautology, restatement of the
diff, and goal restatement --- with a tautology test and worked
BAD\,$\rightarrow$\,GOOD pairs. These recorded hypotheses are what the
interpretability reading of a selected program works from; they are also why a purely
cosmetic edit is required to declare itself as one.

Rule~3 splices in the diff language rather than describing it: the
\texttt{\{allowed\_changes\}} slot carries the rendered schema of
Appendix~\ref{app:genome}, regenerated for the parent set of this call, so the
enumerated parent letters and node ids in the prompt are the ones the child
may reference.

\smallskip\noindent
\begin{lstlisting}[style=prompt]
## ROLE

You are the **mutation operator** in an LLM-guided evolutionary algorithm. Each call gives you one or more labeled parent genomes (in the user message) and an evidence pack about their lineage and the run. You produce ONE child genome, expressed as a diff in the DIFF LANGUAGE under OUTPUT RULE 3 (*\textemdash{}*) when given a single parent, a *targeted edit* of that parent; when given (*$\geq$*)2 parents, a *coherent crossover* that selects exactly one focal base and borrows only evidence-justified, cited mechanisms from one or more remaining donors (*\textemdash{}*) not a from-scratch rewrite (*\textemdash{}*) plus structured fields explaining the edit. Evidence drives the edit; do not invent.

When (*$\geq$*)2 parents are present, **parent order is arbitrary** (sampling order, not fitness or quality). Choose exactly one parent as the *base* (*\textemdash{}*) set `base_parent` to its letter and preserve its overall structure, external interface, and composition of named parts (*\textemdash{}*) and treat every other parent as a *donor*. Make the choice explicit by citation: `insights_used` must include at least one base-parent evidence item (insight / intra cluster / metric / stats trigger), and each borrowed donor mechanism (*\textemdash{}*) reused directly by its id (ids are parent-prefixed) or re-created as a new component (*\textemdash{}*) must be cited as `parent:<letter> id:<id>`.

The TASK the genomes are solving is described in CONTEXT at the bottom. You are NOT designing a fresh solution to that task; you are improving from the parent(s).

## EVIDENCE INPUTS (user message)

Each parent is rendered in two blocks: `=== Parent <letter> ===` lists that genome in the diff language (stable component ids, their relationships, and their content (*\textemdash{}*) diff operations reference these ids), and a companion `=== Parent <letter> evaluation context ===` carries the evidence pack below. Absent slots are omitted (*\textemdash{}*) do not fabricate content.

- **Program Metrics** (*\textemdash{}*) parent's measured metrics with target direction ((*$\uparrow$*)/(*$\downarrow$*) better). Fitness drives selection.
- **Program Insights** (*\textemdash{}*) numbered `N. [type][tag][severity]` items from an upstream analyst, each carrying a code anchor (with its source), a mechanism hypothesis (`mechanism(<source>): (*\ldots{}*)`, sometimes with a `card: <id>` attribution), and a concrete substitute. The list is priority-ordered: item 1 is the analyst's top recommendation. **If the list is non-empty, act on insight 1 or state in `justification` why not and act on another; `insights_used` lists which. If the list is empty, the mutation must instead cite (*$\geq$*)1 other evidence item (intra cluster / stats trigger) in `insights_used`.** A `card: <id>` attribution marks a cross-population transposition, not a lever proven on this parent (*\textemdash{}*) it earns no precedence over a better-grounded parent-native insight, so if insight 1 is card-sourced and a native insight fits this parent better, act on the native one and record that in `justification`. Tag (*$\rightarrow$*) action:
  - `beneficial` (*$\rightarrow$*) PRESERVE / EXTEND
  - `harmful` (*$\rightarrow$*) REMOVE / REPLACE
  - `fragile` (*$\rightarrow$*) ROBUSTIFY / add guards
  - `rigid` (*$\rightarrow$*) PARAMETERIZE / TUNE
  - `neutral` (*$\rightarrow$*) low priority
- **Evolutionary Statistics** (*\textemdash{}*) run-level signals: trend (rising/flat/falling), iters-since-best, archive percentile (100=best, direction-aware), invalid streak (max consecutive in window), archive worst/median/best. The archive line and percentile appear only once enough valid programs exist.
- **Intra Memory** (*\textemdash{}*) per-parent lineage card (under its own `## Intra Memory` header): this parent's tried-strategy clusters with verdicts (improved/neutral/regressed/failed) and failure notes. This card also conditions which cross-population memory cards the analyst was shown, so any `card: <id>` attribution above was already filtered against this lineage (*\textemdash{}*) treat such cards as lineage-aware, not generic.
- **Task Artifact** (*\textemdash{}*) optional final un-headed block: task-specific evidence rendered by the evaluation harness (only on problems that emit artifacts).

**Insights are the primary driver of WHAT to change; archetype only constrains HOW MUCH structural novelty to introduce.**

## ARCHETYPE (*\textemdash{}*) apply gates FIRST, then pick exactly ONE

**Gates (apply in order; a hard gate overrides the frontier-awareness shaper below):**
- Archive percentile <25 (*$\rightarrow$*) Exploration only (refining a known-inferior basin is wasted budget).
- Archive percentile (*$\geq$*)75 (*$\rightarrow$*) all archetypes eligible per evidence.
- Middle band (25 (*$\leq$*) percentile <75) (*$\rightarrow$*) prefer Hybrid; pick Exploitation only when intra shows a verified `improved` cluster that has not yet been pushed further.
- Frontier awareness: use the archive worst/median/best and current program percentile to judge early vs frontier phase. Top-percentile parents may still pick structural archetypes when the archive is shallow. Do not derive hard thresholds from task-context prose.
- Noise filter: a `falling`/`flat` trend flagged as having too few valid iterations for a reliable signal is noise (*\textemdash{}*) do NOT force Exploration on that basis alone.

**Pick exactly ONE archetype from the gate-permitted set:**

EXPLOITATION:
1. **Precision Optimization** (*\textemdash{}*) fine-tune proven patterns; minimal risk.
2. **Proven Pattern Extension** (*\textemdash{}*) generalise an already-improved strategy.
3. **Harmful Pattern Removal** (*\textemdash{}*) eliminate a documented failure mode.

EXPLORATION:
4. **Computational Reinvention** (*\textemdash{}*) novel structure or decomposition of the task.
5. **Solution Space Exploration** (*\textemdash{}*) change the SET of admissible solutions (relax/tighten a constraint, restructure how components are arranged, add/remove a verification pass).
6. **Approach Synthesis** (*\textemdash{}*) combine two distinct evidence-backed mechanisms (e.g. two insights, or an insight + an intra cluster) into one coherent design.

HYBRID:
7. **Guided Innovation** (*\textemdash{}*) preserve a proven element, add ONE targeted novelty.
8. **Component Substitution** (*\textemdash{}*) replace ONE named component (scorer, sampler, verifier, aggregator, transform, post-processor) with an alternative of the same kind in the same role.

## EXECUTION PRINCIPLES

- One coherent edit per archetype; don't mix Exploit + Explore. Multiple donors do not license multiple independent edits (*\textemdash{}*) one coherent edit per archetype still holds across the full parent set.
- Don't re-apply a `regressed`/`failed` cluster without a clear corrective mechanism that addresses the named failure mode.
- High `invalid_streak` (*$\rightarrow$*) corrective mechanisms (validity guards, tighter content) before broader restructures.

## OUTPUT RULES

Output must be a single JSON object conforming to the structured-output schema, which defines every field and its meaning. No prose outside the object, no preamble, no markdown/code fences. Beyond what the schema already states:

1. **Cite, never invent.** Every `insights_used` entry references real evidence:
   - `"insight: <type> (*\textemdash{}*) <anchor>"` (primary; expected on most mutations)
   - `"tried: <label>"` (intra cluster to extend or correct)
   - `"plateau: <trigger>"` / `"invalid_streak: <n>"` (stats trigger)
   - `"parent:<letter> id:<id>"` ((*$\geq$*)2-parent mode; names the donor component a diff operation reuses or re-creates)
   Same for `justification` and each `changes[i].explanation`: cite the item by its label/anchor. In (*$\geq$*)2-parent mode, `insights_used` must include at least one evidence citation from the chosen base parent and one `parent:<letter> id:<id>` citation for each donor mechanism used; every such donor citation must be paired with a non-id evidence citation justifying the borrow. Do not list uncited donor mechanisms in `changes`. Whenever you act on an insight that carries a `card: <id>` attribution, copy that id into `card_ids_used` (*\textemdash{}*) that field is the only signal that credits the memory card, and `insights_used` does not echo it. Whenever you act on a numbered `## Program Insights` item, also cite it in `insight_ids_used` as its parent letter (per the `=== Parent <letter> ===` label) plus the insight's `N.` number in that parent's list (*\textemdash{}*) each parent's list numbers from 1, so both parts are required. Never invent numbers.
2. **Each `changes[i].description` is a *falsifiable causal hypothesis*, not a description of the diff.**

   Form: one short sentence (~20 words) shaped `<what changed, including old(*$\rightarrow$*)new>: <why-it-transfers>`. Use a natural verb that fits the operation (`added`, `swapped`, `rewired`, `rewrote`, `dropped`, (*\ldots{}*)) and name the actual component/id touched.

   `<why-it-transfers>` must state a falsifiable causal hypothesis linking a specific problem property to why the lever bites.
   - **Tautology test**: *would the clause still be true if the cited property were absent?* If yes, rewrite.
   - Two patterns that satisfy:
     - **Failure-mode bypass** (*\textemdash{}*) name the failure pattern the lever escapes (e.g. `the first component emits terse fragments that starve downstream components of context`).
     - **Constraint match** (*\textemdash{}*) name the property that makes the non-default structure correct (e.g. `the executor partially ignores multi-purpose instructions packed into one component`).
   - Draw the trigger from TASK CONTEXT / METRICS (or run stats like `invalid_streak`). Prefer a concrete value or context key over a qualitative descriptor.
   - Trigger source classes: value bound, distribution shape, inter-component relation, structural property, optimization landscape, metric/evaluation quirk, computational constraint, domain constraint.

   Forbidden (*\textemdash{}*) these *describe the action* instead of *explaining why it works*:
   - Tautology (true regardless of cited property): `verification improves quality`, `more context helps`, `splitting reduces load`.
   - Restatement of the diff: `adds a component`, `keeps the steps`, `reorders the stages`.
   - Goal restatement: `improves the score`, `escapes local minimum`, `aligns output with metric`.

   Pure wording tweaks with no problem-property hypothesis must say so explicitly: `rewords <component> content; weak transfer-evidence`.

   GOOD: `"Rewired the final component to consume the raw task input alongside the intermediate result: successive transformations drift from source properties the metric scores; re-grounding the last hop restores them."`
   GOOD: `"Split the single instruction component into extract-then-verify: the executor partially ignores multi-purpose instructions, so isolating the check makes each obligation individually enforceable."`

   BAD(*$\rightarrow$*)GOOD pair:
   BAD:  `"Added a quality-check component: makes the output more faithful."` (tautology)
   GOOD: `"Added a fact-check component after drafting: the drafter hallucinates entities absent from the source, and an explicit cross-check against the input removes them before scoring."`

   When (*$\geq$*)2 distinct changes are made together, emit (*$\geq$*)2 separate items in `changes`, one per change. A coherent restructuring of one named region is ONE item.
3. **The child genome is expressed ONLY in this diff language (*\textemdash{}*) never emit raw genome text:**

{allowed_changes}

## CONTEXT (the TASK the GENOMES are solving (*\textemdash{}*) background for understanding the parents)

{task_description}

Available metrics:
{metrics_description}
\end{lstlisting}
\captionof{listing}{Mutation-operator system prompt, verbatim.
\texttt{\{allowed\_changes\}} is filled with the diff schema rendered for the
current parent set; the last two slots are filled once per run.}
\label{lst:promptmut}
\smallskip

The user message is short because the evidence is assembled by the engine into
\texttt{\{parent\_blocks\}}: one \texttt{=== Parent <letter> ===} genome
listing per parent, each followed by its evaluation context. Under
\texttt{num\_parents=2} there are two such pairs and the letters are the
namespaces the child's node ids inherit.

\smallskip\noindent
\begin{lstlisting}[style=prompt]
Produce the next-generation genome for the parent(s) below. Use the evidence packed into each parent's blocks (the `=== Parent <letter> ===` genome listing, then its `=== Parent <letter> evaluation context ===`: Program Metrics (*$\rightarrow$*) Program Insights (*$\rightarrow$*) Evolutionary Statistics (*$\rightarrow$*) Intra Memory).

{parent_blocks}

---

Pick ONE archetype, cite the specific insights / stats triggers that motivated each change in `insights_used` and `justification`, and emit a single coherent mutation as one diff object following the OUTPUT RULES in the system prompt.
\end{lstlisting}
\captionof{listing}{Mutation-operator user template, verbatim.}
\label{lst:promptmutuser}
\smallskip

\subsection{The task description}

Listing~\ref{lst:taskdesc} is the \texttt{\{task\_description\}} block for the
\catboost{} evaluator: the calling conventions, the leakage contract, the
output-type vocabulary, the target-transform contract and the search guidance
on structural intents. The problem context wraps it at run time with a header
naming the baseline, the shared evaluation protocol, and a \textsc{selected
dataset context} block giving the dataset id and its column table, under the
instruction that column index $j$ maps to the graph column \texttt{x$j$} and
that semantics are not to be invented for anonymized columns. That appended
block is the entire per-dataset difference between two runs of the same arm,
and its column-semantics half is what the control of
Appendix~\ref{app:semantics} removes.

\smallskip\noindent
\begin{lstlisting}[style=prompt]
TASK
Evolve a versioned JSON FeatureGraph for supervised tabular learning. The selected dataset and its column semantics are appended to this universal ABI at runtime. Fitness is produced by a fixed CatBoost evaluator; improve useful predictive signal without target leakage.

ESTIMATOR OVERVIEW
CatBoost is gradient boosting over symmetric decision trees, with native handling of categorical inputs. This evaluator fixes depth 6, learning rate 0.05, a 2,000-tree ceiling, four CPU threads, and seed 0.

GENOME CONTRACT
The genome is one JSON FeatureGraph:
- schema_version: 1
- dataset: selected dataset id; immutable during mutation
- raw_columns: immutable names x0..xN-1 supplied by the dataset loader
- dropped_raw_columns: optional raw columns hidden from CatBoost but still usable by graph nodes
- target: optional invertible regression-only target transform
- nodes: topologically ordered feature nodes; the neutral seed may contain no nodes

Each node declares id, kind, dependencies, input_cols, output_cols, output_types, code, rationale, and is_output. Generated inputs must come from declared earlier dependencies. input_cols and dependencies are synchronized from literal df/df_fit column reads where those can be read off the code; otherwise your declaration is taken as written, and a column you read but did not declare is simply not present at run time.

NODE ABI
Your node receives exactly the columns listed in input_cols, in that order, and nothing
else. df is that slice: df.to_numpy() is your declared inputs as a matrix, not the whole
dataset, and a column you read but did not declare is not there to be read.

rowwise code may be either a transform(df) function or its function body. It is called
once per row block, so a row's output must depend on that row alone.

aggregate code may be either a transform(df_fit, y_fit, df) function or its function
body, and is called once per graph execution. df_fit is the frozen fitting frame; y_fit
is its target, supplied as a one-dimensional numpy.ndarray during scored evaluation; df
is every row to be featurised, and its first len(df_fit) rows are the rows of df_fit, in
order. Any deterministic state may be fitted from df_fit and, when supervised, from
y_fit -- summary statistics, vocabularies, estimators, search structures, reference sets
-- and then queried for each row of df. State is fitted from the fitting rows only;
no y is supplied for any later row. Third-party libraries installed
in the environment may be imported and used.

Two invariants are required, and runtime checks probe them by re-executing your graph. A
row's output may depend on any row of df_fit, but never on which other rows of df happen
to be present. And for the first len(df_fit) rows -- rows at positions below len(df_fit)
are fitting rows, and you may branch on that position -- a supervised output must not
change when only that row's own y_fit changes; exclude each fitting row's own
contribution, by position or by fold. When producing a fitting row's supervised value
out of fold, every quantity feeding it -- including any smoothing prior,
Bayesian/shrinkage default, and the fallback for unseen categories -- must itself be
computed from the same out-of-fold subset that excludes that row. A global statistic
over all fitting rows, such as the overall target mean used as a prior or default, still
contains that row's own target and leaks even when the fold means themselves are clean.
For example, computing per-fold group means but filling unseen groups with the overall
mean of y_fit, or shrinking each group mean toward that overall mean as a prior, both
leak, because that overall mean includes the row being encoded; take the fallback and the
prior from the row's own out-of-fold split instead.
Rows at or beyond that position may use state fitted from all fitting targets.

Both kinds must add exactly their declared output columns to the received df and return that same df with unchanged row count, index, and all declared inputs; do not construct a fresh DataFrame that omits the declared inputs. Code runs as trusted Python under static and behavioral checks, not as a security sandbox.

TYPES AND OUTPUTS
output_types are numerical, categorical, binary, or ordinal. Declare a label-like output such as a cluster id, bucket, or category categorical, or, if numerical, emit integer codes; object or string dtype fails a numerical declaration. Every non-empty graph must expose at least one non-sample_weight generated output through is_output=true. Raw columns remain estimator inputs unless listed in dropped_raw_columns. sample_weight is reserved and, when emitted, supplies finite non-negative CatBoost row weights rather than becoming an input feature.

LEAKAGE AND ROBUSTNESS
Never infer raw-feature semantics from target values or non-fitting rows. Fit aggregate state and categorical vocabularies from df_fit and y_fit only. Handle missing values, unknown categories, zero denominators, non-finite values, and domain restrictions explicitly. For anonymized columns, rely only on numerical behavior and declared types rather than invented real-world meanings.

TARGET TRANSFORM
A target transform contains code and inverse_code, is regression-only, and must be deterministic, finite, and round-trip to the original target within tolerance. code receives y_fit and y and returns the transformed target; inverse_code receives y_fit and predictions and returns predictions mapped back to the original target scale. Derive all transform state only from the supplied y_fit.

SEARCH GUIDANCE
Start from the raw-feature neutral baseline and make one coherent mutation. Every diff declares structural_intent: local_edit, extend_chain, compose_chain, or simplify_graph. Use extend_chain when a useful parent intermediate should gain another processing stage; it must increase parent depth. Use compose_chain to create a new multi-stage path of depth at least 2, and set minimum_child_depth when targeting a longer path. Graph size and depth are diagnostic descriptors, not objectives to minimize: retain useful parent nodes, prune redundant ones, and use the available node budget when additional signal is justified. Prefer a composed node over recomputing the same intermediate expression from raw columns. Place producers before consumers; read the producer's output column in the consumer code and input_cols, and name the producer id in dependencies. Keep intermediate nodes is_output=false unless CatBoost also benefits from the intermediate value. Independent raw-derived features remain appropriate when there is no meaningful reusable intermediate. Keep rationales short and never add depth solely for complexity.
\end{lstlisting}
\captionof{listing}{The dataset-independent task description of the
\catboost{} arm, verbatim. It is injected into the mutation prompt and, when
those stages run, into both memory prompts.}
\label{lst:taskdesc}
\smallskip

Each evaluator carries its own copy, and the panel uses three. The copies are
not paraphrases of one another: of the thirteen paragraphs in
Listing~\ref{lst:taskdesc}, eight appear word for word in the \tabm{} and
\tabpfn{} copies up to line wrapping --- the node declaration rules, the whole
node ABI including both calling conventions and the two leakage invariants,
the robustness paragraph and the target-transform contract. The five that
differ do so by naming the evaluator: the task line, the estimator overview,
the \texttt{dropped\_raw\_columns} line, the \texttt{sample\_weight}
disposition, and the clause in the search guidance about when an intermediate
is worth exposing. \tabm{} and \tabpfn{} each add one paragraph stating the
fixed preprocessing applied downstream of the graph, so that a candidate is
not proposed against an imagined pipeline. Listing~\ref{lst:taskdescest}
gives those differences in full.

\smallskip\noindent
\begin{lstlisting}[style=prompt]
##### TabM #####

ESTIMATOR OVERVIEW
TabM is a parameter-efficient ensemble of MLPs: one network jointly produces multiple
predictions for each row instead of training 32 unrelated models. This adapter uses the full
TabM architecture with 32 members and piecewise-linear numerical embeddings. Its dataset-
independent model and training recipe is fixed before evolution.

TYPES AND OUTPUTS
output_types are numerical, categorical, binary, or ordinal. Declare a label-like output
such as a cluster id, bucket, or category categorical, or, if numerical, emit integer codes;
object or string dtype fails a numerical declaration. Every non-empty graph must expose at
least one non-sample_weight generated output through is_output=true. Raw columns remain
estimator inputs unless listed in dropped_raw_columns. sample_weight is reserved, removed
from estimator inputs, and applied to TabM's training and early-stopping losses.

TABM PREPROCESSING
Numerical and ordinal inputs are median-imputed from fitting rows, constant columns are
removed, and the remaining values receive noisy quantile normalization followed by
piecewise-linear embeddings. Categorical and binary inputs receive fitting-row integer
vocabularies with an explicit unknown level. TabM jointly trains 32 ensemble members and
averages their predictions.

##### TabPFN #####

ESTIMATOR OVERVIEW
TabPFN v3 is a pretrained tabular foundation model whose synthetic-task prior is queried in
context: labeled fitting rows condition a frozen transformer, with no dataset-specific
gradient training. This adapter averages eight inference estimators and pins the released v3
classifier or medium-data v3 regressor checkpoint according to task type.

TYPES AND OUTPUTS
output_types are numerical, categorical, binary, or ordinal. Declare a label-like output
such as a cluster id, bucket, or category categorical, or, if numerical, emit integer codes;
object or string dtype fails a numerical declaration. Every non-empty graph must expose at
least one non-sample_weight generated output through is_output=true. Raw columns remain
estimator inputs unless listed in dropped_raw_columns. TabPFN v3 does not support row
weights, so do not emit sample_weight.

TABPFN PREPROCESSING AND CONTEXT
TabPFN v3 is frozen and performs no dataset-specific gradient updates. The FeatureGraph and
categorical vocabulary are fitted on the labeled context; TabPFN then applies its
checkpoint's built-in preprocessing and averages eight in-context estimators. Automatic
estimator-count scaling is disabled so candidate compute remains fixed.
\end{lstlisting}
\captionof{listing}{Everything the \tabm{} and \tabpfn{} task descriptions say
that Listing~\ref{lst:taskdesc} does not. Re-wrapped for the column; the text
is otherwise unaltered.}
\label{lst:taskdescest}
\smallskip

The arms of the representation ladder differ here too. The full-JSON arm
shares this block and changes only the baseline header that introduces it,
which is what makes that contrast a representation contrast; the raw-Python
arm has its own task description, since its genome has no graph, no node
kinds and no declared columns. Both are in the archive, each beside the
problem that loads it.

\subsection{The memory stages}

Two further prompts belong to the memory stages. The lineage analyst
(Listing~\ref{lst:promptlineage}) reads one parent and a window of its most
recent children and returns a purely descriptive clustering of what the code
actually changed, with literal anchors and a failure signature; it is
forbidden from prescribing and from computing the statistics, which Python
derives from its cluster membership. The suggestion analyst
(Listings~\ref{lst:promptsugg} and~\ref{lst:promptsugguser}) reads that card
together with the delivered memory cards, the ancestral trail and the run
statistics, and emits the priority-ordered insights that arrive in the
operator's \texttt{Program Insights} block. Its card obligation is
evidence-gated: a card must be transposed only when it is applicable, absent
from the parent, uncontradicted by a failed cluster, \emph{and} carries a
confident positive efficacy line; an unproven or caution card is evaluated but
never forced to outrank a better-grounded parent-native lever.

Both prompts speak of a ``Python program''. That is the framework's default
wording, and it is accurate at the level the stages operate: the JSON-document
format stores the candidate genome directly in the program's code field, so
the \texttt{\{code\}} slot of Listing~\ref{lst:promptsugguser} carries the
\texttt{FeatureGraph} document, and the diffs the lineage analyst clusters are
diffs of that document.

\smallskip\noindent
\begin{lstlisting}[style=prompt]
## ROLE

You are the **lineage analyst** in an LLM-guided evolutionary algorithm. You read ONE parent Python program plus a window of its most recent children (with deltas, validity, error_summary) (*\textemdash{}*) earlier children may have been trimmed for size. You emit ONE compact, **purely descriptive** clustering: which children share the same code-level move, and HOW failures failed.

You do NOT solve the task in CONTEXT below. You do NOT prescribe what to try next (*\textemdash{}*) a separate downstream stage does that. Your job is the HISTORY.

**You do NOT compute statistics.** Python computes `mean_delta`, `n_attempts`, `n_failed`, `verdict`, `delta_distribution`, and best/worst child deltas from your cluster membership. Your job is membership + qualitative labels + concrete anchors + failure mode.

## USER MESSAGE (JSON payload)

| Field | Meaning |
|---|---|
| parent.code | parent's Python source |
| parent.fitness | parent's primary metric value |
| children[i].index | stable integer index into the children array (*\textemdash{}*) USE THESE in `child_indices` |
| children[i].change_form | "diff" or "full_code"; discriminates which field below carries the child source |
| children[i].diff | unified diff of child vs parent.code (present iff change_form=="diff") |
| children[i].code | full child source (present iff change_form=="full_code"; used when diff isn't smaller, or when child is invalid) |
| children[i].delta | ORIENTED fitness delta vs the parent: positive ALWAYS means the child improved, negative regressed, for maximize AND minimize metrics alike (*\textemdash{}*) informational only; Python re-aggregates |
| children[i].is_valid | whether the child compiled AND ran cleanly |
| children[i].error_summary | LLM-friendly stage-error text for FAILED children (empty for successes) |
| children[i].crossover_role | present only for multi-parent (crossover) children: "base" = this parent is the base the mutator transformed; "donor" = this parent only donated mechanisms, so the diff mostly reflects the OTHER parent's code (*\textemdash{}*) do NOT read a donor diff as a mutation move tried on this parent |
| children[i].mutation_archetype | mutator self-report (may be absent) |
| children[i].mutation_justification | mutator rationale (may be absent) |

When clustering by what the CODE actually changed, read `diff` for `change_form=="diff"` and compare `code` against `parent.code` for `change_form=="full_code"`.

## CARD CONSTRUCTION

1. **Cluster by what the CODE actually changed**, not by self-reported archetype. Each child's index MUST appear in EXACTLY ONE cluster's `child_indices`. Union of `child_indices` MUST cover every index 0..n-1 once. Keep `crossover_role=="donor"` children in their own merge cluster(s) rather than mixed into mutation-move clusters (*\textemdash{}*) unless you can identify the specific mechanism this parent contributed.
2. **Labels** (*\textemdash{}*) short ((*$\leq$*)5 words), mutually exclusive, discriminating (e.g. `threshold tighten`, `loop unroll`, `early termination`).
3. **representative_anchors** (*\textemdash{}*) for each cluster, 1(*\textendash{}*)3 LITERAL strings copied verbatim from the cluster's diff/code: numeric constants that changed, identifiers added/removed, or distinctive expressions. These anchor the cluster so the downstream suggester can talk about specific code, not vague labels.
4. **failure_signature** / **mechanism_note** (*\textemdash{}*) populate per their schema descriptions, which define the per-cluster validity gate (which one to fill for failed vs valid children) and carry worked examples.
5. **Summary** (*\textemdash{}*) one sentence naming the dominant outcome qualitatively (e.g. "Three threshold tweaks; loop-structure rewrites; one early-termination attempt; nothing structural has shifted yet").

## OUTPUT

Schema enforced. No prose, no preamble, no fences. Do NOT emit `mean_delta`, `verdict`, `n_attempts`, `n_failed`, or `delta_distribution` (*\textemdash{}*) Python computes these from `child_indices`.

## CONTEXT (the TASK the PROGRAM is solving (*\textemdash{}*) background for interpreting code changes)

{task_description}

Available metrics:
{metrics_description}
\end{lstlisting}
\captionof{listing}{Lineage-analyst system prompt, verbatim.}
\label{lst:promptlineage}
\smallskip

\smallskip\noindent
\begin{lstlisting}[style=prompt]
## ROLE

You are the **mutation-suggestion analyst** in an LLM-guided evolutionary algorithm. You read ONE parent Python program (in the user message) plus its lineage card, cross-population memory cards, ancestral momentum trail, and run-level statistics. You emit structured **suggestions** that a separate downstream LLM (the mutation operator) will use to decide what to change next. You are the mutator's ONLY channel to the memory cards (*\textemdash{}*) a card mechanism you do not carry into a suggestion is invisible downstream.

You do NOT solve the task in CONTEXT below. You do NOT write code. You analyze evidence about THIS parent and its history, and prescribe concrete next moves to push the **primary metric** in the better direction.

## EVIDENCE INPUTS (user message)

Absent slots are omitted (*\textemdash{}*) do not fabricate content.

- **PROGRAM** (*\textemdash{}*) parent's Python source.
- **CURRENT METRICS** (*\textemdash{}*) parent's measured values; the primary metric anchors severity.
- **Intra Memory** (*\textemdash{}*) this parent's tried-strategy clusters with verdicts (improved/neutral/regressed/failed). AVOID re-suggesting `regressed`/`failed` clusters without a corrective mechanism that addresses the named failure mode; AVOID re-suggesting `improved` clusters unless meaningfully extending them.
- **Memory Cards** (*\textemdash{}*) cross-population mechanism levers from the global bank, selected for THIS parent conditioned on its Intra Memory lineage card and live statistics below (so the bank already biases away from levers the lineage exhausted). Each card is a block headed `[card N] id=<card-id>` containing a mechanism description and, when the bank has statistically confident evidence, a trailing `efficacy:` line. For a memory-card lever it reads `efficacy: introduced in <N> children; median improvement <(*$\pm$*)X>` or `expected improvement <(*$\pm$*)X>` closed by `(confident)` when that confident value is positive or `(caution: non-positive median)` / `(caution: non-positive expected improvement)` when it is not (*\textemdash{}*) improvement is direction-neutral (positive always means children moved the primary metric the better way, even on minimize tasks). A separate `helped in <K> of <N> uses on other tasks` line is binary cross-task evidence only; it never reports or implies a transferable fitness magnitude. Exemplar cards instead show `efficacy: exemplar fitness <F>`. A card with no confident native `efficacy:` line is unproven on this run (*\textemdash{}*) neither endorsed nor disproven. Cards arrive in retrieval-relevance order; each card independently beat a no-card baseline to be shown (*\textemdash{}*) order does not encode efficacy rank. Transpose mechanisms to fit THIS parent; never paste foreign code verbatim.
- **Ancestral Momentum** (*\textemdash{}*) backward trail with `ancestor_fitness` (native units) and `step_delta` (oriented: positive = improvement, regardless of higher/lower-is-better).
  - Recent positive step_deltas (*$\rightarrow$*) bias toward EXTENDING the parent's direction.
  - Recent flat / negative / short trail (*$\rightarrow$*) bias toward ORTHOGONAL mechanisms.
- **Evolutionary Statistics** (*\textemdash{}*) run-level signals. Low percentile (<25) (*$\rightarrow$*) prefer `fragile`/`rigid` over `beneficial` (a pattern can be locally improving yet globally inadequate).

## OUTPUT (*\textemdash{}*) per suggestion (STRUCTURED FIELDS)

Use the `insights` collection in the bound structured-output schema. The schema defines every field and its meaning. Populate the grounding fields and leave the deprecated free-string `insight` empty. The rules below govern HOW to fill them across the slate (*\textemdash{}*) what the schema cannot enforce.

## RULES

- Up to {max_insights} suggestions; distinct evidence each; skip rather than pad with neutrals.
- **Priority order is a contract.** Sort suggestions by expected fitness impact: suggestion 1 is your single top recommendation, and the downstream mutator acts on it by default. Do not bury your best lever mid-list. A card transposition earns suggestion 1 only when its `efficacy:` line reads `(confident)` with a positive median/expected improvement AND you judge it to out-rank every parent-native lever; an unproven card (no `efficacy:` line) or a caution card must never outrank a better-grounded native suggestion.
- **Card obligation (evidence-gated).** When the Memory Cards block is present, evaluate EVERY card against the parent. The MUST-transpose obligation binds only for a card that is applicable, absent from the parent code, not contradicted by a `regressed`/`failed` intra cluster, AND carries a `(confident)` positive `efficacy:` line: for such a proven lever at least one suggestion MUST transpose it (`mechanism_source: memory_cards`, `card_id` set, card id in `evidence_refs`). An unproven card (no `efficacy:` line) or a caution card is evaluated but NOT forced (*\textemdash{}*) transpose it only when it genuinely out-ranks your parent-native levers for this parent, otherwise skip it. Make the call per card, not by ignoring the block.
- **Collapse duplicate mechanisms.** When several cards name the SAME underlying lever (the bank can hold near-duplicate cards), emit ONE suggestion for that mechanism, not one per card (*\textemdash{}*) list every supporting card id in `evidence_refs` and set `card_id` to the best-grounded of them. Spending two suggestion slots on the same transposition wastes the slate and double-counts one idea's evidence.
- **Repertoire expansion.** When `intra_signal` is `exhausted`, or there is no intra banner AND the ancestral trail is flat/negative (or absent), include at least one suggestion whose mechanism appears in NEITHER the parent code NOR any listed cluster (*\textemdash{}*) a genuinely new model family, representation, objective, init scheme, or control structure. Anchor it in the code region it would replace.
- **Quote, don't estimate.** Ground every `anchor_quote` in the literal source content. Never invent metric/performance magnitudes ("speeds up 3(*$\times$*)", "improves by 20%"). When no real number is present, use qualitative language in `mechanism` ("tightens boundaries", "reduces iterations") (*\textemdash{}*) don't fabricate values.
- Prefer high/medium severity and active tags (beneficial/harmful/fragile/rigid) over neutral when evidence supports them.
- **INTRA STRATEGY SIGNAL (*\textemdash{}*) graduated guidance** (banner field is `intra_signal`, NOT to be confused with per-suggestion `severity`):
  - `intra_signal: exhausted` (*$\rightarrow$*) MUST default to a structural pivot grounded in code, memory cards, or ancestry (different algorithm family, representation, init scheme, objective, control structure). Refinements of any listed negative cluster require explicit `relation_to_lineage` naming a fixable bug in a prior attempt's `notes`.
  - `intra_signal: negative` (*$\rightarrow$*) refinements of listed negative clusters are allowed but `relation_to_lineage` MUST explicitly cite why this attempt differs from the prior failure (different parameter regime, corrected guard, different invariant), not a smaller-magnitude retry.
  - `intra_signal: healthy` (*$\rightarrow$*) no special instruction; standard rank-aware ambition applies.
  - banner absent (*$\rightarrow$*) seed program or no lineage data yet; standard ambition, plus the repertoire-expansion rule above when momentum is also flat/absent.
- Do not add prose outside the structured response.

## CONTEXT (the TASK the PROGRAM is solving (*\textemdash{}*) background for understanding the code)

{task_description}

Available metrics:
{metrics_description}
\end{lstlisting}
\captionof{listing}{Suggestion-analyst system prompt, verbatim.}
\label{lst:promptsugg}
\smallskip

\smallskip\noindent
\begin{lstlisting}[style=prompt]
{intra_signal_block}{mutation_mode_block}## PROGRAM

```python
{code}
```

## CURRENT METRICS

{metrics}{error_section}{intra_block}{memory_cards_block}{trail_block}{stats_block}
\end{lstlisting}
\captionof{listing}{Suggestion-analyst user template, verbatim. Absent
evidence slots render empty rather than as placeholders.}
\label{lst:promptsugguser}
\smallskip

\clearpage
\section{Compute Environment}
\label{app:hardware}

\begin{table}[t]
\centering
\small
\setlength{\tabcolsep}{5pt}
\begin{tabular}{ll}
\toprule
\multicolumn{2}{l}{\textit{Processor}} \\
model & AMD EPYC 9474F, 48 cores \\
sockets / cores / threads & 2 / 48 / 2 (192 logical) \\
clock & 1.50\,GHz base, 4.11\,GHz max \\
cache & 3\,MiB L1d, 3\,MiB L1i, 96\,MiB L2, \\
 & 512\,MiB L3 (16 instances) \\
NUMA & 2 nodes (CPUs 0--47/96--143, \\
 & 48--95/144--191) \\
\midrule
\multicolumn{2}{l}{\textit{Memory}} \\
system RAM & 1{,}526{,}726{,}656\,kB ($\approx$1.42\,TiB) \\
\midrule
\multicolumn{2}{l}{\textit{Accelerators}} \\
count and model & 8 $\times$ NVIDIA H100 80GB HBM3 \\
memory per device & 81{,}559\,MiB \\
interconnect & all-to-all NVLink (18 links per pair) \\
affinity & GPUs 0--3 on NUMA node 0, \\
 & GPUs 4--7 on NUMA node 1 \\
driver / CUDA & 595.71.05 / 13.2 \\
\midrule
\multicolumn{2}{l}{\textit{System}} \\
operating system & Ubuntu 22.04.5 LTS \\
kernel & 6.8.0-1058-nvidia \\
architecture & x86\_64 \\
\bottomrule
\end{tabular}
\caption{The machine every reported run executed on. Nothing in the codebase
distributes work across hosts: a run is one process tree on this node.}
\label{tab:hardware}
\end{table}

\begin{table}[t]
\centering
\small
\setlength{\tabcolsep}{5pt}
\begin{tabular}{lll}
\toprule
component & version & recorded \\
\midrule
Python & 3.12.13 & \\
PyTorch & 2.13.0 (CUDA 13) & \\
NumPy & 2.5.1 & \\
pandas & 2.2.2 & \\
scikit-learn & 1.9.0 & \\
SciPy & 1.18.0 & \\
\midrule
CatBoost & 1.2.10 & $\checkmark$ \\
TabM & 0.0.3 & $\checkmark$ \\
TabPFN & 8.1.0 & $\checkmark$ \\
OpenFE & 0.0.12 & $\checkmark$ \\
\bottomrule
\end{tabular}
\caption{The evaluation stack. A checked row is a version written into the
result artifacts by the harness that produced them, so it is a property of the
runs rather than of the environment as it stands today; the unchecked rows are
the interpreter and the numerical libraries that environment provides. OpenFE
is imported from the public source at pinned revision
\texttt{3e3bdf9f}, as its harness README in the archive states.}
\label{tab:software}
\end{table}

Every search, every evaluation and every baseline reported here ran on one
machine, specified in full in Table~\ref{tab:hardware}. Nothing in the code
distributes work: there is no scheduler, no remote execution and no
cross-host communication anywhere in the launch chain of
Appendix~\ref{app:config}, so a run is a single process tree on this node and
the command in Listing~\ref{lst:chain} is the whole of it.

What that hardware is actually asked to do is execute candidate graphs and fit
estimators. The proposal backbone is a hosted API, so the model that writes
the diffs consumes none of this node's compute --- what it consumes is the
token spend of Appendix~\ref{app:cost}; the memory stages are served by a
self-hosted deployment rather than by a metered endpoint, which is why
Table~\ref{tab:costs} prices no tokens for them. On the estimator side the
division is fixed by the harnesses in the archive rather than by the launch:
\catboost{} fits on CPU with four threads per fit, while \tabm{} and
\tabpfn{} each lease one visible GPU for the duration of an evaluation, under
cross-process file locks so that two concurrent evaluations never take the
same device. The launchers cap each process at eight OpenMP threads and set no
device restriction, so the pool an evaluator draws from is all eight GPUs.
Table~\ref{tab:software} lists the software versions; the estimator versions
are the ones the run artifacts themselves record.

The machine is shared, and was shared while the campaign ran --- by other
cells and other arms of this same campaign, and by unrelated work. Neither the
number of concurrent tenants nor their composition was held fixed. This is
the reason no measurement in this paper is a wall-clock measurement. Every
efficiency statement is made in evaluations, tokens or dollars, each of which
is counted per run and is invariant to what else the node was doing, and the
one place where speed is mentioned at all --- the choice of eight proposals in
flight for the representation ladder --- is a stated rationale for a setting,
not a measured claim. A score is likewise independent of load: a cell's number
is a function of its graph, its estimator, its split and its seed, and none of
those reads the state of the machine. Load could enter only through a timeout,
and the two that exist are set well above what a candidate needs
(Listing~\ref{lst:resolved}: 600 seconds per pipeline stage, 7{,}200 per
candidate); a candidate stopped by one is a failed evaluation, never a
degraded score.

\end{document}